\documentclass[12pt]{article}
\usepackage{graphicx}
\usepackage[T1]{fontenc}
\usepackage[utf8]{inputenc}
\usepackage{lmodern}
\usepackage[margin=1in]{geometry}
\usepackage{amsmath,amssymb,amsthm,mathtools,bm,mathrsfs,aliascnt}
\usepackage{booktabs,array,enumitem,microtype,setspace}
\usepackage[authoryear,round]{natbib}
\usepackage[hidelinks]{hyperref}
\usepackage[nameinlink,capitalise,noabbrev]{cleveref}
\allowdisplaybreaks[2]
\setlist[enumerate]{leftmargin=2em,itemsep=3pt,topsep=4pt}
\numberwithin{equation}{section}
\theoremstyle{plain}
\newtheorem{theorem}{Theorem}
\newtheorem{lemma}{Lemma}[section]
\newaliascnt{proposition}{lemma}
\newtheorem{proposition}[proposition]{Proposition}
\aliascntresetthe{proposition}
\newaliascnt{corollary}{lemma}
\newtheorem{corollary}[corollary]{Corollary}
\aliascntresetthe{corollary}
\theoremstyle{definition}
\newtheorem{assumption}{Assumption}
\theoremstyle{remark}

\crefname{assumption}{Assumption}{Assumptions}
\crefname{lemma}{Lemma}{Lemmas}
\crefname{proposition}{Proposition}{Propositions}
\crefname{corollary}{Corollary}{Corollaries}
\crefname{equation}{Equation}{Equations}
\newcounter{algorithm}
\newenvironment{procedure}[1]{\par\medskip\noindent\begin{minipage}{\linewidth}
\refstepcounter{algorithm}\hrule\smallskip
\noindent\textbf{Algorithm \thealgorithm\ (#1).}\par\smallskip}
{\par\smallskip\hrule\end{minipage}\par\medskip}
\crefname{algorithm}{Algorithm}{Algorithms}
\Crefname{algorithm}{Algorithm}{Algorithms}
\newcommand{\R}{\mathbb R}
\newcommand{\N}{\mathbb N}
\newcommand{\E}{\mathbb E}
\newcommand{\PP}{\mathbb P}
\newcommand{\cH}{\mathcal H}
\newcommand{\cV}{\mathcal V}
\newcommand{\cF}{\mathcal F}
\newcommand{\cL}{\mathcal L}
\newcommand{\cD}{\mathcal D}
\newcommand{\sD}{\mathscr D}
\newcommand{\cE}{\mathcal E}
\newcommand{\cA}{\mathcal A}
\newcommand{\cK}{\mathcal K}
\newcommand{\cW}{\mathcal W}
\newcommand{\cN}{\mathcal N}
\newcommand{\cS}{\mathscr S}
\newcommand{\Lip}{\operatorname{Lip}}
\newcommand{\clip}{\operatorname{clip}}
\newcommand{\supp}{\operatorname{supp}}
\newcommand{\tr}{\operatorname{tr}}
\newcommand{\Var}{\operatorname{Var}}
\newcommand{\argmin}{\operatorname*{arg\,min}}
\newcommand{\dd}{\,\mathrm d}
\newcommand{\ind}{\mathbf 1}

\newcommand{\ip}[2]{\left\langle #1,#2\right\rangle}
\newcommand{\proofanchor}[1]{\leavevmode\phantomsection\label{proof:#1}}
\newcommand{\resultnote}[3]{\par\smallskip\noindent
\textit{Proof.} \hyperref[proof:#2]{Appendix~\ref*{#3}}.
\par\noindent #1
\par\smallskip}

\title{Knowledge Transfer for Conditional Flow Matching:\\
Frozen Estimation and  Adaptation}

\author{
	Yian Huang\hyperlink{author-email-a}{\textsuperscript{a}} \qquad Jian Huang\hyperlink{author-email-b}{\textsuperscript{b}}\\[0.5em]
	{\small The Hong Kong Polytechnic University}\\[0.5em]
	{\small \hypertarget{author-email-a}{\textsuperscript{a}}\href{mailto:author1@example.com}{\nolinkurl{yian.huang@polyu.edu.hk
		}}
		\qquad \hypertarget{author-email-b}{\textsuperscript{b}}\href{mailto:author2@example.com}{\nolinkurl{j.huang@polyu.edu.hk
		}}}
}
\date{}

\begin{document}
\maketitle

\begin{abstract}
	Conditional generative modeling with limited target data can benefit from
	representations learned in a related source population. We develop a
	transfer learning framework for conditional flow matching that accommodates
	distinct source and target response distributions and dimensions. Two
	procedures exploit a representation learned from source data: TFM-Freeze
	estimates a target velocity head with the representation fixed, whereas
	TFM-Adapt jointly learns the head and a representation adaptation under a
	functional anchoring penalty. For explicit neural network classes, we
	establish bounds on conditional squared Wasserstein risk that account for
	source estimation, target fitting, representation mismatch, and numerical
	integration. When the source representation admits a smooth correction, adaptation replaces the frozen mismatch
	contribution by the cost of learning the correction and a controllable
	regularization bias. An information lower bound identifies regimes in which
	adaptation attains a strictly faster conditional squared Wasserstein risk
	rate than any generator restricted to the frozen context information. We
	also develop independent  validation procedure for selecting among fitted
	conditional generators. 
	We show that the selection procedure satisfies
	an oracle inequality, and the switching rule guarantees, with
	high probability, no increase in conditional energy risk relative to a
	target-only generator. Simulations and real data applications illustrate the gains from our proposed methods and the effectiveness of target validation in guarding against
	negative transfer.
\end{abstract}

\section{Introduction}\label{sec:introduction}
Conditional generative modeling seeks to recover the distribution of a
response given its covariates \citep{ZhouJiaoLiuHuang2023}. A distributional description accommodates
heterogeneity, tail behavior, and dependence, and supports prediction
beyond the conditional mean \citep{Meinshausen2006,Cevid2022}. In applications with limited observations
from the population of interest, a related source population can provide
valuable information about how the covariates should be represented
\citep{Maurer2016,Tripuraneni2020}.
This opportunity is especially attractive when a high-dimensional context
acts through a lower-dimensional feature map \citep{LeeLiChiaromonte2013}. Transferring that map can
concentrate target learning on the task-specific distribution, while
allowing the source and target response laws, and even their response
dimensions, to differ \citep{TianShen2026}.

Continuous normalizing flows construct distributions by transporting a
reference law through an ordinary differential equation
\citep{Chen2018}.
Flow matching learns the transporting velocity through
regression against velocities associated with a prescribed probability
path \citep{Lipman2023}. This formulation provides a direct route to
representation transfer for conditional generation: a source encoder
summarizes common covariates, and a target velocity head combines this
summary with the current state, time, and any task-specific covariates.
With independent source and target samples of sizes \(n_s\) and \(n_t\),
the statistical objective is to use the information in the source sample
to improve the conditional distribution generated for the target task.

Two complementary uses of a source representation are particularly
useful. Freezing the representation reduces target fitting to learning
a velocity head with a smaller input dimension. Adapting it permits the
target sample to recover additional relevant information. The structure of
the required change is central: a smooth correction involving a few
context coordinates can be considerably easier to estimate than a new
representation of the entire context. This observation motivates an
analysis that distinguishes the information supplied by the source,
the complexity of a target correction, and the accuracy of the resulting
generated distribution.

We develop this analysis for two procedures, TFM-Freeze and TFM-Adapt.
TFM-Freeze estimates a target velocity head while retaining the
source-fitted encoder. TFM-Adapt jointly estimates a target head and an
updated representation, with a functional penalty that anchors the
representation to its source fit. 
An independent validation step then compares the
generated conditional distributions and permits a data-dependent choice
between transfer and target-only learning.

\subsection{Main contributions}\label{sec:intro-contributions}
The paper makes four principal contributions.

\begin{enumerate}[label=\arabic*.]
\item \textbf{Statistical guarantees from source learning to generation.}
We establish explicit neural risk bounds for frozen transfer and a
target-only counterpart under stated smoothness and flow regularity
conditions, with source-to-target calibration for frozen transfer
(\cref{thm:freeze,thm:nt}). The bounds account for joint estimation of
the source encoder and head, target fitting, representation discrepancy,
and numerical integration. They identify how a sufficiently informative
source sample can reduce the representation-learning contribution to
target risk. The analysis uses constructive ReLU approximations that
simultaneously control function values and the actual state,
representation, and time Lipschitz constants needed for generation.

\item \textbf{A quantitative benefit from learning target corrections.}
Under smooth correctability, we compare TFM-Adapt and
TFM-Freeze using the same source fit and the same target head class
(\cref{thm:anchor}).
Adaptation
replaces a frozen mismatch contribution by
the cost
of learning the adaptation.
The target complexity counts the fitted head and adaptation.
The resulting comparison identifies
regimes in which learning a correction improves the corresponding risk
bound over freezing and reduces the representation-learning contribution
relative to fitting the full target representation from scratch.

\item \textbf{A strict advantage in actual generation risk.}
We derive an information lower bound for every generator whose
evaluation-time context is restricted to the frozen representation
together with the task-specific covariates.
The bound is determined by the component of the target conditional mean
that this reduced context does not retain. When this information loss
dominates the remaining estimation and integration errors and the
anchoring weight tends to zero, the adaptive
generator achieves a strictly faster conditional squared Wasserstein
risk rate than any generator using only the frozen context information
(\cref{cor:strict-adaptation,lem:frozen-information}). A fully specified
conditional-flow example verifies the conditions and exhibits both
vanishing adaptive risk under fixed mismatch and a polynomial-rate
separation under shrinking mismatch
(\cref{prop:adaptation-example}).

\item \textbf{Validation with protection for negative transfer.}
We construct an independent energy-score validation rule for a finite
dictionary of deployed conditional generators, including a target-only
candidate. Under bounded endpoint support, selection satisfies a
leading-constant-one oracle inequality in conditional energy risk.
A conservative switching rule also guarantees, with the prescribed
high probability, no deterioration relative to the fitted target-only
generator (\cref{thm:energy-selection}). Validation only requires one observed
response at each sampled context and simulated draws from each candidate.
The procedure thus provides a direct distributional
criterion for deciding when to retain transfer.
\end{enumerate}

Simulation studies with nonlinear representations, heteroscedasticity,
and mixture responses illustrate the gains available from adaptation.
Applications to superconducting materials and diamond prices further
demonstrate the value of combining transfer with target validation.

\subsection{Related literature}\label{sec:related-literature}
Our work connects four strands of literature.

\paragraph{Statistical transfer and shared representations.}
Statistical transfer theory quantifies how the relation between source
and target populations determines the value of auxiliary observations.
\citet{CaiWei2021} characterize this relation through minimax rates for
nonparametric classification.
For high-dimensional linear regression, \citet{LiCaiLi2022} establish
gains from informative auxiliary studies and use aggregation to obtain
an oracle comparison that includes a target-only estimator.
In shared representation learning, \mbox{\citet{Tripuraneni2020}} separate
the complexity of estimating a common representation from that of
learning a new task and quantify the role of task diversity.
For generative models, \citet{TianShen2026} introduce a shared-embedding
framework and establish distributional error bounds for diffusion
models and coupling flows.
The present paper studies conditional flow matching and also considers learning target
adaptation, with explicit adaptation complexity and a frozen-information
lower bound.

\paragraph{Flow matching and distributional error analysis.}
The regression formulation of \citet{Lipman2023} gives a tractable
conditional objective with the same population gradient as the marginal
flow-matching loss.
\citet{Benton2024} bound Wasserstein error by integrated velocity approximation
error under spatial Lipschitz control.
\citet{Gao2024} develop Gaussian interpolation flows and establish
well-posedness, Lipschitz regularity, and Wasserstein stability under
perturbations of the velocity field.
More recent statistical analyses establish nearly minimax Wasserstein
convergence rates over smooth density classes with suitable probability
paths \citep{Fukumizu2025} and account for neural estimation and Euler
discretization in convergence guarantees for deep flow-matching
estimators \citep{ZhouLiu2025}.
These results provide a foundation for studying the accuracy of a fitted
flow. Our analysis addresses the additional statistical structure created
by an estimated source representation and a target correction, and links
that structure to both upper bounds and an information-based separation
between adaptive and frozen generation.

\paragraph{Neural approximation and statistical estimation.}
Quantitative approximation theory for ReLU networks includes uniform
approximation over smooth function classes \citep{Yarotsky2017} and
Sobolev approximation that controls function values together with first
weak derivatives \citep{Guhring2020}.
\citet{SchmidtHieber2020} derives statistical risk bounds for sparse
ReLU regression estimators under compositional structure, with rates
that reflect the dimensions of the component functions.
These developments motivate explicit treatment of both approximation
accuracy and the complexity of the functions estimated from data.
For transferred conditional flows, we construct admissible approximants
under prescribed blockwise Lipschitz budgets and account separately
for source learning, the target head, and the adaptation.

\paragraph{Proper scoring rules and distributional validation.}
\citet{GneitingRaftery2007} develop strictly proper scoring rules for
probabilistic prediction, including the energy score and its 
CRPS specialization.
\citet{Sejdinovic2013} establish the equivalence between energy
distance and squared distances between kernel mean embeddings for
the corresponding kernels.
These results provide the distributional and Hilbert-space foundations
for energy-score assessment. We use this foundation to validate
conditional endpoint laws with continuously distributed contexts,
explicitly account for score simulation, and derive a finite-dictionary
oracle inequality together with a confidence-calibrated switch from
target-only generation to transfer.

\Cref{sec:model} introduces the sampling framework, neural classes,
and training procedures.
\Cref{sec:theory} presents the transfer bounds, adaptation comparison,
information lower bound, and target-only counterpart.
\Cref{sec:endpoint-safeguards} develops endpoint validation.
\Cref{sec:numerical-simulations,sec:numerical-real} report the simulation
studies and real-data applications. The appendices contain the proofs,
constructive approximation results, and numerical details.
\section{Model, neural classes, and training procedures}\label{sec:model}
\subsection{Sampling experiment and conditional generation}
For a task index \(j\in\{s,t\}\), let
\begin{equation}\label{eq:data}
 \sD_j=\{(X_{j,i},Z_{j,i}):1\le i\le n_j\},\qquad
 (X_{j,i},Z_{j,i})\stackrel{\mathrm{iid}}\sim P_j^0.
\end{equation}
The samples \(\sD_s\) and \(\sD_t\) are independent. The response
\(X_j\) is in \(\R^{d_j}\). Write the context as
\(Z_j=(Z_{j,0},Z_{j,c})\), where the common block
\(Z_{j,0}\in\mathcal Z_0=[-1,1]^p\) uses the same coordinates in both
tasks, and the task-specific block \(Z_{j,c}\in[-1,1]^{p_{j,c}}\) may
be absent when \(p_{j,c}=0\). The marginal laws of the two common blocks
need not be equal. The integers \(p,d_s,d_t,d_h\) are positive.
A representation \(h:\mathcal Z_0\to[-1,1]^{d_h}\) defines the
reduced context
\(H_{j,h}(z)=(h(z_0),z_c)\).

Fix a base law \(\gamma_j\) on \(\R^{d_j}\) with a finite second
moment. Independently of \((X_j,Z_j)\), draw \(G_j\sim\gamma_j\)
and \(T\sim\operatorname{Unif}[0,1]\). A specified interpolation
\(S_{j,\tau}=\Psi_{j,\tau}(G_j,X_j)\) is almost surely absolutely
continuous in \(\tau\), with
\begin{equation}\label{eq:path}
 S_{j,0}=G_j,\quad S_{j,1}=X_j,\quad
 U_{j,\tau}=\partial_\tau S_{j,\tau},\quad
 \E\int_0^1\|U_{j,\tau}\|^2\dd\tau<\infty.
\end{equation}
For example, the independent linear path is
\(S_{j,\tau}=(1-\tau)G_j+\tau X_j\) and
\(U_{j,\tau}=X_j-G_j\).

Define the raw regression input, its law, and the full conditional mean by
\begin{equation}\label{eq:full-mean}
 W_j=(S_{j,T},Z_j,T),\qquad \nu_j=\mathcal L(W_j),\qquad
 v_j^0(W_j)=\E(U_{j,T}\mid W_j).
\end{equation}
For a head \(v\) and representation \(h\), the composed field is
\begin{equation}\label{eq:field}
 F_{j,v,h}(x,z,\tau)=v(x,h(z_0),z_c,\tau).
\end{equation}
The symbols \(\|a\|\), \(\|A\|_{\rm op}\), and \(\|A\|_F\)
denote the Euclidean, matrix operator, and Frobenius norms.
The notation \(\tr(A)\) denotes a matrix trace, \(I_q\) is the
\(q\times q\) identity, and \(A\succeq B\) means that \(A-B\) is
positive semidefinite. The symbols \(\ind(E)\), \(\sigma(\cdot)\),
and \(\supp(P)\) denote an event indicator, a generated sigma-field,
and a topological support. A function supremum norm uses the Euclidean
output norm. Put
\(\|f\|_j^2=\int\|f(w)\|^2\nu_j(\dd w)\), and define
\begin{align}
 \cL_j(v,h)&=\E\|F_{j,v,h}(W_j)-U_{j,T}\|^2,
       \label{eq:loss}\\
 \cL_j^*&=\E\|v_j^0(W_j)-U_{j,T}\|^2,\qquad
 \cE_j(F)=\|F-v_j^0\|_j^2.\label{eq:excess}
\end{align}
Thus \(\cL_j(v,h)-\cL_j^*=\cE_j(F_{j,v,h})\).
A population risk of a fitted field always uses a fresh evaluation
observation, with the fitted field held fixed. An outer expectation
averages training and auxiliary randomness.

The exact endpoint law \(P_F(\cdot\mid z)\) is defined by
\begin{equation}\label{eq:fitted-ode}
 \dot X_F(\tau)=F(X_F(\tau),z,\tau),\qquad X_F(0)=G_t\sim\gamma_t.
\end{equation}
The numerical endpoint law \(P_F^{[J]}(\cdot\mid z)\) uses the explicit
Euler method with \(J\in\N\) equal steps and the same initial law.
For conditional probability laws with finite second moments, set
\begin{equation}\label{eq:generation-loss}
 \cW_t^2(P,P_t^0)=\E_{Z_t}W_2^2\{P(\cdot\mid Z_t),P_t^0(\cdot\mid Z_t)\},
 \quad
 W_2^2(P,Q)=\inf_{\pi\in\Pi(P,Q)}\int\|x-y\|^2\pi(\dd x,\dd y),
\end{equation}
where \(\Pi(P,Q)\) is the set of couplings of \(P,Q\). The expectation
over \(Z_t\) uses the target context marginal.

\subsection{Representation deficiency}
For a fixed representation \(h\), define
\begin{equation}\label{eq:reduced-mean}
 m_{j,h}(W_j)=\E\{U_{j,T}\mid S_{j,T},H_{j,h}(Z_j),T\},
 \qquad \cD_j(h)=\|v_j^0-m_{j,h}\|_j^2.
\end{equation}
The conditional mean \(m_{j,h}\) is viewed as a function of the raw input
\(W_j\). When a reduced-input version is needed, write
\(m_{j,h}(W_j)=g_{j,h}(S_{j,T},h(Z_{j,0}),Z_{j,c},T)\).
For a specified head sieve \(\cV_{j,N}\), define
\begin{equation}\label{eq:profile}
 \cA_{j,N}(h)=\inf_{v\in\cV_{j,N}}\|F_{j,v,h}-m_{j,h}\|_j^2,
 \qquad
 Q_{j,N}(h)=\inf_{v\in\cV_{j,N}}\cE_j(F_{j,v,h}).
\end{equation}
The exact projection identity is
\begin{equation}\label{eq:decomposition}
 \cE_j(F_{j,v,h})=\|F_{j,v,h}-m_{j,h}\|_j^2+\cD_j(h),
 \qquad Q_{j,N}(h)=\cD_j(h)+\cA_{j,N}(h).
\end{equation}
The deficiency \(\cD_j(h)\) is independent of the head sieve, unlike
\(\cA_{j,N}(h)\). \Cref{lem:projection} proves
\cref{eq:decomposition}. In particular, a common sufficient representation
\(h_0\) satisfying
\(P_j^0(\cdot\mid Z_j)=P_j^0(\cdot\mid H_{j,h_0}(Z_j))\)
has \(\cD_j(h_0)=0\).

For each original observation in \cref{eq:data}, independently generate
one auxiliary pair \((G_{j,i},T_{j,i})\). Define the indexed path by
\(S_{j,i,\tau}=\Psi_{j,\tau}(G_{j,i},X_{j,i})\) and
\(U_{j,i,\tau}=\partial_\tau S_{j,i,\tau}\). The observed regression pair is
\(W_{j,i}=(S_{j,i,T_{j,i}},Z_{j,i},T_{j,i})\),
\(U_{j,i}=U_{j,i,T_{j,i}}\). Define the empirical loss
\begin{equation}\label{eq:empirical}
 \widehat\cL_j(v,h)=\frac1{n_j}\sum_{i=1}^{n_j}
       \|F_{j,v,h}(W_{j,i})-U_{j,i}\|^2.
\end{equation}
The effective independent sample size is \(n_j\). Repeated auxiliary
draws for the same observation are a single cluster, not additional
independent data; \cref{lem:relative} also covers this implementation.
All predictor classes are taken in separable measurable versions, and all
approximate minimizers and conditional means used below are measurable.

\subsection{Explicit neural sieves}\label{sec:sieves}
For fixed input and output dimensions \(D,m\), let \(\cN_{D,m}(N)\)
be a ReLU network sieve at integer resolution \(N\ge2\).
Here \(\operatorname{ReLU}(a)=\max(a,0)\) coordinatewise. Take \(L_N\)
hidden layers of width \(W_N\), an affine output layer, at most \(S_N\)
nonzero trainable weights and biases, and entrywise parameter bound \(R_N\), where
\begin{equation}\label{eq:architecture}
 \begin{aligned}
 L_N&=\lceil a_L\log(eN)\rceil,& W_N&=\lceil a_WN\rceil,\\
 S_N&=\lceil a_SN\log(eN)\rceil,& R_N&=a_RN^{a_R}.
 \end{aligned}
\end{equation}
The positive constants \(a_L,a_W,a_S,a_R\) are fixed, sufficiently large
for the constructive comparators in \cref{lem:approximation}.
Fixed input and output clipping operations add only a fixed-dimensional
number of nonzero parameters; they are included in the network implementation
and do not change the orders in \cref{eq:architecture}.

Let \(\Pi_K\) denote coordinatewise projection onto a rectangle \(K\)
and \(\clip_b\) coordinatewise clipping to \([-b,b]\). The representation
sieve \(\cH_M\) consists of maps
\(h(z_0)=\clip_1 f(\Pi_{\mathcal Z_0}z_0)\), with
\(f\in\cN_{p,d_h}(M)\). The reduced head input dimension is
\begin{equation}\label{eq:dimensions}
 D_j=d_j+d_h+p_{j,c}+1,
\end{equation}
and its fixed rectangle is
\(\cK_j=[-K_{x,j},K_{x,j}]^{d_j}\times[-1,1]^{d_h}
 \times[-1,1]^{p_{j,c}}\times[0,1]\), where \(K_{x,j}>0\).
The head sieve \(\cV_{j,N}\) contains the zero head and all heads of the
form \(v=\clip_{b_j}f\circ\Pi_{\cK_j}\),
\(f\in\cN_{D_j,d_j}(N)\), satisfying the actual functional constraints
\begin{align}
 \Lip_x(v)&\le L_{x,j},&
 \Lip_u(v)&\le L_{u,j},&
 \Lip_\tau(v)&\le L_{\tau,j}.
 \label{eq:lipschitz}
\end{align}
The constants \(b_j,L_{x,j},L_{u,j},L_{\tau,j}\) are fixed and positive.
Each blockwise Lipschitz constant varies only its indicated argument,
with all other arguments held fixed. Set \(B_{V,j}=\sqrt{d_j}\,b_j\).
These heads are globally defined, bounded by \(B_{V,j}\), and satisfy
\cref{eq:lipschitz} outside the training rectangle as well.
\Cref{lem:approximation} constructs approximants satisfying the actual
constraints, and \cref{prop:cell-certificate} gives a finite cell-slope
characterization. 

\subsection{The two training algorithms}\label{sec:algorithms}
We introduce our proposed  algorithms in the following.
A measurable \(\epsilon\)-minimizer of an objective \(J\) over a
nonempty class means an output \(\widehat\theta\) satisfying
\(J(\widehat\theta)\le\inf_\theta J(\theta)+\epsilon\).
We use tolerances \(n_s^{-2}\) and \(n_t^{-2}\).
 This defines the statistical estimators
without asserting a convergence guarantee for a particular numerical
optimizer. The resolution parameters \(N_s,M_s,N_t\) are fixed before
training; their rate-optimal choices are given in \cref{eq:resolution-choice}.

\begin{procedure}{TFM-Freeze}\label{alg:freeze}
\textbf{Input:} independent source and target training samples;
head sieves \(\cV_{s,N_s},\cV_{t,N_t}\); representation sieve \(\cH_{M_s}\).
\begin{enumerate}
\item On the source sample, compute
\begin{equation}\label{eq:source-estimator}
 \widehat\cL_s(\widehat v_s,\widehat h_s)
 \le\inf_{v\in\cV_{s,N_s},\,h\in\cH_{M_s}}\widehat\cL_s(v,h)+n_s^{-2}.
\end{equation}
\item Keep \(\widehat h_s\) fixed. On the target sample, compute
\begin{equation}\label{eq:freeze-estimator}
 \widehat\cL_t(\widehat v_F,\widehat h_s)
 \le\inf_{v\in\cV_{t,N_t}}\widehat\cL_t(v,\widehat h_s)+n_t^{-2}.
\end{equation}
\item Return \(\widehat F_F=F_{t,\widehat v_F,\widehat h_s}\).
Generate using \cref{eq:fitted-ode}, exactly or by \(J\) Euler steps.
\end{enumerate}
\end{procedure}

For functional anchoring, let \(\sD_a=\{z_{a,i}:1\le i\le n_a\}\)
be a declared design of common-block contexts, independent of the target
training sample and its auxiliary draws. The design may be deterministic
or source-derived. A default empirical functional anchor is
\begin{equation}\label{eq:functional-anchor}
 A_s(h)=\frac1{n_a}\sum_{i=1}^{n_a}
       \|h(z_{a,i})-\widehat h_s(z_{a,i})\|^2.
\end{equation}
A second, mathematically defined functional anchor is
\begin{equation}\label{eq:sup-anchor}
 A_s^\infty(h)=\sup_{z_0\in\mathcal Z_0}
                      \|h(z_0)-\widehat h_s(z_0)\|^2.
\end{equation}
Either is a discrepancy between representation functions, not a distance
between network weight vectors. In particular, reparameterizations of
network weights that preserve the function do not change the anchor.
Other nonnegative functional
anchors with \(A_s(\widehat h_s)=0\) are allowed when the conditions of the particular theorem being used are verified.

\begin{procedure}{TFM-Adapt with a functional anchor}\label{alg:anchor}
\textbf{Input:} independent source and target training samples;
a target-independent anchor design; a fixed \(\lambda\ge0\).
\begin{enumerate}
\item Compute \((\widehat v_s,\widehat h_s)\) by
\cref{eq:source-estimator}. Construct the functional anchor \(A_s\),
for example \cref{eq:functional-anchor}.
\item Choose a declared bounded-output neural encoder class \(\cH_A\)
containing the source fit \(\widehat h_s\).
The class may
depend on source training and the anchor design, but not on target training. Jointly fit
\((\widehat v_\lambda,\widehat h_\lambda)\in\cV_{t,N_t}\times\cH_A\) so that
\begin{align}\label{eq:anchor-estimator}
 \widehat\cL_t(\widehat v_\lambda,\widehat h_\lambda)
       +\lambda A_s(\widehat h_\lambda)
 &\le\inf_{v\in\cV_{t,N_t},\,h\in\cH_A}
       \{\widehat\cL_t(v,h)+\lambda A_s(h)\}+n_t^{-2}.
\end{align}
\item Return \(\widehat F_\lambda=F_{t,\widehat v_\lambda,\widehat h_\lambda}\).
Generate using \cref{eq:fitted-ode}, exactly or by \(J\) Euler steps.
\end{enumerate}
\end{procedure}
At \(\lambda=0\), \cref{eq:anchor-estimator} defines unpenalized
joint fitting over the same declared class \(\cV_{t,N_t}\times\cH_A\).

\section{Theoretical  guarantees}\label{sec:theory}
\subsection{Assumptions and explicit rates}\label{sec:assumptions}

\begin{assumption}[Regression experiment and true flow]\label{ass:path}
For \(j=s,t\), there is a fixed \(B_{U,j}<\infty\) such that
\(\|U_{j,T}\|\le B_{U,j}\) almost surely, and
\(S_{j,T}\in[-K_{x,j},K_{x,j}]^{d_j}\) almost surely.
For almost every context \(z\), the true velocity \(v_j^0\) has a unique
characteristic solution \(\Phi_{j,\tau}(g,z)\) on \([0,1]\), for
\(\gamma_j\)-almost every initial \(g\), whose law at time \(\tau\)
is \(\mathcal L(S_{j,\tau}\mid Z_j=z)\).
These characteristics have integrable second moments. All conditional
versions used in the ODE are jointly measurable.
\end{assumption}

For \(a>0\), write \(k=\lceil a\rceil-1\) and \(\alpha=a-k\in(0,1]\).
The class \(C^a(K;B)\) consists of functions on a fixed rectangle \(K\)
with derivatives through order \(k\) bounded by \(B\), and derivatives
of order \(k\) that are \(\alpha\)-H\"older with constant \(B\).
We use Euclidean output norms and Euclidean input distance. At integer
\(a\), this means Lipschitz derivatives of order \(a-1\).
Derivatives are continuous up to the boundary.

\begin{assumption}[Smooth task-specific comparison functions]\label{ass:smooth}
For each task to which this assumption is applied, fix \(r_j>1\),
\(\beta_j>0\), and finite positive smoothness bounds. There are functions
\(h_{j,0}\in C^{\beta_j}(\mathcal Z_0;B_{h,j})\), taking values in
\([-1,1]^{d_h}\), and \(g_{j,0}\in C^{r_j}(\cK_j;B_{g,j})\),
taking values in \([-b_j,b_j]^{d_j}\), such that
\begin{equation}\label{eq:structural-bias}
 \zeta_j^2:=\|v_j^0-g_{j,0}(\,\cdot,h_{j,0}(\,\cdot),\,\cdot,\,\cdot)\|_j^2
 <\infty.
\end{equation}
For positive fixed margins \(\eta_{x,j},\eta_{u,j},\eta_{\tau,j}\),
\begin{equation}\label{eq:smooth-margins}
 \Lip_b(g_{j,0};\cK_j)\le L_{b,j}-\eta_{b,j},
 \qquad b\in\{x,u,\tau\}.
\end{equation}
The source and target comparison representations need not be identical.
Exact shared structure is the special case
\(h_{s,0}=h_{t,0}=h_0\) and \(\zeta_s=\zeta_t=0\).
\end{assumption}
The notation in \cref{eq:structural-bias} abbreviates the fully specified
map \((x,z,\tau)\mapsto g_{j,0}(x,h_{j,0}(z_0),z_c,\tau)\).
Only the source part of \cref{ass:smooth} is needed for the transfer
source rate. The target part is needed for the target-only theorem.

\begin{assumption}[Calibrated source representations and reduced-mean regularity]
\label{ass:transfer}
Fix \(r_t>1\), \(B_{g,t}<\infty\), and positive margins
\(\eta_{x,t},\eta_{u,t},\eta_{\tau,t}\). There are a deterministic
measurable set \(\cH_{\rm good}\), constants
\(\kappa\ge0\), \(\Delta\ge0\), and a deterministic number
\(\pi_s\in[0,1]\) such that
\(\PP(\widehat h_s\notin\cH_{\rm good})\le\pi_s\).
For every \(h\in\cH_{\rm good}\),
\begin{equation}\label{eq:calibration}
 \cD_t(h)\le\kappa\cD_s(h)+\Delta^2.
\end{equation}
Also, the reduced conditional mean in \cref{eq:reduced-mean} has a version
\(g_{t,h}\in C^{r_t}(\cK_t;B_{g,t})\), taking values in
\([-b_t,b_t]^{d_t}\), with
\begin{equation}\label{eq:reduced-regularity}
 \Lip_b(g_{t,h};\cK_t)\le L_{b,t}-\eta_{b,t},
 \qquad b\in\{x,u,\tau\},
\end{equation}
uniformly in \(h\). The constants are independent of sample sizes.
\end{assumption}
A direct sufficient formulation is that these conditions hold for every
source-feasible \(h\in\cH_{M_s}\), in which case \(\pi_s=0\).
A local formulation is allowed, but its failure contribution is not
discarded. Only \(\widehat h_s\), the source-fitted representation used
as the frozen comparator, needs the reduced-mean regularity in
\cref{ass:transfer}. The final jointly fitted representation
\(\widehat h_\lambda\) need not have a smooth reduced conditional mean.
\Cref{prop:density,prop:plugin} give a density-based verification and an
alternative based on a single smooth target head and functional
representation recovery. 

For integers \(n\ge2\), smoothness \(a>0\), and dimension \(D\ge1\), define
\begin{equation}\label{eq:rho}
 \ell_n=\log(en),\qquad
 \rho_{a,D}(n)=\left(\frac{\ell_n^3}{n}\right)^{2a/(2a+D)},\qquad
 K_{a,D}(n)=\max\left\{2,\left\lceil
      \left(\frac{n}{\ell_n^3}\right)^{D/(2a+D)}\right\rceil\right\}.
\end{equation}
Choose
\begin{equation}\label{eq:resolution-choice}
 N_s=K_{r_s,D_s}(n_s),\quad M_s=K_{\beta_s,p}(n_s),\quad
 N_t=K_{r_t,D_t}(n_t).
\end{equation}

For compact statements, define the source-learning and transferred
representation contributions by
\begin{align}
 R_s(n_s)&=\zeta_s^2+\rho_{r_s,D_s}(n_s)+\rho_{\beta_s,p}(n_s),
      \label{eq:source-rate-symbol}\\
 T_s(n_s)&=\kappa R_s(n_s)+\Delta^2+B_{U,t}^2\pi_s.
      \label{eq:transfer-rate-symbol}
\end{align}
Throughout, \(C\) denotes a finite positive constant that may change
between displays; \(a\lesssim b\) means \(a\le Cb\) with the stated
constant dependencies. The constants in subsequent inequalities may depend on the dimensions,
fixed rectangles, envelopes, smoothness bounds, Lipschitz budgets, and
margins, but not on \(n_s,n_t,\lambda,J\). They are not claimed to be
uniform in growing ambient dimension \(p\).

\subsection{Frozen neural transfer}
\begin{theorem}[End-to-end neural bound for TFM-Freeze]\label{thm:freeze}
Use \cref{alg:freeze} with \cref{eq:resolution-choice}. Suppose
\cref{ass:path}, the source part of \cref{ass:smooth}, and
\cref{ass:transfer} hold. Then
\begin{align}
 \E\cD_s(\widehat h_s)&\le C R_s(n_s),\label{eq:source-final}\\
 \E\cE_t(\widehat F_F)&\le C\{\rho_{r_t,D_t}(n_t)+T_s(n_s)\},
       \label{eq:freeze-risk}\\
 \E\cW_t^2(P_{\widehat F_F}^{[J]},P_t^0)
 &\le C\{\rho_{r_t,D_t}(n_t)+T_s(n_s)+J^{-2}\}.
       \label{eq:freeze-main}
\end{align}
For exact ODE integration, omit \(J^{-2}\). In the exact, globally
calibrated case \(\zeta_s=\Delta=\pi_s=0\), the bound is
\begin{equation}\label{eq:freeze-clean}
 C\left\{
 \left(\frac{\ell_{n_t}^3}{n_t}\right)^{\!2r_t/(2r_t+D_t)}
 +\kappa\left(\frac{\ell_{n_s}^3}{n_s}\right)^{\!2r_s/(2r_s+D_s)}
 +\kappa\left(\frac{\ell_{n_s}^3}{n_s}\right)^{\!2\beta_s/(2\beta_s+p)}
 +J^{-2}\right\}.
\end{equation}
\end{theorem}
\resultnote{The target fit pays only for a reduced-input neural head.
The source contribution is derived from joint neural source regression,
including its head and representation complexities. Projection and
calibration transfer this bound to the target; ODE stability and the Euler
bound complete the generation analysis. This is the main guarantee for
Algorithm~\ref{alg:freeze}.}{thm:freeze}{app:mainproofs}

\subsection{Functionally anchored neural adaptation}\label{sec:adaptation}
We analyze an explicit
neural function class and compare the two training algorithms under
the same correctability condition. 

Let \(\cS=\sigma(\sD_s,\text{source auxiliary draws},\sD_a)\),
which is independent of target training, and put
\(K_t=B_{U,t}+B_{V,t}\). Fix a nonempty coordinate set
\(I_A\subseteq\{1,\ldots,p\}\), write \(q_A=|I_A|\), and let
\(z_{0,I_A}\in[-1,1]^{q_A}\) be the selected common-context coordinates.
For a fixed \(B_R>0\), define the residual neural sieve and the
source-conditioned encoder class by
\begin{align}
 \mathcal R_{A,M}
 &=\{a=\clip_{B_R} f\circ\Pi_{[-1,1]^{q_A}}:
                  f\in\cN_{q_A,d_h}(M)\},\label{eq:adapter-sieve}\\
 h_{h,a}(z_0)&=\clip_1\{h(z_0)+a(z_{0,I_A})\},\qquad
 \cH_A(M)=\{h_{\widehat h_s,a}:a\in\mathcal R_{A,M}\}.
                  \label{eq:residual-encoder}
\end{align}
The zero residual belongs to \(\mathcal R_{A,M}\), so
\(\widehat h_s\in\cH_A(M)\). The sum, the two clipping maps, and the
parallel branches in \cref{eq:residual-encoder} are ReLU implementations.
The source branch stays fixed, but the residual network \(a\) and target
head \(v\) are fitted jointly. 
The residual can use
coordinates missing from \(\widehat h_s\). Restricting every updated
encoder to be a function of \(\widehat h_s\) alone would not recover
such missing information.

The mean empirical anchor in \cref{eq:functional-anchor}, on any declared
finite design, and the supremum anchor in \cref{eq:sup-anchor} satisfy
\begin{equation}\label{eq:anchor-upper-control}
 0\le A_s(h_{\widehat h_s,a})\le\|a\|_\infty^2,
 \qquad A_s(\widehat h_s)=0.
\end{equation}
Indeed, output clipping fixes \(\widehat h_s\) and is nonexpansive.
Any other nonnegative, target-independent functional anchor satisfying
\cref{eq:anchor-upper-control} is also permitted here. 

\begin{assumption}[Smooth low-dimensional correctability]\label{ass:correctable}
Fix \(\beta_A>0\), \(r_t>1\), finite smoothness bounds
\(B_{a,A},B_{g,A}\), and fixed positive target block-Lipschitz margins.
There are a deterministic measurable set \(\cH_{\rm good}^A\), constants
\(\kappa_A\ge0\), \(\zeta_A\ge0\), \(0\le\delta_A\le B_R\), and
\(\pi_{s,A}\in[0,1]\), with
\(\PP(\widehat h_s\notin\cH_{\rm good}^A)\le\pi_{s,A}\), such that
for every \(h\in\cH_{\rm good}^A\) there exist functions
\begin{equation}\label{eq:correctable-functions}
 a_h\in C^{\beta_A}([-1,1]^{q_A};B_{a,A}),\quad
 \|a_h\|_\infty\le\delta_A,\qquad
 g_h^+\in C^{r_t}(\cK_t;B_{g,A}).
\end{equation}
The head \(g_h^+\) takes values in \([-b_t,b_t]^{d_t}\) and satisfies
\(\Lip_b(g_h^+;\cK_t)\le L_{b,t}-\eta_{b,A}\),
\(b\in\{x,u,\tau\}\), for fixed \(\eta_{b,A}>0\). With
\(h_h^+=h_{h,a_h}\) as in \cref{eq:residual-encoder}, require
\begin{equation}\label{eq:correctability}
 \|F_{t,g_h^+,h_h^+}-v_t^0\|_t^2
       \le\kappa_A\cD_s(h)+\zeta_A^2.
\end{equation}
All smoothness bounds and margins are uniform over
\(\cH_{\rm good}^A\). They are independent of sample sizes, even when
\(\delta_A,\zeta_A,\pi_{s,A}\) vary along a sequence of experiments.
\end{assumption}
The notation \(F_{t,g_h^+,h_h^+}\) uses the composition in
\cref{eq:field}; \(g_h^+\) need not itself be a neural network or the reduced
conditional mean for its encoder.
This assumption describes a target velocity that becomes well approximated
after a smooth \(q_A\)-dimensional correction to the source representation.
The term \(\delta_A^2\) measures the size of a correctable change;
\(\zeta_A^2\) is the remaining error after correction. 
A fully specified experiment verifying it is given
in \cref{prop:adaptation-example}. In particular, no identification of a
particular source encoder in Euclidean parameter distance is assumed.

Use \(N_s,M_s,N_t\) from \cref{eq:resolution-choice}, and set
\begin{align}
 M_A&=K_{\beta_A,q_A}(n_t),&
 \mathfrak R_A(n_t)&=\rho_{r_t,D_t}(n_t)+\rho_{\beta_A,q_A}(n_t),
                  \label{eq:adapter-resolution}\\
 T_A(n_s)&=\kappa_A R_s(n_s)+\zeta_A^2+K_t^2\pi_{s,A}.
                  \label{eq:adapter-source-term}
\end{align}
Here \(R_s(n_s)\) is the explicitly derived joint source rate in
\cref{eq:source-rate-symbol}.

\begin{theorem}[End-to-end neural adaptation and the frozen comparison]
\label{thm:anchor}
Suppose \cref{ass:path}, the source part of \cref{ass:smooth}, and
\cref{ass:correctable} hold. Run \cref{alg:anchor} with
\(\cH_A=\cH_A(M_A)\), the resolutions in
\cref{eq:resolution-choice,eq:adapter-resolution}, an anchor satisfying
\cref{eq:anchor-upper-control}, and a fixed \(0 \leq \lambda\le1\).
Then
\begin{align}
 \E\cE_t(\widehat F_\lambda)
 &\le C\{\mathfrak R_A(n_t)+T_A(n_s)+\lambda\delta_A^2\},
                     \label{eq:anchor-risk}\\
 \E\cW_t^2(P_{\widehat F_\lambda}^{[J]},P_t^0)
 &\le C\{\mathfrak R_A(n_t)+T_A(n_s)+\lambda\delta_A^2+J^{-2}\}.
                     \label{eq:anchor-main}
\end{align}
For \cref{alg:freeze} using the same source fit and target head sieve,
under exactly these assumptions,
\begin{align}
 \E\cE_t(\widehat F_F)
 &\le C\{\rho_{r_t,D_t}(n_t)+T_A(n_s)+L_{u,t}^2\delta_A^2\},
                     \label{eq:freeze-correctable-risk}\\
 \E\cW_t^2(P_{\widehat F_F}^{[J]},P_t^0)
 &\le C\{\rho_{r_t,D_t}(n_t)+T_A(n_s)
                       +L_{u,t}^2\delta_A^2+J^{-2}\}.
                     \label{eq:freeze-correctable-main}
\end{align}
All constants are independent of \(n_s,n_t,\lambda,J,\delta_A\).
For exact integration omit \(J^{-2}\). The source bound
\(\E\cD_s(\widehat h_s)\le C R_s(n_s)\) remains valid.
In particular, any deterministic sequence satisfying
\begin{equation}\label{eq:lambda-choice}
 0 \leq \lambda_{n_t} \leq \min\{1,\mathfrak R_A(n_t)\}
\end{equation}
yields
\begin{equation}\label{eq:anchor-clean}
\begin{aligned}
 \E\cW_t^2(P_{\widehat F_{\lambda_{n_t}}}^{[J]},P_t^0)
 \le C\big\{&\rho_{r_t,D_t}(n_t)+\rho_{\beta_A,q_A}(n_t)
                  +\zeta_A^2+K_t^2\pi_{s,A}+J^{-2}\\
 &+\kappa_A[\zeta_s^2+\rho_{r_s,D_s}(n_s)
                            +\rho_{\beta_s,p}(n_s)]\big\}.
\end{aligned}
\end{equation}
\end{theorem}
\resultnote{Adaptation replaces the frozen correction-size contribution
\(L_{u,t}^2\delta_A^2\) by a learned residual contribution
\(\rho_{\beta_A,q_A}(n_t)+\lambda\delta_A^2\). Its proof counts the
conditional entropy of the target head and residual network, but not the
already trained source branch. A penalized oracle, valid also at
\(\lambda=0\), is applied to the actual objective; admissible neural
approximation and flow stability then
give the end-to-end bound. The frozen comparison is proved under the same
conditions.}
{thm:anchor}{app:mainproofs}
The rate improvement requires
\(\rho_{\beta_A,q_A}(n_t)+\lambda\delta_A^2=o(\delta_A^2)\) when $\lambda \to 0$, and
also requires that the common source, head, and solver terms not dominate.
The range in
\cref{eq:lambda-choice} includes \(\lambda_{n_t}=0\) at every sample
size. A zero penalty removes the contribution \(\lambda_{n_t}\delta_A^2\),
but not the residual-learning term \(\rho_{\beta_A,q_A}(n_t)\).

\subsection{When the advantage is an actual risk separation}
We next
give a separate, checkable information-loss condition that orders actual risks.
Let \(\mu_t(z)=\E(X_t\mid Z_t=z)\), and, for a fixed representation
\(h\), define
\begin{align}
 \mathcal I_t(h)
 &=\E_{Z_t}\left\|\mu_t(Z_t)
       -\E\{\mu_t(Z_t)\mid H_{t,h}(Z_t)\}\right\|^2,
                     \label{eq:frozen-mean-defect}\\
 \mathcal I_F(n_s)&=\E\mathcal I_t(\widehat h_s).
                     \label{eq:expected-mean-defect}
\end{align}
The inner conditional expectation uses a fresh target context and holds
\(h\) fixed. The outer expectation in \(\mathcal I_F(n_s)\) averages
source training. This endpoint conditional-mean defect is different from
the velocity deficiency \(\cD_t(h)\).

\begin{corollary}[Strict adaptation advantage over frozen generation]
\label{cor:strict-adaptation}
Under \cref{thm:anchor}, consider a sequence with \(\delta_A>0\).
If, for a fixed \(c_0>0\),
\begin{equation}\label{eq:strict-gap-conditions}
 \mathcal I_F(n_s)\ge c_0\delta_A^2,\qquad
 \mathfrak R_A(n_t)+T_A(n_s)+J^{-2}=o(\delta_A^2),
 \qquad 0 \leq \lambda\le1,\quad\lambda\longrightarrow0,
\end{equation}
then the numerical generation risks satisfy
\begin{align}
 \E\cW_t^2(P_{\widehat F_F}^{[J]},P_t^0)&\asymp\delta_A^2,
                     \label{eq:freeze-matched-gap}\\
 \frac{\E\cW_t^2(P_{\widehat F_\lambda}^{[J]},P_t^0)}
      {\E\cW_t^2(P_{\widehat F_F}^{[J]},P_t^0)}
       &\longrightarrow0.\label{eq:strict-generation-ratio}
\end{align}
The same assertions hold for exact integration, omitting \(J^{-2}\).
The corresponding ratio of expected flow-matching excess risks also tends
to zero. Here \(a_n\asymp b_n\) means two-sided inequalities with fixed
positive constants for all sufficiently large sample sizes.
\end{corollary}
\resultnote{Every generator whose evaluation-time context is only
\(H_{t,\widehat h_s}\) has squared Wasserstein loss at least
\(\mathcal I_t(\widehat h_s)\). This lower bound applies to Algorithm~1
regardless of its fitted head or number of Euler steps. Combining it with
Theorem~\ref{thm:anchor} proves a genuine separation. 
}
{cor:strict-adaptation}{app:mainproofs}

For example, if
\(\mathfrak R_A(n_t)+T_A(n_s)+J^{-2}
 =O(n_t^{-a}\log^b n_t)\), where \(a>0\) and \(b\ge0\), and
\(\delta_A^2\asymp n_t^{-\gamma}\) with \(0\le\gamma<a\),
then \cref{eq:lambda-choice} and the mean-defect condition give the
strict advantage in \cref{cor:strict-adaptation}. A fixed nonzero
correctable mismatch corresponds to \(\gamma=0\).

\paragraph{A completely verified conditional-flow example.}
In \cref{prop:adaptation-example}, the source response is \(V/4\), the
target response is \(V/4+\delta f(W)\), and \(V\) is independent of
the \(q_A\)-dimensional vector \(W\). A valid neural source minimizer
has representation \(\widehat c V\), with
\(1/4\le\widehat c\le1/2\). The target residual learns
\(4\widehat c\delta f(W)\). 
For
\(q_A=2\), \(\beta_A=2\), \(r_t=4\), and \(D_t=3\), the adaptive
upper rate with sufficiently accurate integration is
\((\log^3(en_t)/n_t)^{2/3}\), while every frozen generator has a
nonvanishing error for fixed \(\delta>0\). A shrinking mismatch gives a
polynomial-rate separation as well: taking
\(\delta=\delta_0n_t^{-1/4}\), \(0<\delta_0\le1/8\), and
\(J^{-2}\lesssim(\ell_{n_t}^3/n_t)^{2/3}\) gives frozen risk
\(\asymp n_t^{-1/2}\), whereas the adaptive risk is
\(O(n_t^{-2/3}\ell_{n_t}^2)\). This bounded point-mass conditional-law
example shows nonvacuity of the separation conditions.

\subsection{Target-only estimation and the transfer comparison}
A non-transfer estimator uses only \(\sD_t\) and its auxiliary draws.
Set \(M_t=K_{\beta_t,p}(n_t)\), use the same target head resolution
\(N_t=K_{r_t,D_t}(n_t)\), and define
\begin{equation}\label{eq:nt-estimator}
 \widehat\cL_t(\widehat v_{\rm NT},\widehat h_{\rm NT})
 \le\inf_{v\in\cV_{t,N_t},\,h\in\cH_{M_t}}
       \widehat\cL_t(v,h)+n_t^{-2},\qquad
 \widehat F_{\rm NT}=F_{t,\widehat v_{\rm NT},\widehat h_{\rm NT}}.
\end{equation}
This is a separately trained target-only estimator. It does not use a
source initialization, a source representation, or a source-based anchor.

\begin{theorem}[End-to-end neural non-transfer counterpart]\label{thm:nt}
Suppose the target part of \cref{ass:path,ass:smooth} holds. The estimator
in \cref{eq:nt-estimator} satisfies
\begin{align}
 \E\cE_t(\widehat F_{\rm NT})
 &\le C R_{\rm NT}(n_t),\label{eq:nt-risk}\\
 R_{\rm NT}(n_t)
 &:=\zeta_t^2+\rho_{r_t,D_t}(n_t)+\rho_{\beta_t,p}(n_t),
       \label{eq:nt-rate-symbol}\\
 \E\cW_t^2(P_{\widehat F_{\rm NT}}^{[J]},P_t^0)
 &\le C\{R_{\rm NT}(n_t)+J^{-2}\}.
       \label{eq:nt-main}
\end{align}
For exact integration, omit \(J^{-2}\). 
\end{theorem}
\resultnote{This is the neural non-transfer counterpart of the two
transfer theorems. Unlike a frozen fit, joint target fitting pays for both
the target head and target representation. Its proof uses a joint neural
cover and a simultaneous head--encoder comparator. It is also the
source-independent baseline for the validation theorem.}{thm:nt}{app:mainproofs}

The following comparison uses the common correctability conditions of
\cref{thm:anchor}; the target-only row uses its own smooth-comparator
conditions. Fixed constants and the common Euler term are suppressed.
\begin{center}
\begin{tabular}{@{}>{\raggedright\arraybackslash}p{0.23\linewidth}>{\raggedright\arraybackslash}p{0.28\linewidth}>{\raggedright\arraybackslash}p{0.42\linewidth}@{}}
\toprule
Estimator & Target-head contribution & Other upper-bound contributions\\
\midrule
TFM-Freeze & \(\rho_{r_t,D_t}(n_t)\) &
 \(T_A(n_s)+L_{u,t}^2\delta_A^2\)\\[5pt]
TFM-Adapt  & \(\rho_{r_t,D_t}(n_t)\) &
 \(T_A(n_s)+\rho_{\beta_A,q_A}(n_t)\)\newline
 \(\quad+\lambda\delta_A^2\)\\[5pt]
Target-only & \(\rho_{r_t,D_t}(n_t)\) &
 \(\rho_{\beta_t,p}(n_t)+\zeta_t^2\)\\
\bottomrule
\end{tabular}
\end{center}
Theorem~\ref{thm:freeze} additionally supplies the frozen bound with
\(T_s(n_s)\) under its calibrated reduced-mean conditions. These are
alternative sufficient bounds; their assumptions are not interchangeable.

For target-only comparison in an exact target model, a sufficient
condition for a smaller adaptation-specific upper contribution is
\begin{equation}\label{eq:benefit}
 T_A(n_s)+\rho_{\beta_A,q_A}(n_t)+\lambda\delta_A^2
          =o\{\rho_{\beta_t,p}(n_t)\}.
\end{equation}
For example, when \(\beta_A=\beta_t\) and \(q_A<p\), the residual-learning term
has a strictly larger power exponent than the generic target
representation-learning term. The source contribution and regularization
bias must still be small. 
Under the frozen theorem's conditions,
the analogous sufficient comparison is
\(T_s(n_s)=o\{\rho_{\beta_t,p}(n_t)\}\).

\section{Endpoint validation and target-only safeguards}
\label{sec:endpoint-safeguards}
In this section, we present the procedure to prevent negative transfer.
Before fitting candidates, split the target sample into independent parts
\(\sD_{\rm tr}\) and \(\sD_{\rm val}\), of sizes
\(n_{\rm tr}\) and \(n_{\rm val}\). Train candidates with
\(n_t=n_{\rm tr}\), including architecture and tolerance choices.
The validation observations and simulation draws are excluded from
representation learning, anchor construction, fitting, and dictionary
construction.

Condition on the pre-validation information \(\cS_{\rm tr}\), fixing
\(\mathscr F=(F_1,\ldots,F_M)\) with deterministic \(M\ge2\).
Include the target-only candidate \(F_{\rm NT}\) trained on
\(n_{\rm tr}\) observations. Other entries may be frozen fits, adaptive
fits at predeclared nonnegative penalties \(\lambda\), possibly including
zero, or fits using different declared architectures.
All fields satisfy the common target envelopes and state/time Lipschitz
bounds in \cref{eq:lipschitz}.

\subsection{Endpoint laws, conditional scores, and the validation experiment}
\label{sec:endpoint-validation}
We validate the deployed endpoint laws.
Throughout the rest of this section write
\(n=n_{\rm val}\), \(d=d_t\), and
\begin{equation}\label{eq:endpoint-dictionary}
 P_z=P_t^0(\cdot\mid z),\qquad
 Q_{m,z}=P_{F_m}^{[J_m]}(\cdot\mid z),\quad 1\le m\le M.
\end{equation}
Exact integration is also allowed. Each solver and its resolution \(J_m\)
are fixed as part of the pre-validation candidate. The notation
\(Q_m=(Q_{m,z})_z\) denotes a probability kernel, and
\(m_0\) indexes the included target-only kernel \(Q_{\rm NT}=Q_{m_0}\).
Conditionally on \(\cS_{\rm tr}\), these kernels are fixed. Endpoint
validation uses the original pairs
\((X_i,Z_i)\stackrel{\rm iid}{\sim}P_t^0\).
Generated endpoints used to approximate a score are auxiliary randomness;
they do not increase the number \(n\) of target observations.

\begin{assumption}[Common bounded endpoint support]\label{ass:endpoint-support}
There is a fixed compact set \(\mathcal K_{\rm e}\subset\R^d\) of
Euclidean diameter at most \(D_{\rm e}>0\) such that, for target-almost
every \(z\), \(P_z\) and every \(Q_{m,z}\) are supported on
\(\mathcal K_{\rm e}\). All kernels and sampling maps are measurable.
The validation sample is independent of \(\cS_{\rm tr}\).
\end{assumption}
This is sufficient for the distribution-free bounded-score statements
below; it does not require calibration, correctability, or a correct
flow-matching model. It is  available in the paper's 
regime. Indeed, absolute continuity of the interpolation paths and
\cref{ass:path} imply that \(G_t,X_t\) lie in
\([-K_{x,t},K_{x,t}]^d\) almost surely. A field bounded by \(B_{V,t}\)
moves its initial point by at most \(B_{V,t}\), for both exact and Euler
integration. Thus one may take
\(\mathcal K_{\rm e}=[-K_{x,t}-B_{V,t},K_{x,t}+B_{V,t}]^d\).
\Cref{lem:endpoint-support} supplies the endpoint and solver details.

Fix an energy exponent \(\upsilon\in(0,2)\), usually \(\upsilon=1\),
and set \(A_{\rm e}=D_{\rm e}^{\upsilon}\). Define the negatively
oriented energy score and energy discrepancy by
\begin{align}
 S_{\upsilon}(Q,x)
  &=\E_{Y\sim Q}\|Y-x\|^{\upsilon}
       -\tfrac12\E_{Y,Y'\stackrel{\rm iid}{\sim}Q}\|Y-Y'\|^{\upsilon},
          \label{eq:energy-score}\\
 \mathsf D_{\upsilon}(P,Q)
  &=2\E\|X-Y\|^{\upsilon}-\E\|X-X'\|^{\upsilon}
                            -\E\|Y-Y'\|^{\upsilon},
          \label{eq:energy-discrepancy}\\
 \mathcal R_{{\rm E},\upsilon}(Q)
  &=\tfrac12\E_{Z_t}\mathsf D_{\upsilon}(P_{Z_t},Q_{Z_t}).
          \label{eq:endpoint-risks}
\end{align}
The four variables in \cref{eq:energy-discrepancy} are independent, with
\(X,X'\sim P\) and \(Y,Y'\sim Q\). Lower scores are better.
We use \(\mathsf D_{\upsilon}\) for the conventional energy discrepancy.
Most importantly,
\begin{equation}\label{eq:conditional-energy-identity}
 \E S_{\upsilon}(Q_{Z_t},X_t)
 -\E S_{\upsilon}(P_{Z_t},X_t)
       =\mathcal R_{{\rm E},\upsilon}(Q)\ge0,
\end{equation}
with equality exactly when \(Q_z=P_z\) for target-almost every \(z\).
The unknown second term is common to all candidates and need not be
estimated. Thus one observed endpoint per context suffices for conditional
energy-score validation, even when the contexts are continuously distributed.
The underlying proper-score and Hilbert-space identities are classical
\citep{GneitingRaftery2007,Sejdinovic2013}; \cref{lem:energy-hilbert}
gives a self-contained proof with the normalization used here.

\subsection{Energy-score selection and   target-only safeguard}
\label{sec:energy-selection}
At validation context \(Z_i\), simulate \(B\ge2\) endpoints for
each candidate \(m\). Conditionally on the validation pair and fitted
dictionary, these draws are independent with
\(Y_{m,i,1},\ldots,Y_{m,i,B}\sim Q_{m,Z_i}\). Use the unbiased score
\begin{equation}\label{eq:energy-mc-score}
 \widehat S_{m,i}
 =\frac1B\sum_{b=1}^B\|Y_{m,i,b}-X_i\|^{\upsilon}
 -\frac1{2B(B-1)}\sum_{b\ne b'}
                       \|Y_{m,i,b}-Y_{m,i,b'}\|^{\upsilon}.
\end{equation}
Simulation rows are independent across \(i\).
Set
\begin{equation}\label{eq:energy-selection-rule}
 \widehat s_m=\frac1n\sum_{i=1}^n\widehat S_{m,i},\quad
 \widehat m_{\rm E}\in\argmin_m\widehat s_m,\qquad
 \varepsilon_{\rm E}(\delta)
 =c_{\rm E}A_{\rm e}\sqrt{\frac{\log(2M/\delta)}{2n}},
\end{equation}
where \(c_{\rm E}=1\) for exact scores and \(c_{\rm E}=3/2\) for
\cref{eq:energy-mc-score}. 
For a predeclared confidence level \(\delta\in(0,1)\), define
\begin{equation}\label{eq:energy-safe-rule}
 \widehat m_{{\rm E},{\rm safe}}=
 \begin{cases}
 \widehat m_{\rm E},&
 \widehat s_{m_0}-\widehat s_{\widehat m_{\rm E}}
                                      >2\varepsilon_{\rm E}(\delta),\\
 m_0,&\text{otherwise}.
 \end{cases}
\end{equation}
This rule switches away from target-only only when the improvement exceeds
a simultaneous confidence margin.

\begin{theorem}[Conditional energy-score selection and safe switching]
\label{thm:energy-selection}
Under \cref{ass:endpoint-support} and the simulation scheme above,
conditionally on \(\cS_{\rm tr}\), with probability at least
\(1-\delta\) over validation and simulation, all of the following hold:
\begin{align}
 \mathcal R_{{\rm E},\upsilon}(Q_{\widehat m_{\rm E}})
 &\le\min_m\mathcal R_{{\rm E},\upsilon}(Q_m)
                                  +2\varepsilon_{\rm E}(\delta),
                    \label{eq:energy-selection-oracle}\\
 \mathcal R_{{\rm E},\upsilon}(Q_{\widehat m_{{\rm E},{\rm safe}}})
 &\le\mathcal R_{{\rm E},\upsilon}(Q_{\rm NT}),
                    \label{eq:energy-safe-baseline}\\
 \mathcal R_{{\rm E},\upsilon}(Q_{\widehat m_{{\rm E},{\rm safe}}})
 &\le\min_m\mathcal R_{{\rm E},\upsilon}(Q_m)
                                  +4\varepsilon_{\rm E}(\delta).
                    \label{eq:energy-safe-oracle}
\end{align}
Moreover,
\begin{align}
 \E_{{\rm val},{\rm sim}}[
    \mathcal R_{{\rm E},\upsilon}(Q_{\widehat m_{\rm E}})
                       \mid\cS_{\rm tr}]
 &\le\min_m\mathcal R_{{\rm E},\upsilon}(Q_m)
       +2c_{\rm E}A_{\rm e}\sqrt{\frac{\log(2M)}{2n}},
                    \label{eq:energy-selection-expectation}\\
 \E_{{\rm val},{\rm sim}}[
    \mathcal R_{{\rm E},\upsilon}(Q_{\widehat m_{{\rm E},{\rm safe}}})
                       \mid\cS_{\rm tr}]
 &\le\mathcal R_{{\rm E},\upsilon}(Q_{\rm NT})+A_{\rm e}\delta.
                    \label{eq:energy-safe-expectation}
\end{align}
On the same high-probability event, if the best candidate improves on
\(Q_{\rm NT}\) in this risk by more than
\(4\varepsilon_{\rm E}(\delta)\), the safe rule switches.
\end{theorem}
\resultnote{Note that this is a leading-constant-one oracle for a 
deployed endpoint generator. The conservative version has no deterioration
relative to the actual target-only endpoint law on its confidence event,
in the declared energy risk. Fixed \(B\ge2\) suffices for consistency of
finite-dictionary selection: the Monte Carlo score is unbiased and its
rows are independent. No target conditional-density estimator is required.}
{thm:energy-selection}{app:endpoint}

\section{Simulation studies}\label{sec:numerical-simulations}
We investigate the accuracy gained by adapting a source representation and
the ability of target validation to guard against negative transfer. The
design varies the source sample size and the discrepancy between the
source and target representations, while keeping the target sample size
fixed. It includes nonlinear sufficient representations, heteroscedastic
responses, and a target conditional distribution with two mixture
components.

\subsection{Data-generating process}\label{sec:numerical-generator}
In each domain, let $Z\sim N_p(0,I_p)$, with independent observations
across and within domains. Fix $p\times4$ matrices $B$ and $C$ such that
$[B,C]$ has orthonormal columns, and set
$B_s=B$ and $B_t=\cos(\theta)B+\sin(\theta)C$.
For $u=B_j^\top z$, $j\in\{s,t\}$, define
\begin{equation}\label{eq:num-representation}
 q(v)=\frac{v^2-1}{\sqrt2},\qquad
 h_j(z)_\ell=2.5\tanh\!\left\{
 \frac{u_\ell+q(u_{\ell+2})}{2.5\sqrt2}\right\},\quad \ell=1,2.
\end{equation}
We consider $\theta=0.5$ and $\theta=\pi/2$. The first setting has
related representations. In the second, the source and target depend
on independent Gaussian projections, so that $h_s(Z)$ and $X_t$ are
independent under the target law.

Let $d=d_s=d_t$ be the response dimension. Write
$\xi_j=h_j(z)$, $a_k=0.8+0.2\cos k$,
$b_k=0.6\sin(k+0.5)$, and $s(v)=(1+e^{-v})^{-1}$.
For $k=1,\ldots,d$, the source response is
\begin{align}
 X_{s,k}&=\mu_{s,k}+\sigma_{s,k}\epsilon_{s,k},
                  \label{eq:num-source}\\
 \mu_{s,k}&=1.3a_k\xi_{s,1}+b_k\xi_{s,2}
                  +0.3\sin(\xi_{s,1}+\xi_{s,2}+k),\nonumber\\
 \sigma_{s,k}&=0.55+0.12s(\xi_{s,2}+\cos k).\nonumber
\end{align}
For the target response, define
\begin{align}
 \mu_{t,k}&=a_k\{0.8\xi_{t,1}+0.6\sin\xi_{t,1}\}+b_k\xi_{t,2}
                       \label{eq:num-target-parameters}\\
 &\quad+0.35\sin(\xi_{t,1}\xi_{t,2}/2+k),\nonumber\\
 \pi_t&=0.15+0.7s(0.9\xi_{t,1}-0.6\xi_{t,2}),\nonumber\\
 \delta_k&=0.9\cos(0.7k)+0.7,\qquad
 \sigma_{t,k}=0.3+0.25s(\xi_{t,2}+0.3\sin k).\nonumber
\end{align}
Draw $L\mid Z\sim\operatorname{Bernoulli}(\pi_t)$ and set
\begin{equation}\label{eq:num-target}
 X_{t,k}=\mu_{t,k}+(2L-1)\delta_k
                    +\sigma_{t,k}(1+0.25L)\epsilon_{t,k}.
\end{equation}
All errors $\epsilon_{j,k}$ are independent standard normal variables,
independent of the covariates and mixture labels. The label $L$ is shared
across target response coordinates and induces conditional dependence
when $d>1$. Thus the source and target response laws differ even when
their representations are closely related. The projection matrices,
representations, and mixture parameters are not supplied to the fitted
procedures.

The dimension pairs are $(p,d)=(50,1)$, $(200,5)$, and $(500,10)$.
We fix $n_t=1024$ and take $n_s=1000$, $2000$, $4000$, or $8000$.
Each setting has 100 independent
training repetitions. Within a repetition, the methods share the same
observations and evaluation samples.

\subsection{Methods and evaluation}
We compare target-only diffusion, target-only flow matching (FM), the
frozen-representation diffusion construction of \citet{TianShen2026},
TFM-Freeze, TFM-Adapt, and TFM-Adapt with $\lambda=0$.
The positive-penalty procedure
selects $\lambda\in\{0.01,0.1,1\}$ by tuning; the unpenalized procedure
fixes $\lambda=0$.
We also apply a target-validation safeguard to each of the three TFM
procedures, giving nine methods in total. These versions are denoted by
$+\mathrm{TV}$. They compare a tuned transfer candidate with target-only
FM on separate target observations and select the target-only fit when
the validation evidence does not favor transfer. The common neural
architecture, tuning protocol, and validation rule
are given in \cref{app:numerical-training}.
Accuracy is measured by the context-averaged energy discrepancy and
joint empirical Wasserstein distances $W_1$ and $W_2$.
Lower values indicate more accurate
generation. Evaluation details and uncertainty calculations are given in
\cref{app:numerical-evaluation}.

\subsection{Generation accuracy and negative transfer}
\begin{figure}[p]\centering
\includegraphics[width=\textwidth,height=.82\textheight,keepaspectratio]{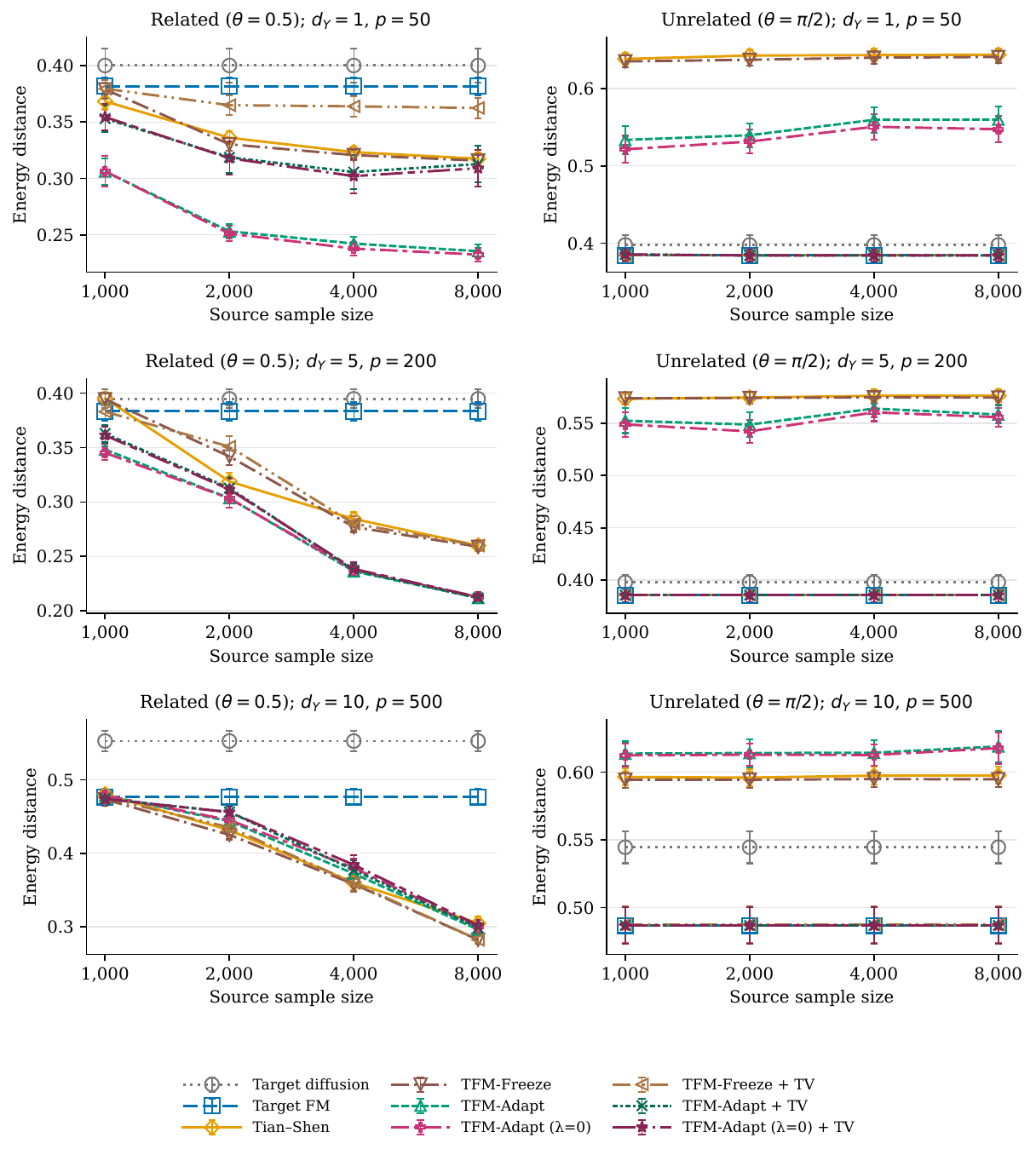}
\caption{Mean energy discrepancy against source sample size, with
$n_t=1024$. Columns correspond to related and unrelated representations;
rows correspond to $(p,d)=(50,1),(200,5),(500,10)$. Error bars are
pointwise 95\% Monte Carlo intervals from 100 repetitions.
$+\mathrm{TV}$ denotes target validation. 
}
\label{fig:num-energy}
\end{figure}

\Cref{fig:num-energy} shows substantial gains from adaptation in the
related setting. For $d=1$ and $d=5$, both adaptive procedures have
smaller mean energy discrepancy than the frozen and target-only
alternatives throughout the source-size grid. At $n_s=8000$, the
unpenalized procedure attains 0.2325 for $d=1$, compared
with 0.3158 for TFM-Freeze, the strongest nonadaptive
comparator, a reduction of 26.4\%. For $d=5$, TFM-Adapt
reduces the discrepancy from 0.2589 to 0.2118,
an improvement of 18.2\%. Paired comparisons support both adaptive
procedures over all five nonadaptive comparators in these two dimensions
(\cref{app:numerical-comparisons}).

\begin{figure}[p]\centering
\includegraphics[width=\textwidth,height=.82\textheight,keepaspectratio]{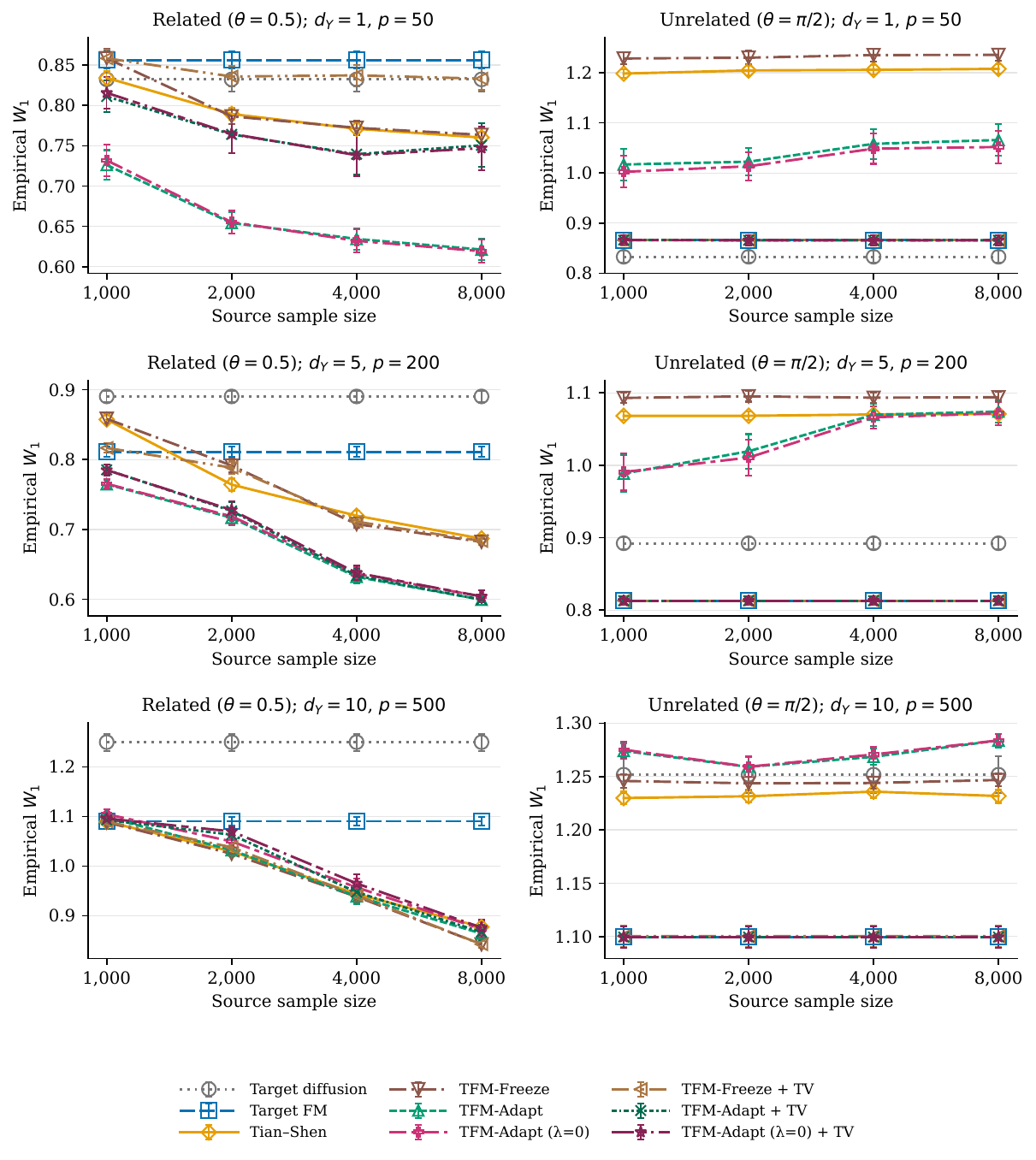}
\caption{Mean joint empirical Wasserstein-1 distance against source
sample size, with $n_t=1024$. Panel arrangement, methods, and intervals
are as in \cref{fig:num-energy}. 
}
\label{fig:num-w1}
\end{figure}

\begin{figure}[p]\centering
\includegraphics[width=\textwidth,height=.82\textheight,keepaspectratio]{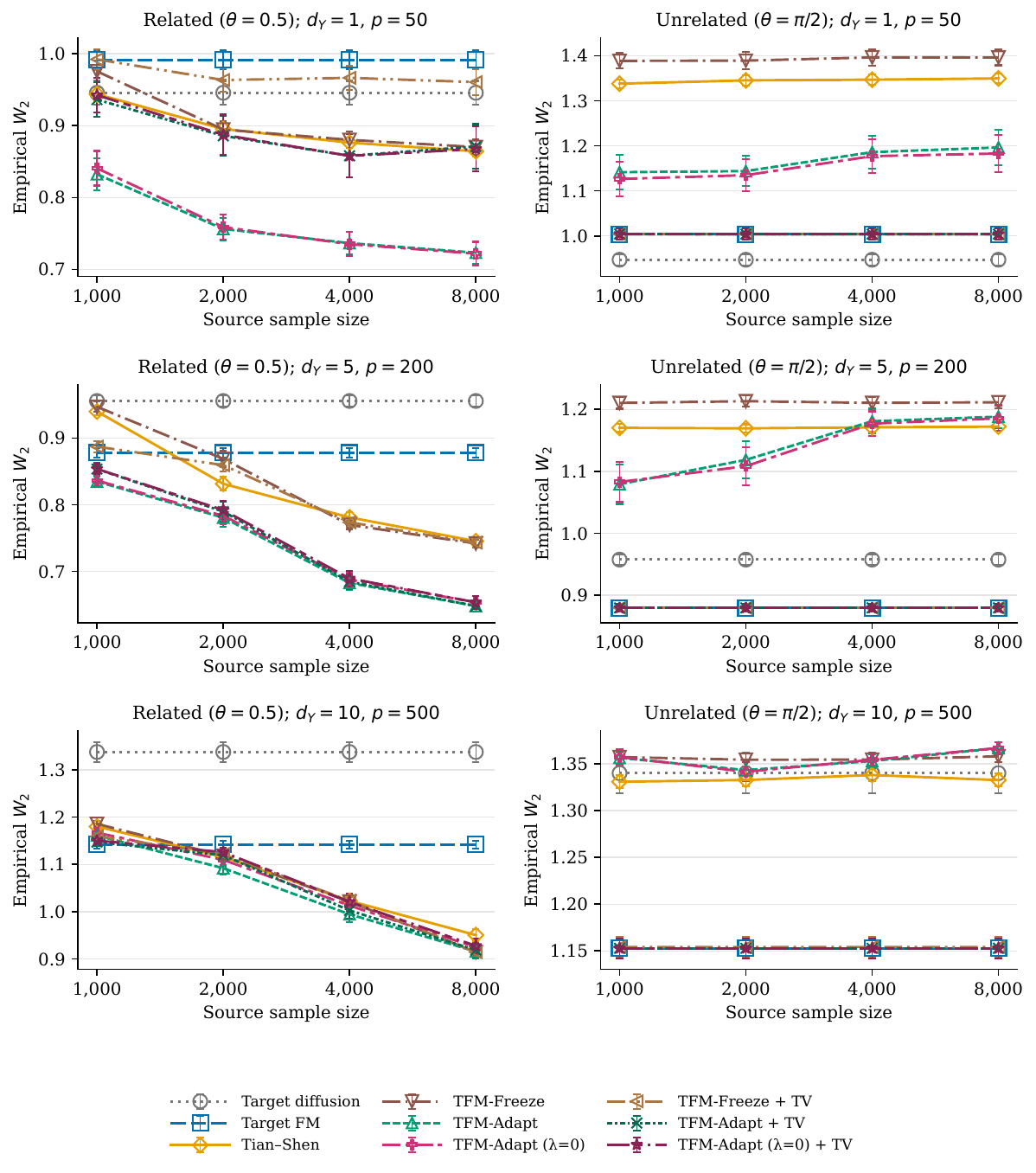}
\caption{Mean joint empirical Wasserstein-2 distance against source
sample size, with $n_t=1024$. Panel arrangement, methods, and intervals
are as in \cref{fig:num-energy}.
}
\label{fig:num-w2}
\end{figure}

\Cref{fig:num-w1,fig:num-w2} corroborate the adaptation gains for
$d=1$ and $d=5$ under both Wasserstein criteria. The ten-dimensional
setting also benefits from transfer: TFM-Adapt reduces energy discrepancy by
37.9\% relative to target-only FM at $n_s=8000$,
with TFM-Freeze attaining the smallest discrepancy in that setting.
The similar performance of positive and zero penalties in the first two
dimensions indicates that allowing the representation to adapt is an
important source of the observed gain.

The right panels examine unrelated representations. With target
validation, each adaptive procedure has at most one harmful repetition
out of 100 in every setting, and none at $n_s=8000$. At that source
size, both recover target-only FM exactly in all three dimensions.
For related five-dimensional responses at the same source size,
validation retains both adaptive fits in all 100 repetitions.
Thus the safeguard preserves the accuracy gain when transfer is useful
in this setting and returns to target-only learning when the
representations are unrelated. Counts use energy discrepancy;
\cref{tab:num-validation} gives the comparisons with and without validation.

\clearpage

\section{Real-data applications}\label{sec:numerical-real}
We consider Materials and Diamonds, two conditional distribution
prediction problems with different source and target populations.
The same nine procedures are compared.
 Each application uses 40 paired training repetitions.
We assess prediction by the continuous ranked probability score
(CRPS) on held-out target observations; lower values are better.
Here, CRPS is the energy score with exponent one
\citep{GneitingRaftery2007}. Data construction and uncertainty
calculations are described in \cref{app:numerical-data,app:numerical-evaluation}.

\subsection{Superconducting materials}
The Materials application uses the UCI superconductivity data
\citep{Hamidieh2018}. The response is $\log(1+T_c)$, where $T_c$ is
the critical temperature in kelvin, and the predictors are 81 chemical
descriptors. Copper-oxide compounds without iron form the source
domain; iron-bearing compounds without copper form the target domain.
Each repetition uses $n_s=6000$ source and $n_t=256$ target
compositions, with 336 target compositions reserved for testing.

This application illustrates the protection provided by target
validation. Both adaptive procedures with validation attain mean CRPS
0.31826, equal to target-only FM, with zero harmful
repetitions out of 40 (\cref{tab:num-real}). In every repetition, the
safeguard selects the target-only fit. It thereby avoids the loss
incurred by transferring between these populations: without validation,
TFM-Adapt and its $\lambda=0$ version have mean CRPS
0.35067 and 0.34833. The gain here is
preservation of target-only FM performance through a data-dependent
decision about whether to transfer.

\subsection{Diamond prices}
The Diamonds application predicts log price from 19 variables describing
carat, depth and table percentages, physical dimensions, color, and
clarity \citep{DiamondsData}. Premium and Very Good cuts form the
source domain, and Ideal cuts form the target domain. Cut grade is
excluded from the predictors. Each repetition uses $n_s=12000$
source and $n_t=2048$ target groups, with 1024 target groups reserved
for testing.

\Cref{tab:num-real} shows that adaptation improves prediction in this
application. TFM-Adapt attains the lowest mean CRPS, 0.06357,
followed closely by its $\lambda=0$ version at 0.06365.
These values improve on target-only FM by 8.05\% and
7.95\%, respectively, and on TFM-Freeze by
3.65\% and 3.54\%.
Pointwise 95\% paired Student-$t$ intervals favor both adaptive procedures
over each of the five nonadaptive comparators
(\cref{tab:num-real-comparisons}). TFM-Adapt improves on target-only FM
in all 40 repetitions. Its validated version retains transfer in
39 repetitions, has zero harmful repetitions, and attains CRPS
0.06371. Thus target validation preserves most of the
adaptation gain on Diamonds while recovering target-only performance
on Materials.

\begin{table}[htbp]\centering
\caption{Mean test CRPS and harmful-repetition counts for Materials and Diamonds, using 40 paired repetitions.
Harm compares TFM procedures with target-only FM and Tian--Shen with target-only diffusion; exact fallback ties are excluded. Bold type marks the lowest mean CRPS in each application.}
\label{tab:num-real}
\setlength{\tabcolsep}{5pt}
\begin{tabular}{lrrrr}
\toprule
 & \multicolumn{2}{c}{Materials} & \multicolumn{2}{c}{Diamonds} \\
\cmidrule(lr){2-3}\cmidrule(lr){4-5}
Method & CRPS & Harm & CRPS & Harm \\
\midrule
Target-only diffusion & \textbf{0.31245} & --- & 0.07064 & --- \\
Target-only FM & 0.31826 & --- & 0.06914 & --- \\
Tian--Shen & 0.41195 & 40/40 & 0.06959 & 11/40 \\
TFM-Freeze & 0.41191 & 40/40 & 0.06598 & 2/40 \\
TFM-Adapt & 0.35067 & 35/40 & \textbf{0.06357} & 0/40 \\
TFM-Adapt ($\lambda=0$) & 0.34833 & 35/40 & 0.06365 & 1/40 \\
TFM-Freeze + TV & 0.31826 & 0/40 & 0.06606 & 1/40 \\
TFM-Adapt + TV & 0.31826 & 0/40 & 0.06371 & 0/40 \\
TFM-Adapt ($\lambda=0$) + TV & 0.31826 & 0/40 & 0.06386 & 1/40 \\
\bottomrule
\end{tabular}
\end{table}

\clearpage
\appendix

\section{Projection identities and flow stability}\label{app:structural}
\subsection{Conditional regression and the probability path}
\begin{lemma}[Orthogonal decomposition and deficiency]\label{lem:projection}
Under the moment condition in \cref{eq:path}, every square-integrable
field \(F=F_{j,v,h}\) satisfies \cref{eq:decomposition} and
\(\cL_j(v,h)-\cL_j^*=\cE_j(F)\). Moreover,
\begin{equation}\label{eq:deficiency-bound}
 0\le\cD_j(h)\le\|v_j^0\|_j^2\le\E\|U_{j,T}\|^2.
\end{equation}
If \(P_j^0(\cdot\mid Z_j)=P_j^0(\cdot\mid H_{j,h}(Z_j))\), then
\(\cD_j(h)=0\) for the independent-base construction in \cref{eq:path}.
\end{lemma}
\begin{proof}\proofanchor{lem:projection}
Fix \(j,h\), and abbreviate \(U=U_{j,T}\), \(W=W_j\),
\(v^0=\E(U\mid W)\), and
\(m=\E(U\mid S_{j,T},H_{j,h}(Z_j),T)\).
Since \(F-v^0\) is measurable with respect to \(W\),
\[
 \E\langle F-v^0,v^0-U\rangle
 =\E\langle F-v^0,\E(v^0-U\mid W)\rangle=0.
\]
Expanding \(F-U=(F-v^0)+(v^0-U)\) proves the excess-risk identity.
Let \(\mathcal B_h=\sigma(S_{j,T},H_{j,h}(Z_j),T)\).
The tower property gives \(\E(v^0\mid\mathcal B_h)=m\), so
\[
 \E\langle F-m,m-v^0\rangle
 =\E\langle F-m,\E(m-v^0\mid\mathcal B_h)\rangle=0.
\]
Expansion of \(F-v^0=(F-m)+(m-v^0)\) proves the first identity in
\cref{eq:decomposition}. Taking the infimum over \(v\) proves the
second, without assuming that the infimum is attained. Orthogonal
projection also gives
\(\cD_j(h)=\|v^0\|_j^2-\|m\|_j^2\), and conditional Jensen gives
\(\|v^0\|_j^2\le\E\|U\|^2\).

For the final assertion, sufficiency and independence of \(G_j,T\)
imply that the conditional law of \((X_j,G_j,T)\) given \(Z_j\)
depends on \(Z_j\) only through \(H_{j,h}(Z_j)\).
The state and label are measurable functions of these variables.
Consequently their conditional joint law has the same property, and
\(\E(U\mid S_{j,T},Z_j,T)
 =\E(U\mid S_{j,T},H_{j,h}(Z_j),T)\) almost surely.
Thus \(v^0=m\) in \(L_2(\nu_j)\).
\end{proof}

\begin{lemma}[Conditional marginal-flow identity]\label{lem:marginal}
Under \cref{eq:path}, the conditional interpolation laws satisfy the
weak continuity equation with velocity \(v_j^0\) given \(Z_j\), and
with velocity \(m_{j,h}\) given the reduced context \(H_{j,h}(Z_j)\).
The characteristic realization in \cref{ass:path} identifies the full
conditional endpoint law with \(P_j^0(\cdot\mid Z_j)\).
A reduced-context characteristic has endpoint law
\(P_j^0(\cdot\mid H_{j,h}(Z_j))\) only when the analogous reduced-context
realization and uniqueness also hold.
\end{lemma}
\begin{proof}\proofanchor{lem:marginal}
Let \(\varphi\) be a continuously differentiable compactly supported
function on \(\R^{d_j}\), and let \(0\le a<b\le1\).
The chain rule along the almost surely absolutely continuous path gives
\[
 \varphi(S_{j,b})-\varphi(S_{j,a})
   =\int_a^b\nabla\varphi(S_{j,u})^\top U_{j,u}\dd u.
\]
The integrand is absolutely integrable in expectation, because
\(\|\nabla\varphi\|_\infty<\infty\) and
\(\E\int_0^1\|U_{j,u}\|\dd u<\infty\) by Cauchy--Schwarz.
Condition on \(Z_j\), use conditional Fubini, and condition the integrand
further on \(S_{j,u},Z_j\). For almost every \(u\), this replaces
\(U_{j,u}\) by \(v_j^0(S_{j,u},Z_j,u)\), yielding
\[
 \E\{\varphi(S_{j,b})-\varphi(S_{j,a})\mid Z_j\}
 =\int_a^b\E\{\nabla\varphi(S_{j,u})^\top
                  v_j^0(S_{j,u},Z_j,u)\mid Z_j\}\dd u.
\]
This is the integrated weak continuity equation. Conditioning instead on
\(H_{j,h}(Z_j)\) gives the reduced version with \(m_{j,h}\).
The endpoint identities follow from \(S_{j,0}=G_j\),
\(S_{j,1}=X_j\), and the respective characteristic realization
assumptions. 
\end{proof}

\subsection{Generation and numerical integration}
\begin{lemma}[Flow stability and an explicit Euler remainder]\label{lem:stability}
Suppose the target characteristic realization in \cref{ass:path} holds.
Let \(F\) be any field bounded by \(B_{V,t}\), globally state-Lipschitz
with constant \(L=L_{x,t}\), and time-Lipschitz with constant
\(L_{\tau,t}\). Its exact and numerical conditional endpoint laws satisfy
\begin{align}
 \cW_t^2(P_F,P_t^0)&\le C_L\cE_t(F),\label{eq:stability}\\
 C_L&=\int_0^1 e^{2L(1-u)}\dd u
       =\begin{cases}(e^{2L}-1)/(2L),&L>0,\\1,&L=0,
         \end{cases}\label{eq:stability-constant}\\
 \cW_t^2(P_F^{[J]},P_t^0)
       &\le2C_L\cE_t(F)+2C_{\rm Euler}J^{-2},\label{eq:numerical-stability}\\
 C_{\rm Euler}&=\frac{(L_{\tau,t}+LB_{V,t})^2}{4}
          \left(\frac{e^L-1}{L}\right)^2,\label{eq:euler-constant}
\end{align}
where the ratio in \cref{eq:euler-constant} equals one at \(L=0\).
The exact-flow bound does not require time Lipschitz continuity.
These bounds also hold after averaging over fitted-field randomness.
\end{lemma}
\begin{proof}\proofanchor{lem:stability}
Condition on training and on a target context \(z\). Couple the fitted
flow \(X_F\) and the true characteristic \(\Phi\) using the same initial
\(G_t\). The bounded, globally state-Lipschitz field has a unique global
solution, and \(\|X_F(\tau)\|\le\|G_t\|+B_{V,t}\).
For \(e(\tau)=\|X_F(\tau)-\Phi(\tau)\|\), the integral equation gives
\[
 e(\tau)\le L\int_0^\tau e(u)\dd u
  +\int_0^\tau\|F(\Phi(u),z,u)-v_t^0(\Phi(u),z,u)\|\dd u.
\]
The integral form of Gronwall's inequality, obtained by integrating the
corresponding scalar linear equation, yields
\[
 e(1)\le\int_0^1 e^{L(1-u)}
       \|F(\Phi(u),z,u)-v_t^0(\Phi(u),z,u)\|\dd u.
\]
Cauchy--Schwarz bounds its square by \(C_L\) times the integral of the
squared integrand without the exponential. Average over \(G_t,Z_t\).
The true characteristic has the same time marginals as the interpolation,
so that integral equals \(\cE_t(F)\). The coupled endpoints are admissible
for the Wasserstein infimum, proving \cref{eq:stability}.

For Euler integration let \(h_J=1/J\), \(t_k=kh_J\), and
\(x_{k+1}=x_k+h_JF(x_k,z,t_k)\), \(x_0=G_t\).
The exact solution satisfies
\[
 \begin{aligned}
 &\left\|X_F(t_{k+1})-X_F(t_k)-h_JF(X_F(t_k),z,t_k)\right\|\\
 &\quad\le\int_{t_k}^{t_{k+1}}(LB_{V,t}+L_{\tau,t})(u-t_k)\dd u
 =\tfrac12(LB_{V,t}+L_{\tau,t})h_J^2.
 \end{aligned}
\]
Writing \(e_k=\|x_k-X_F(t_k)\|\),
\[
 e_{k+1}\le(1+Lh_J)e_k+\tfrac12(LB_{V,t}+L_{\tau,t})h_J^2.
\]
Summing the geometric series and using \((1+L/J)^J\le e^L\) gives
\(e_J\le (LB_{V,t}+L_{\tau,t})(e^L-1)/(2JL)\), with its continuous
extension at \(L=0\). This proves the squared solver bound
\(\E\|x_J-X_F(1)\|^2\le C_{\rm Euler}J^{-2}\).
Couple all three endpoints using the same \(G_t\), and apply
\(\|x_J-\Phi(1)\|^2\le2\|x_J-X_F(1)\|^2+
2\|X_F(1)-\Phi(1)\|^2\). This proves
\cref{eq:numerical-stability}. Uniform constants permit averaging over
training randomness.
\end{proof}

\section{Constructive neural approximation and entropy}\label{app:neural}
\subsection{A multiplication module}
\begin{lemma}[Multiplication with a Lipschitz-small error]\label{lem:multiplier}
Fix \(A\ge1\). For every \(0<\delta<1/2\) there is a ReLU network
\(P_{\delta,A}:\R^2\to\R\) such that, on \([-A,A]^2\),
\begin{equation}\label{eq:multiplier}
 \|P_{\delta,A}(a,b)-ab\|_\infty\le\delta,
 \qquad \Lip(P_{\delta,A}(a,b)-ab)\le\delta.
\end{equation}
Moreover \(P_{\delta,A}(a,0)=P_{\delta,A}(0,b)=0\) exactly on this box.
Its width is bounded by a constant depending on \(A\), its depth and
number of nonzero parameters are at most \(C_A\log(1/\delta)\), and
its parameter magnitudes are bounded by a constant depending on \(A\).
\end{lemma}
\begin{proof}\proofanchor{lem:multiplier}
On \([0,1]\), let
\(T(u)=2\operatorname{ReLU}(u)-4\operatorname{ReLU}(u-1/2)
       +2\operatorname{ReLU}(u-1)\), and let \(T_j\) denote its
\(j\)-fold composition. The function
\[
 S_q(u)=u-\sum_{j=1}^q4^{-j}T_j(u)
\]
is the piecewise-linear interpolant of \(u^2\) on the mesh of spacing
\(2^{-q}\). To check this, start with the chord \(S_0(u)=u\).
On each old mesh interval the next tent is zero at both endpoints and
one at the midpoint, and subtracting \(4^{-(q+1)}\) gives the exact
midpoint value of the quadratic. Both sides are affine on the new
half intervals, completing the induction. On an interval \([a,a+h]\)
with \(h=2^{-q}\), the interpolation error is
\((u-a)(a+h-u)\). Consequently
\[
 \|S_q-u^2\|_\infty\le h^2/4,
 \qquad \Lip(S_q-u^2)\le h.
\]
Set \(A_*=2A\), and define
\(q_q(z)=A_*^2S_q(|z|/A_*)\) for \(|z|\le A_*\).
The function \(q_q\) is even, and its error from \(z^2\) has supremum
norm at most \(A_*^2h^2/4\) and Lipschitz constant at most \(A_*h\).
Polarization gives
\[
 P_{\delta,A}(a,b)=\tfrac14\{q_q(a+b)-q_q(a-b)\}.
\]
The value error is at most \(A_*^2h^2/8\). Since each of the linear maps
\((a,b)\mapsto a\pm b\) has Euclidean operator norm \(\sqrt2\), the
Lipschitz constant of the error is at most \(A_*h/\sqrt2\).
Choose \(q\) so that these two bounds are at most \(\delta\).
Evenness gives the exact zero assertions.

The recursion for \(T_j\), together with a running partial sum, uses
constant width and \(O(q)\) layers and parameters. Signed registers are
represented by their positive and negative parts if required by the ReLU
layers. Its coefficients, including \(4^{-j}\), are uniformly bounded;
the affine rescalings depend only on \(A\). As
\(q\le C_A\log(1/\delta)\), the architecture bounds follow.
\end{proof}

\subsection{Simultaneous approximation and actual Lipschitz budgets}
\begin{lemma}[Constructive admissible neural approximation]\label{lem:approximation}
Let \(K\subset\R^D\) be a fixed nondegenerate rectangle.
For every \(g\in C^a(K;B)\), \(a>0\), with a fixed output dimension,
there is a network in an architecture satisfying \cref{eq:architecture}
for which
\begin{equation}\label{eq:value-approximation}
 \|g-g_N\|_{\infty,K}\le C N^{-a/D}.
\end{equation}
If \(a>1\), it can be chosen so that
\begin{equation}\label{eq:derivative-approximation}
 \|Dg-Dg_N\|_{L_\infty(K)}\le C N^{-(a-1)/D},
\end{equation}
where the derivative of the ReLU network is taken almost everywhere and
the matrix norm is the operator norm. Before output clipping the
approximation has this derivative bound; after coordinatewise output
clipping, \cref{eq:value-approximation} and the resulting upper bounds
on each block Lipschitz constant remain valid whenever \(g\) takes values
inside the clipping box.

In particular, for the fixed positive margins in
\cref{eq:smooth-margins,eq:reduced-regularity}, sufficiently large
resolutions give comparators \(h_M\in\cH_M\) and
\(v_{j,N}\in\cV_{j,N}\) with
\begin{equation}\label{eq:admissible-approximation-proved}
 \|h_M-h_{j,0}\|_\infty\le C M^{-\beta_j/p},\qquad
 \|v_{j,N}-g_{j,0}\|_{\infty,\cK_j}\le C N^{-r_j/D_j}.
\end{equation}
The latter assertion holds uniformly for \(g_{t,h}\) in
\cref{ass:transfer}. All architecture and approximation constants are
uniform over the indicated H\"older balls.
\end{lemma}
\begin{proof}\proofanchor{lem:approximation}
Write \(m\) for the fixed output dimension of the function
\(g:K\to\R^m\). Throughout this proof, a constant \(C\) may depend
on \(D,m,a,B,K\), but not on the particular function \(g\) in the
H\"older ball \(C^a(K;B)\) or on the resolution \(N\). For a map
\(f:K\to\R^m\), the two norms used below are
\begin{equation}\label{eq:approx-proof-norms}
 \|f\|_{\infty,K}=\sup_{q\in K}\|f(q)\|,
 \qquad
 \|Df\|_{L_\infty(K)}
 =\mathop{\mathrm{ess\,sup}}_{q\in K}\|Df(q)\|_{\rm op}.
\end{equation}
The essential supremum in Equation~\eqref{eq:approx-proof-norms}
is with respect to \(D\)-dimensional Lebesgue measure; it ignores the
null set on which the derivative \(Df(q)\) of a ReLU realization
need not exist.

\medskip\noindent
\textbf{Step 1: Reduction of the rectangle \(K\) to the cube \(K_0\).}
Write
\(K=\prod_{d=1}^D[\ell_d,r_d]\), where \(r_d>\ell_d\), and define
\begin{equation}\label{eq:approx-affine-change}
 \begin{aligned}
 K_0&=[0,1]^D,&
 \Lambda_K&=\operatorname{diag}(r_1-\ell_1,\ldots,r_D-\ell_D),\\
 \Theta_K(x)&=\ell+\Lambda_Kx,&
 \bar g(x)&=g(\Theta_K(x)).
 \end{aligned}
\end{equation}
The diagonal matrix \(\Lambda_K\) is invertible because the rectangle
\(K\) is nondegenerate. The chain rule and the definition of
\(C^a(K;B)\) in Section~\ref{sec:assumptions} give
\(\bar g\in C^a(K_0;B')\), where \(B'\le C_KB\) and \(C_K\)
depends only on \(K,D,a\). If a ReLU realization \(\bar g_N\)
approximates \(\bar g\) on \(K_0\), define
\(g_N(q)=\bar g_N(\Theta_K^{-1}(q))\). Then
\begin{equation}\label{eq:approx-affine-errors}
 \begin{aligned}
 \|g_N-g\|_{\infty,K}
   &=\|\bar g_N-\bar g\|_{\infty,K_0},\\
 Dg_N(q)-Dg(q)
   &=\{D\bar g_N(x)-D\bar g(x)\}\Lambda_K^{-1},
     \quad x=\Theta_K^{-1}(q),
 \end{aligned}
\end{equation}
where the derivative identity holds almost everywhere when \(a>1\).
Consequently it suffices to construct \(\bar g_N\) on \(K_0\).
The fixed affine map \(\Theta_K^{-1}\) adds only fixed-dimensional
network operations.

\medskip\noindent
\textbf{Step 2: Hat functions \(\phi_{i,d}\) and the partition \(\phi_i\).}
Let \(Q\ge2\) be an integer mesh resolution, distinct from the network
resolution \(N\) and the representation resolution \(M\) in the
lemma. Define the index set \(\mathcal I_Q=\{0,\ldots,Q\}^D\),
the grid points \(x_i=i/Q\), and the functions
\begin{equation}\label{eq:approx-hats}
 \phi_{i,d}(u)=\{1-Q|u-x_{i,d}|\}_+,
 \qquad
 \phi_i(x)=\prod_{d=1}^D\phi_{i,d}(x_d),
 \qquad i\in\mathcal I_Q.
\end{equation}
Here \(s_+=\max(s,0)\). The scalar function \(\phi_{i,d}\) is a
one-dimensional hat factor: it equals one at \(x_{i,d}\), vanishes
when \(|u-x_{i,d}|\ge Q^{-1}\), and is affine on the intervals
\([x_{i,d}-Q^{-1},x_{i,d}]\) and
\([x_{i,d},x_{i,d}+Q^{-1}]\). The hat factors \(\phi_{i,d}\)
satisfy the bounds
\begin{equation}\label{eq:approx-hat-slopes}
 0\le\phi_{i,d}\le1,
 \qquad \Lip(\phi_{i,d};[0,1])\le Q,
 \qquad \phi_{i,d}'(u)\in\{-Q,0,Q\},
\end{equation}
where the derivative assertion holds at every differentiability point \(u\).
Thus the hat slopes in this construction have magnitude at most \(Q\).
Writing \(\varrho(s)=s_+\), an exact ReLU implementation of each
hat factor \(\phi_{i,d}\) is
\begin{equation}\label{eq:approx-hat-network}
 \phi_{i,d}(u)
 =\varrho(Qu-i_d+1)-2\varrho(Qu-i_d)
                         +\varrho(Qu-i_d-1).
\end{equation}

For \(u\in[j/Q,(j+1)/Q]\), the only potentially nonzero
one-dimensional hats have centers \(j/Q\) and \((j+1)/Q\);
their values are \(j+1-Qu\) and \(Qu-j\). Their sum is one.
Multiplying these one-dimensional identities over \(d=1,\ldots,D\)
proves
\begin{equation}\label{eq:approx-partition}
 \sum_{i\in\mathcal I_Q}\phi_i(x)=1,
 \qquad \phi_i(x)\ge0,
 \qquad x\in K_0.
\end{equation}
The support of \(\phi_i\), relative to \(K_0\), is contained in
\begin{equation}\label{eq:approx-local-box}
 B_i=K_0\cap\prod_{d=1}^D
              [x_{i,d}-Q^{-1},x_{i,d}+Q^{-1}].
\end{equation}
At every point \(x\in K_0\), at most \(2^D\) of the functions
\(\phi_i(x)\) are positive. Outside the finitely many mesh
hyperplanes \(\{x:x_d=j/Q\}\), at most \(2^D\) of the boxes
\(B_i\) contain \(x\). The product rule and
Equation~\eqref{eq:approx-hat-slopes} give, almost everywhere,
\begin{equation}\label{eq:approx-partition-derivative}
 \partial_d\phi_i(x)
   =\phi_{i,d}'(x_d)\prod_{e\ne d}\phi_{i,e}(x_e),
 \qquad
 \sum_{i\in\mathcal I_Q}\|D\phi_i(x)\|
      \le 2^D\sqrt D\,Q.
\end{equation}
Here \(D\phi_i(x)\) is a row vector and its norm is Euclidean.
Differentiating Equation~\eqref{eq:approx-partition} also gives
\(\sum_iD\phi_i(x)=0\) almost everywhere.

\medskip\noindent
\textbf{Step 3: Taylor polynomials \(T_i\) and the interpolant \(H_Q\).}
Set \(k=\lceil a\rceil-1\) and \(\alpha=a-k\in(0,1]\), as in
the definition of the H\"older ball \(C^a(K;B)\). For a multi-index
\(\nu=(\nu_1,\ldots,\nu_D)\) with nonnegative integer entries,
write \(|\nu|=\sum_d\nu_d\), \(\nu!=\prod_d\nu_d!\),
\(x^\nu=\prod_dx_d^{\nu_d}\), and
\(D^\nu=\partial_1^{\nu_1}\cdots\partial_D^{\nu_D}\).
Define
\begin{equation}\label{eq:approx-taylor-definitions}
 \begin{aligned}
 \mathcal M_k&=\{\nu:|\nu|\le k\},&
 c_{i,\nu}&=\frac{D^\nu\bar g(x_i)}{\nu!}\in\R^m,\\
 T_i(x)&=\sum_{\nu\in\mathcal M_k}c_{i,\nu}(x-x_i)^\nu,&
 H_Q(x)&=\sum_{i\in\mathcal I_Q}\phi_i(x)T_i(x).
 \end{aligned}
\end{equation}
The number of Taylor monomials is
\(J_k:=|\mathcal M_k|=\binom{D+k}{k}\), and
\(\|c_{i,\nu}\|\le B'\). When \(k=0\), the polynomial
\(T_i\) is the constant vector \(\bar g(x_i)\).

We first derive the Taylor remainders used in the approximation of
\(\bar g\). For \(k\ge1\), put \(w=x-x_i\) and
\(G_i(t)=\bar g(x_i+tw)\), \(0\le t\le1\).
The segment \(x_i+tw\) lies in the convex cube \(K_0\).
The one-dimensional integral Taylor formula, with the order-\(k\)
term subtracted explicitly, gives
\begin{equation}\label{eq:approx-integral-taylor}
 \bar g(x)-T_i(x)
 =\frac1{(k-1)!}\int_0^1(1-t)^{k-1}
                   \{G_i^{(k)}(t)-G_i^{(k)}(0)\}\dd t.
\end{equation}
If the segment \(x_i+tw\) meets \(\partial K_0\),
Equation~\eqref{eq:approx-integral-taylor} follows by approximating
both segment endpoints from \(\operatorname{int}K_0\) and using
continuity of the derivatives of \(\bar g\) up to the boundary.
The chain rule and the \(\alpha\)-H\"older condition yield
\begin{align}
 G_i^{(k)}(t)
   &=\sum_{|\nu|=k}\frac{k!}{\nu!}
                  D^\nu\bar g(x_i+tw)w^\nu,
                   \label{eq:approx-directional-taylor}\\
 \|G_i^{(k)}(t)-G_i^{(k)}(0)\|
   &\le C B't^\alpha\|w\|^{k+\alpha}.
                   \label{eq:approx-directional-remainder}
\end{align}
Substituting Equation~\eqref{eq:approx-directional-remainder} into
Equation~\eqref{eq:approx-integral-taylor} proves
\(\|\bar g(x)-T_i(x)\|\le C\|x-x_i\|^a\).
For \(k=0\), the same bound follows directly from the
\(a\)-H\"older condition on \(\bar g\).

When \(a>1\), let \(e_d\) be the \(d\)-th coordinate vector in
\(\R^D\). Differentiating the polynomial \(T_i\) gives
\begin{equation}\label{eq:approx-differentiated-taylor}
 \partial_dT_i(x)
 =\sum_{|\mu|\le k-1}
       \frac{D^{\mu+e_d}\bar g(x_i)}{\mu!}(x-x_i)^\mu.
\end{equation}
Thus \(\partial_dT_i\) is the order-\((k-1)\) Taylor polynomial
of \(\partial_d\bar g\). The function \(\partial_d\bar g\)
has smoothness \(a-1=(k-1)+\alpha\). Applying the remainder
argument just proved to \(\partial_d\bar g\), including its
order-zero case when \(k=1\), bounds each derivative column.
Because \(\|x-x_i\|\le\sqrt D/Q\) on \(B_i\), these bounds imply
\begin{equation}\label{eq:approx-local-taylor-errors}
 \begin{aligned}
 \sup_{x\in B_i}\|T_i(x)-\bar g(x)\|&\le C Q^{-a},\\
 \sup_{x\in B_i}\|DT_i(x)-D\bar g(x)\|_{\rm op}
      &\le C Q^{-(a-1)} \quad(a>1).
 \end{aligned}
\end{equation}
This argument also covers integer \(a\): in that case
\(\alpha=1\), and Equation~\eqref{eq:approx-directional-remainder}
uses the Lipschitz continuity of the order-\((a-1)\) derivatives.
It does not assume the existence of continuous order-\(a\) derivatives.

Equation~\eqref{eq:approx-partition} now gives the exact identities
\begin{equation}\label{eq:approx-interpolant-errors-identity}
 \begin{aligned}
 H_Q-\bar g
   &=\sum_i\phi_i(T_i-\bar g),\\
 D(H_Q-\bar g)
   &=\sum_i(T_i-\bar g)D\phi_i
                      +\sum_i\phi_i(DT_i-D\bar g)
                      \quad(a>1),
 \end{aligned}
\end{equation}
where the second identity holds almost everywhere. In the first sum
of the derivative identity, \((T_i-\bar g)D\phi_i\) is the outer
product of an \(m\)-dimensional column vector and a
\(D\)-dimensional row vector. Its operator norm is at most
\(\|T_i-\bar g\|\|D\phi_i\|\). Combining
Equations~\eqref{eq:approx-partition},
\eqref{eq:approx-partition-derivative}, and
\eqref{eq:approx-local-taylor-errors} therefore proves
\begin{equation}\label{eq:approx-interpolant-bounds}
 \begin{aligned}
 \|H_Q-\bar g\|_{\infty,K_0}&\le C Q^{-a},\\
 \|D(H_Q-\bar g)\|_{L_\infty(K_0)}
    &\le (C Q^{-a})(2^D\sqrt D\,Q)+C Q^{-(a-1)}
      \le C Q^{-(a-1)} \quad(a>1).
 \end{aligned}
\end{equation}

\medskip\noindent
\textbf{Step 4: ReLU products \(\widetilde p_{i,\nu}\) and their error recurrence.}
Each scalar localized monomial in
Equation~\eqref{eq:approx-taylor-definitions} is
\begin{equation}\label{eq:approx-local-monomial}
 p_{i,\nu}(x)=\phi_i(x)(x-x_i)^\nu
             =\prod_{r=1}^{s_{i,\nu}}z_r(x),
 \qquad s_{i,\nu}=D+|\nu|\le s_*:=D+k.
\end{equation}
For each fixed pair \((i,\nu)\), the first \(D\) factors in
Equation~\eqref{eq:approx-local-monomial} are
\(z_d(x)=\phi_{i,d}(x_d)\). The remaining factors are
\(x_d-x_{i,d}\), repeated \(\nu_d\) times for each \(d\).
By Equations~\eqref{eq:approx-hats} and
\eqref{eq:approx-hat-slopes}, every factor satisfies
\begin{equation}\label{eq:approx-factor-bounds}
 \|z_r\|_{\infty,K_0}\le1,
 \qquad \Lip(z_r;K_0)\le Q.
\end{equation}

Choose \(0<\delta\le(4s_*)^{-1}\). Apply
Lemma~\ref{lem:multiplier} with \(A=2\), and denote its error by
\(R_\delta(s,t)=P_{\delta,2}(s,t)-st\). Equation~\eqref{eq:multiplier}
states precisely that
\begin{equation}\label{eq:approx-multiplier-error}
 \sup_{(s,t)\in[-2,2]^2}|R_\delta(s,t)|\le\delta,
 \qquad \Lip(R_\delta;[-2,2]^2)\le\delta.
\end{equation}
For the factors associated with \((i,\nu)\), suppress the indices
\((i,\nu)\) temporarily and define the partial products
\begin{equation}\label{eq:approx-product-recursion}
 \begin{aligned}
 p^{(1)}&=\widetilde p^{(1)}=z_1,\\
 p^{(r)}&=p^{(r-1)}z_r,&
 \widetilde p^{(r)}&=P_{\delta,2}(\widetilde p^{(r-1)},z_r)
                 \quad(2\le r\le s_{i,\nu}),\\
 e_r&=\widetilde p^{(r)}-p^{(r)}.&
 \end{aligned}
\end{equation}
The final network product is
\(\widetilde p_{i,\nu}=\widetilde p^{(s_{i,\nu})}\).
If \(s_{i,\nu}=1\), this definition uses no multiplication module
and gives \(\widetilde p_{i,\nu}=p_{i,\nu}\) exactly.

We verify that every call to \(P_{\delta,2}\) remains in its
certified domain \([-2,2]^2\). Whenever the preceding calls are
valid, Equations~\eqref{eq:approx-product-recursion} and
\eqref{eq:approx-multiplier-error} give
\begin{equation}\label{eq:approx-product-error-identity}
 e_r=R_\delta(\widetilde p^{(r-1)},z_r)+e_{r-1}z_r,
 \qquad
 \|e_r\|_{\infty,K_0}\le\delta+\|e_{r-1}\|_{\infty,K_0}.
\end{equation}
Starting from \(e_1=0\), induction proves
\begin{equation}\label{eq:approx-product-value-control}
 \|e_r\|_{\infty,K_0}\le(r-1)\delta,
 \qquad
 \|\widetilde p^{(r)}\|_{\infty,K_0}
      \le1+(r-1)\delta\le\tfrac54.
\end{equation}
Indeed, the last bound at step \(r-1\), together with
\(\|z_r\|_\infty\le1\), first validates the next module input;
Equation~\eqref{eq:approx-product-error-identity} then proves the
bound at step \(r\). This establishes the induction without assuming
in advance that an approximate product lies in \([-2,2]\).

For scalar Lipschitz functions \(f_1,f_2\) on \(K_0\), adding
and subtracting \(f_1(x)f_2(y)\) gives
\begin{equation}\label{eq:approx-product-lipschitz-rule}
 \Lip(f_1f_2;K_0)
 \le\|f_1\|_{\infty,K_0}\Lip(f_2;K_0)
       +\|f_2\|_{\infty,K_0}\Lip(f_1;K_0).
\end{equation}
Equations~\eqref{eq:approx-factor-bounds} and
\eqref{eq:approx-product-lipschitz-rule} imply
\(\Lip(p^{(r)};K_0)\le rQ\). Set
\(L_r=\Lip(e_r;K_0)\). The map
\(x\mapsto(\widetilde p^{(r-1)}(x),z_r(x))\) has Lipschitz
constant at most \(\Lip(\widetilde p^{(r-1)};K_0)+Q\).
Applying Equation~\eqref{eq:approx-product-lipschitz-rule} and the
composition inequality for Lipschitz constants to
Equation~\eqref{eq:approx-product-error-identity} yields
\begin{align}
 L_r
 &\le\delta\{\Lip(\widetilde p^{(r-1)};K_0)+Q\}
             +\|e_{r-1}\|_{\infty,K_0}Q+L_{r-1}\nonumber\\
 &\le\delta\{(r-1)Q+L_{r-1}+Q\}
             +(r-2)\delta Q+L_{r-1}\nonumber\\
 &=(1+\delta)L_{r-1}+2(r-1)Q\delta.
                         \label{eq:approx-lipschitz-recurrence}
\end{align}
Since \(L_1=0\) and \(r\le s_*\), iterating
Equation~\eqref{eq:approx-lipschitz-recurrence} gives
\begin{equation}\label{eq:approx-local-product-errors}
 \begin{aligned}
 \|\widetilde p_{i,\nu}-p_{i,\nu}\|_{\infty,K_0}
       &\le (s_*-1)\delta,\\
 \Lip(\widetilde p_{i,\nu}-p_{i,\nu};K_0)
       &\le2Q\delta\sum_{j=2}^{s_{i,\nu}}
                    (j-1)(1+\delta)^{s_{i,\nu}-j}
        \le C_{s_*}Q\delta.
 \end{aligned}
\end{equation}
An empty sum in Equation~\eqref{eq:approx-local-product-errors} is
zero. Also
\(\Lip(\widetilde p_{i,\nu};K_0)\le C_{s_*}Q\).
All constants in these bounds are independent of \(i,\nu,Q,\delta\).

\medskip\noindent
\textbf{Step 5: Exact support preservation and summation of local errors.}
The exact-zero assertion in Lemma~\ref{lem:multiplier} states
\(P_{\delta,2}(s,0)=P_{\delta,2}(0,t)=0\) on \([-2,2]^2\).
If \(\phi_{i,d}(x_d)=0\) for any \(d\), the recursion in
Equation~\eqref{eq:approx-product-recursion} produces zero at the
step containing that hat factor and at every subsequent step.
The validity of these module inputs was proved in
Equation~\eqref{eq:approx-product-value-control}. Consequently,
\begin{equation}\label{eq:approx-support-preservation}
 \phi_{i,d}(x_d)=0\text{ for some }d
 \quad\Longrightarrow\quad
 \widetilde p_{i,\nu}(x)=p_{i,\nu}(x)=0.
\end{equation}
In particular, both products in
Equation~\eqref{eq:approx-support-preservation}, and their difference,
are supported inside the box \(B_i\) from
Equation~\eqref{eq:approx-local-box}. The network product
\(\widetilde p_{i,\nu}\) may have additional zeros; equality of the
two supports is neither needed nor asserted.

Define the vector-valued ReLU realization
\begin{equation}\label{eq:approx-neural-interpolant}
 \widetilde H_{Q,\delta}(x)
   =\sum_{i\in\mathcal I_Q}\sum_{\nu\in\mathcal M_k}
                      c_{i,\nu}\widetilde p_{i,\nu}(x).
\end{equation}
The exact interpolant \(H_Q\) in
Equation~\eqref{eq:approx-taylor-definitions} has the same expression
with \(p_{i,\nu}\) in place of \(\widetilde p_{i,\nu}\).
At every \(x\in K_0\), Equation~\eqref{eq:approx-support-preservation}
leaves at most \(2^DJ_k\) nonzero product errors. Therefore
\begin{equation}\label{eq:approx-summed-value-error}
 \|\widetilde H_{Q,\delta}-H_Q\|_{\infty,K_0}
 \le 2^DJ_kB'(s_*-1)\delta\le C\delta.
\end{equation}
To justify the derivative bound, exclude the mesh hyperplanes and the
finitely many nondifferentiability boundaries of the ReLU products
\(\widetilde p_{i,\nu}\). At each remaining point, a product error
whose box \(B_i\) does not contain that point vanishes on a
neighborhood and has derivative zero there. Each of the remaining,
at most \(2^DJ_k\), scalar product errors has derivative norm bounded
by its Lipschitz constant in
Equation~\eqref{eq:approx-local-product-errors}. It follows that
\begin{equation}\label{eq:approx-summed-derivative-error}
 \|D\widetilde H_{Q,\delta}-DH_Q\|_{L_\infty(K_0)}
      \le 2^DJ_kB'C_{s_*}Q\delta\le CQ\delta.
\end{equation}
The exceptional sets just excluded have Lebesgue measure zero: a finite
ReLU realization is affine on each member of a finite polyhedral
partition, and the exact products \(p_{i,\nu}\) are polynomial on
each mesh cell. Thus Equation~\eqref{eq:approx-summed-derivative-error}
is an essential-supremum bound in the sense of
Equation~\eqref{eq:approx-proof-norms}. In particular, neither
Equation~\eqref{eq:approx-summed-value-error} nor
Equation~\eqref{eq:approx-summed-derivative-error} incurs a factor
\(Q^D\) from the total number of grid points.

Fix \(0<c_0\le(4s_*)^{-1}\), set
\(\delta_Q=c_0Q^{-a}\), and put
\(\bar g_Q=\widetilde H_{Q,\delta_Q}\). Combining
Equations~\eqref{eq:approx-interpolant-bounds},
\eqref{eq:approx-summed-value-error}, and
\eqref{eq:approx-summed-derivative-error} gives
\begin{equation}\label{eq:approx-mesh-final-errors}
 \begin{aligned}
 \|\bar g_Q-\bar g\|_{\infty,K_0}
    &\le C(Q^{-a}+\delta_Q)\le C Q^{-a},\\
 \|D\bar g_Q-D\bar g\|_{L_\infty(K_0)}
    &\le C\{Q^{-(a-1)}+Q\delta_Q\}
      \le C Q^{-(a-1)}\quad(a>1).
 \end{aligned}
\end{equation}
For \(0<a\le1\), the construction uses \(k=0\), and only the
first inequality in Equation~\eqref{eq:approx-mesh-final-errors}
is asserted.

\medskip\noindent
\textbf{Step 6: Architecture of \(\bar g_Q\) and the choice of \(Q\).}
There are \((Q+1)^D\) grid points \(x_i\) and \(J_k\) Taylor
monomials at each point. Each product network
\(\widetilde p_{i,\nu}\) uses at most \(s_*-1\) multiplication
modules \(P_{\delta_Q,2}\). By Lemma~\ref{lem:multiplier}, each
such module has fixed width, depth at most
\(C\log(1/\delta_Q)\), at most \(C\log(1/\delta_Q)\) nonzero
parameters, and bounded parameter magnitudes. Since
\(\log(1/\delta_Q)=a\log Q+\log(1/c_0)\), the depth and
nonzero-parameter count of one product network are \(O(\log(eQ))\).
The exact hat formulas in Equation~\eqref{eq:approx-hat-network}
require only a fixed number of operations per product network, with
parameter magnitudes bounded by \(C Q\). The coordinate factors
\(x_d-x_{i,d}\) have coefficients bounded by one.

For completeness, a signed scalar quantity \(s\) that must be retained
between hidden layers is stored in two nonnegative coordinates
\(s^+=\varrho(s)\) and \(s^-=\varrho(-s)\), with
\(s=s^+-s^-\). A ReLU layer copies the two coordinates \(s^+,s^-\)
exactly using weights equal to one, and an affine operation reconstructs
\(s\) using weights \(1,-1\). The factors \(z_r\) and intermediate
products \(\widetilde p^{(r)}\) can therefore be propagated through
all multiplication layers using a fixed number of additional channels
per product network. These identity channels are included in the
nonzero-parameter count.

Run the \((Q+1)^DJ_k\) product networks in parallel and use an affine
output layer with coefficients \(c_{i,\nu}\) to form
Equation~\eqref{eq:approx-neural-interpolant}. Since \(m,D,k\)
are fixed and \(\|c_{i,\nu}\|\le B'\), the resulting network
\(\bar g_Q\) has bounds
\begin{equation}\label{eq:approx-architecture-counts}
 \begin{aligned}
 \operatorname{depth}(\bar g_Q)&\le C\log(eQ),&
 \operatorname{width}(\bar g_Q)&\le C Q^D,\\
 \#\{\text{nonzero parameters of }\bar g_Q\}
     &\le C Q^D\log(eQ),&
 \max\{|\text{parameter of }\bar g_Q|\}&\le C Q^{c_*},
 \end{aligned}
\end{equation}
where \(c_*\) is a fixed finite constant. The counts in
Equation~\eqref{eq:approx-architecture-counts} include the affine
summation layer, identity channels, and any padding needed to align
parallel branches. Fixed input transformations and fixed clipping
operations add only fixed-dimensional operations and satisfy the same
bounds after increasing \(C,c_*\).

For the given network resolution \(N\ge2\), choose
\begin{equation}\label{eq:approx-mesh-choice}
 Q=Q_N:=\max\{2,\lfloor N^{1/D}\rfloor\}.
\end{equation}
Then
\(\tfrac12N^{1/D}\le Q_N\le2N^{1/D}\) and
\(Q_N^D\le2^DN\). Substitution into
Equation~\eqref{eq:approx-architecture-counts} bounds the depth by
\(C\log(eN)\), the width by \(CN\), the number of nonzero
parameters by \(CN\log(eN)\), and the parameter magnitudes by
a fixed power of \(N\). Choosing the fixed constants
\(a_L,a_W,a_S,a_R\) in Equation~\eqref{eq:architecture} sufficiently
large embeds \(\bar g_{Q_N}\), and its affine-input transform
\(g_N=\bar g_{Q_N}\circ\Theta_K^{-1}\), in the declared architecture.
Padding to the declared hidden-layer count costs at most
\(CN\log(eN)\) additional nonzero parameters. Finally,
Equations~\eqref{eq:approx-affine-errors},
\eqref{eq:approx-mesh-final-errors}, and
\eqref{eq:approx-mesh-choice} prove
Equations~\eqref{eq:value-approximation} and
\eqref{eq:derivative-approximation}.

\medskip\noindent
\textbf{Step 7: From derivative error to block Lipschitz bounds.}
Suppose \(a>1\), and retain the pre-output-clipping realization
\(g_N\) constructed in Step~6. Define
\(E_N=g_N-g\) on \(K\) and
\(\varepsilon_N=C N^{-(a-1)/D}\), increasing \(C\) if necessary.
Equation~\eqref{eq:derivative-approximation} gives
\(\|DE_N\|_{L_\infty(K)}\le\varepsilon_N\).
We justify carefully why this almost-everywhere statement implies
\(\Lip(E_N;K)\le\varepsilon_N\), including line segments that
lie on boundaries of ReLU cells.

The map \(E_N\) is Lipschitz on \(K\), because the finite ReLU
realization \(g_N\) is Lipschitz and \(g\) is continuously
differentiable with bounded derivative on the convex rectangle \(K\).
Let \(\theta_\epsilon\) be a nonnegative smooth mollifier of integral
one, supported in the Euclidean ball of radius \(\epsilon>0\), and
set
\begin{equation}\label{eq:approx-mollification}
 K_\epsilon=\{q\in\operatorname{int}K:
                  \operatorname{dist}(q,\partial K)>\epsilon\},
 \qquad
 E_{N,\epsilon}(q)=\int\theta_\epsilon(z)E_N(q-z)\dd z.
\end{equation}
The latter integral is defined for \(q\in K_\epsilon\), so it
requires no extension of \(E_N\) outside \(K\). The weak derivative
of the Lipschitz map \(E_N\) equals its almost-everywhere derivative.
Differentiating its convolution therefore gives
\begin{equation}\label{eq:approx-mollified-derivative}
 DE_{N,\epsilon}(q)=\int\theta_\epsilon(z)DE_N(q-z)\dd z,
 \qquad
 \|DE_{N,\epsilon}(q)\|_{\rm op}\le\varepsilon_N.
\end{equation}
For interior points \(q,q'\in K\), choose \(\epsilon\) smaller
than the distance of their segment to \(\partial K\). The fundamental
theorem of calculus and
Equation~\eqref{eq:approx-mollified-derivative} imply
\begin{equation}\label{eq:approx-mollified-segment}
 \begin{aligned}
 E_{N,\epsilon}(q')-E_{N,\epsilon}(q)
   &=\int_0^1 DE_{N,\epsilon}(q+t(q'-q))(q'-q)\dd t,\\
 \|E_{N,\epsilon}(q')-E_{N,\epsilon}(q)\|
   &\le\varepsilon_N\|q'-q\|.
 \end{aligned}
\end{equation}
Letting \(\epsilon\downarrow0\) in
Equation~\eqref{eq:approx-mollified-segment} proves the same inequality
for \(E_N(q'),E_N(q)\), since convolution converges to the continuous
map \(E_N\) at both points. For arbitrary \(q,q'\in K\), fix
\(q_\circ\in\operatorname{int}K\) and apply the interior inequality
to \(q^{(t)}=(1-t)q+tq_\circ\) and
\(q'^{(t)}=(1-t)q'+tq_\circ\), where \(0<t<1\). Continuity
and \(t\downarrow0\) prove
\begin{equation}\label{eq:approx-error-global-lipschitz}
 \|E_N(q')-E_N(q)\|\le\varepsilon_N\|q'-q\|
 \quad(q,q'\in K),
 \qquad \Lip(E_N;K)\le\varepsilon_N.
\end{equation}

For a coordinate block \(b\) of dimension \(d_b\), let
\(J_b:\R^{d_b}\to\R^D\) insert a vector into the coordinates
of block \(b\), with zeros in the other coordinates. Then
\(\|J_bw\|=\|w\|\), and the block Lipschitz constant is
\begin{equation}\label{eq:approx-block-definition}
 \Lip_b(f;K)
 =\sup_{\substack{w\ne0,\ q\in K\\q+J_bw\in K}}
       \frac{\|f(q+J_bw)-f(q)\|}{\|w\|}.
\end{equation}
Adding and subtracting \(g(q+J_bw)\) and \(g(q)\), and applying
Equation~\eqref{eq:approx-error-global-lipschitz}, gives
\begin{equation}\label{eq:approx-block-bound}
 \Lip_b(g_N;K)\le\Lip_b(g;K)+\varepsilon_N.
\end{equation}

\medskip\noindent
\textbf{Step 8: Input projection \(\Pi_K\) and output clipping.}
Suppose \(g(K)\subset[-b_{\rm out},b_{\rm out}]^m\), and define
\(\widetilde g_N=\clip_{b_{\rm out}}\circ g_N\circ\Pi_K\)
on \(\R^D\). Coordinatewise projection onto a rectangle and
coordinatewise clipping onto a box are Euclidean nonexpansive:
the scalar projection inequality in each coordinate, followed by
summing squared coordinate differences, proves this assertion.
Moreover \(\Pi_K(q)=q\) for \(q\in K\), and
\(\clip_{b_{\rm out}}(g(q))=g(q)\). Hence
\begin{equation}\label{eq:approx-clipped-value}
 \|\widetilde g_N(q)-g(q)\|
 =\|\clip_{b_{\rm out}}(g_N(q))
                  -\clip_{b_{\rm out}}(g(q))\|
 \le\|g_N(q)-g(q)\| \quad(q\in K).
\end{equation}
If \(q,q'\in\R^D\) differ only in block \(b\), then
\(\Pi_K(q),\Pi_K(q')\) also differ only in block \(b\), and
\(\|\Pi_K(q')-\Pi_K(q)\|\le\|q'-q\|\).
Equation~\eqref{eq:approx-block-bound} therefore implies
\begin{equation}\label{eq:approx-clipped-lipschitz}
 \begin{aligned}
 \|\widetilde g_N(q')-\widetilde g_N(q)\|
 &\le\|g_N(\Pi_K(q'))-g_N(\Pi_K(q))\|\\
 &\le\{\Lip_b(g;K)+\varepsilon_N\}\|q'-q\|,
 \end{aligned}
\end{equation}
so \(\Lip_b(\widetilde g_N)\le\Lip_b(g;K)+\varepsilon_N\)
on the full input space. Equation~\eqref{eq:approx-clipped-value}
preserves the value error for every \(a>0\), whereas
Equation~\eqref{eq:approx-clipped-lipschitz} uses \(a>1\).
No claim that \(D\widetilde g_N-Dg\) retains the derivative error
in Equation~\eqref{eq:derivative-approximation} is needed: output
clipping can change derivatives where \(g_N\) leaves the clipping box.

\medskip\noindent
\textbf{Step 9: Membership of \(h_M\) and \(v_{j,N}\) in the sieves.}
Under Assumption~\ref{ass:smooth}, apply the value construction with
\((D,m,a,K)=(p,d_h,\beta_j,\mathcal Z_0)\) and network resolution
\(M\) to the comparison representation \(h_{j,0}\). Denote its
unclipped network by \(f_{h,M}\), and define
\begin{equation}\label{eq:approx-representation-comparator}
 h_M=\clip_1\circ f_{h,M}\circ\Pi_{\mathcal Z_0},
 \qquad
 \|h_M-h_{j,0}\|_{\infty,\mathcal Z_0}
                  \le C M^{-\beta_j/p}.
\end{equation}
The representation \(h_M\) belongs to \(\cH_M\) by its definition
in Section~\ref{sec:sieves}. This step needs only \(\beta_j>0\),
not \(\beta_j>1\), because \(\cH_M\) imposes no derivative
approximation requirement.

For the comparison head \(g_{j,0}\), use
\((D,m,a,K)=(D_j,d_j,r_j,\cK_j)\), where \(D_j\) is defined
in Equation~\eqref{eq:dimensions}. Let \(f_{j,N}\) be the
pre-output-clipping realization, and set
\(v_{j,N}=\clip_{b_j}\circ f_{j,N}\circ\Pi_{\cK_j}\).
Equations~\eqref{eq:approx-clipped-value} and
\eqref{eq:approx-clipped-lipschitz}, together with the margins in
Equation~\eqref{eq:smooth-margins}, give
\begin{equation}\label{eq:approx-head-comparator}
 \begin{aligned}
 \|v_{j,N}-g_{j,0}\|_{\infty,\cK_j}&\le C N^{-r_j/D_j},\\
 \Lip_b(v_{j,N})
 &\le L_{b,j}-\eta_{b,j}
                 +C_{{\rm der},j}N^{-(r_j-1)/D_j},
       \qquad b\in\{x,u,\tau\}.
 \end{aligned}
\end{equation}
Set \(\eta_{\min,j}=\min_{b\in\{x,u,\tau\}}\eta_{b,j}>0\)
and choose the fixed threshold
\begin{equation}\label{eq:approx-admissible-threshold}
 N_{0,j}=\max\left\{2,
   \left\lceil
     \left(\frac{C_{{\rm der},j}}{\eta_{\min,j}}\right)^{D_j/(r_j-1)}
   \right\rceil\right\}.
\end{equation}
For \(N\ge N_{0,j}\), the derivative-error term in
Equation~\eqref{eq:approx-head-comparator} is at most
\(\eta_{\min,j}\). Thus every constraint in
Equation~\eqref{eq:lipschitz} holds for \(v_{j,N}\), including
outside \(\cK_j\), and \(v_{j,N}\in\cV_{j,N}\).
Equations~\eqref{eq:approx-representation-comparator} and
\eqref{eq:approx-head-comparator} prove
Equation~\eqref{eq:admissible-approximation-proved}.

Finally, Assumption~\ref{ass:transfer} places every reduced conditional
mean \(g_{t,h}\), with representation \(h\in\cH_{\rm good}\),
in the same H\"older ball \(C^{r_t}(\cK_t;B_{g,t})\), the same
output box \([-b_t,b_t]^{d_t}\), and the same three positive margins
in Equation~\eqref{eq:reduced-regularity}. Repeating the construction
of \(v_{j,N}\) with \(g_{t,h}\) in place of \(g_{j,0}\) therefore
gives a head \(v_{t,N,h}\in\cV_{t,N}\) such that
\begin{equation}\label{eq:approx-uniform-reduced-head}
 \|v_{t,N,h}-g_{t,h}\|_{\infty,\cK_t}
        \le C N^{-r_t/D_t},
 \qquad h\in\cH_{\rm good},\quad N\ge N_{0,t}.
\end{equation}
The architecture constants, the approximation constant \(C\), and
the threshold \(N_{0,t}\) in
Equation~\eqref{eq:approx-uniform-reduced-head} are independent of
\(h\) and of the sample sizes. This proves the asserted uniformity.
\end{proof}

\begin{proposition}[Finite cell-slope characterization]\label{prop:cell-certificate}
A finite ReLU network with the declared input and output clips is continuous
and piecewise affine on a finite polyhedral partition. On each full-dimensional
cell \(C\), write its head as \(v(q)=A_Cq+b_C\).
For a block \(b\in\{x,u,\tau\}\), let \(A_{C,b}\) be the corresponding
columns. On the full input space,
\begin{equation}\label{eq:cell-certificate}
 \Lip_b(v)=\max_{C\text{ full-dimensional}}\|A_{C,b}\|_{\rm op}.
\end{equation}
Thus the actual constraints in \cref{eq:lipschitz} have a finite exact
characterization, although enumerating the cells may be expensive.
\end{proposition}
\begin{proof}\proofanchor{prop:cell-certificate}
Write \(v:\R^D\to\R^m\) for the head including its declared input
projection and output clipping. For the head sieve \(\cV_{j,N}\),
the dimensions are \(D=D_j\) and \(m=d_j\), with \(D_j\) as in
Equation~\eqref{eq:dimensions}. Fix a block \(b\in\{x,u,\tau\}\)
of dimension \(d_b\), and let \(J_b:\R^{d_b}\to\R^D\) be the
coordinate insertion map defined in
Equation~\eqref{eq:approx-block-definition}. In particular,
\(\|J_bw\|=\|w\|\). The definition on the full input space is
\begin{equation}\label{eq:cell-block-definition}
 \Lip_b(v)
   =\sup_{\substack{q\in\R^D\\w\in\R^{d_b}\setminus\{0\}}}
       \frac{\|v(q+J_bw)-v(q)\|}{\|w\|}.
\end{equation}
The matrix of block columns in the proposition is
\(A_{C,b}=A_CJ_b\).

\medskip\noindent
\textbf{Step 1: A finite affine representation on polyhedral cells.}
Let \(\varrho(s)=\max(s,0)\). For any interval \([\ell,r]\),
scalar coordinate clipping has the exact representation
\begin{equation}\label{eq:cell-clipping-formula}
 \operatorname{clip}_{[\ell,r]}(s)
   =\ell+\varrho(s-\ell)-\varrho(s-r).
\end{equation}
Thus each coordinate of the input projection and the output clipping
is itself a finite ReLU realization. Absorbing these operations into a
finite network, write its hidden-layer computation as
\begin{equation}\label{eq:cell-network-recursion}
 \begin{aligned}
 z^{(0)}(q)&=q,\\
 u^{(\ell)}(q)&=W_\ell z^{(\ell-1)}(q)+c_\ell,&
 z^{(\ell)}(q)&=\varrho(u^{(\ell)}(q)),
                       \quad1\le\ell\le L,\\
 v(q)&=W_{L+1}z^{(L)}(q)+c_{L+1}.&
 \end{aligned}
\end{equation}
Here \(L\) is the finite hidden-layer count, \(n_\ell\) is the
width of layer \(\ell\), \(W_\ell\) is its weight matrix, and
\(c_\ell\) is its bias vector. All affine maps and all coordinatewise
ReLU maps in Equation~\eqref{eq:cell-network-recursion} are continuous;
therefore their composition \(v\) is continuous on \(\R^D\).

An activation pattern is a collection
\(\omega=(\omega_1,\ldots,\omega_L)\), with
\(\omega_\ell\in\{0,1\}^{n_\ell}\). Define its diagonal gate
matrix \(\Gamma_{\omega,\ell}=\operatorname{diag}(\omega_\ell)\).
For this pattern, set \(A_{\omega,0}=I_D\),
\(b_{\omega,0}=0\), and recursively define
\begin{equation}\label{eq:cell-pattern-affine-recursion}
 \begin{aligned}
 B_{\omega,\ell}&=W_\ell A_{\omega,\ell-1},&
 d_{\omega,\ell}&=W_\ell b_{\omega,\ell-1}+c_\ell,\\
 A_{\omega,\ell}&=\Gamma_{\omega,\ell}B_{\omega,\ell},&
 b_{\omega,\ell}&=\Gamma_{\omega,\ell}d_{\omega,\ell}.
 \end{aligned}
\end{equation}
The consistency region for the activation pattern \(\omega\) is
\begin{equation}\label{eq:cell-pattern-region}
 C_\omega
 =\bigcap_{\ell=1}^L\bigcap_{r=1}^{n_\ell}
   \left\{q\in\R^D:
    (2\omega_{\ell,r}-1)
          [B_{\omega,\ell}q+d_{\omega,\ell}]_r\ge0\right\}.
\end{equation}
Every set in this intersection is a closed half-space, the whole
space, or the empty set. Hence \(C_\omega\) is a closed convex
polyhedron, possibly empty or lower-dimensional. If \(q\in C_\omega\),
induction over \(\ell\) using
Equations~\eqref{eq:cell-network-recursion} and
\eqref{eq:cell-pattern-region} proves
\(z^{(\ell)}(q)=A_{\omega,\ell}q+b_{\omega,\ell}\).
This remains true for zero preactivations, because both choices of
a gate on a zero coordinate produce zero. Consequently,
\begin{equation}\label{eq:cell-pattern-output}
 v(q)=A_\omega q+b_\omega\quad(q\in C_\omega),
 \qquad
 \begin{cases}
 A_\omega=W_{L+1}A_{\omega,L},\\
 b_\omega=W_{L+1}b_{\omega,L}+c_{L+1}.
 \end{cases}
\end{equation}
There are at most \(2^{\sum_{\ell=1}^Ln_\ell}\) activation patterns,
and their regions \(C_\omega\) cover \(\R^D\), because the actual
preactivation signs at every input \(q\) provide a consistent pattern.

To obtain a partition with disjoint interiors, collect the finitely
many hyperplanes defined by all nonconstant affine expressions in
Equation~\eqref{eq:cell-pattern-region}. Constant affine expressions
need not be included: a nonzero constant has a fixed sign, and an
identically zero expression imposes no sign restriction. The nonempty
open cells of this finite hyperplane arrangement are intersections of
strict half-spaces and therefore are convex and full-dimensional.
On an open cell, every relevant affine expression has a fixed sign,
so that cell lies in at least one region \(C_\omega\); Equation~\eqref{eq:cell-pattern-output}
then gives one affine formula for \(v\) on the cell. If two patterns
are consistent there, their affine formulas agree on an open set and
hence have identical coefficients.

Let \(\mathscr C\) be the finite collection of closures of these
nonempty open cells. Every \(C\in\mathscr C\) is a closed,
full-dimensional convex polyhedron. Their union is \(\R^D\): the
complement of the finitely many hyperplanes is dense, and any convergent
sequence in that complement has a subsequence lying in one of its
finitely many cells. Continuity of \(v\) extends each affine formula
to the cell closure, giving
\begin{equation}\label{eq:cell-closed-partition}
 \bigcup_{C\in\mathscr C}C=\R^D,
 \qquad v(q)=A_Cq+b_C\quad(q\in C).
\end{equation}
If the hyperplane collection is empty, take
\(\mathscr C=\{\R^D\}\). The cell closures and their faces give
a finite polyhedral partition in the usual sense. Subdividing an
already affine cell does not change its matrix \(A_C\), and therefore
does not change the maximum of the block operator norms.

\medskip\noindent
\textbf{Step 2: An exact subdivision of every blockwise segment.}
Define the finite nonnegative number
\begin{equation}\label{eq:cell-maximum-slope}
 L_b^*=\max_{C\in\mathscr C}\|A_CJ_b\|_{\rm op}.
\end{equation}
Fix an arbitrary input \(q\in\R^D\) and a nonzero block increment
\(w\in\R^{d_b}\), and parameterize their segment by
\begin{equation}\label{eq:cell-segment}
 q(t)=q+tJ_bw,\qquad 0\le t\le1.
\end{equation}
For each closed convex cell \(C\in\mathscr C\), the set
\(I_C=\{t\in[0,1]:q(t)\in C\}\) is empty or a closed interval,
possibly a singleton. The intervals \(I_C\) cover \([0,1]\) by
Equation~\eqref{eq:cell-closed-partition}. List all their endpoints,
together with 0 and 1, in strictly increasing order:
\begin{equation}\label{eq:cell-segment-subdivision}
 0=t_0<t_1<\cdots<t_J=1.
\end{equation}
For each \(j=1,\ldots,J\), choose a cell \(C_j\in\mathscr C\)
whose interval \(I_{C_j}\) contains the midpoint of
\((t_{j-1},t_j)\). No endpoint of \(I_{C_j}\) can lie strictly
between \(t_{j-1}\) and \(t_j\), by the definition of
Equation~\eqref{eq:cell-segment-subdivision}. Since \(I_{C_j}\)
is closed, it follows that
\begin{equation}\label{eq:cell-segment-contained}
 [t_{j-1},t_j]\subset I_{C_j},
 \qquad q([t_{j-1},t_j])\subset C_j.
\end{equation}
This argument does not require a transverse crossing of cell boundaries.
It applies even when the entire segment \(q([0,1])\) lies in a
lower-dimensional face, because Equation~\eqref{eq:cell-closed-partition}
is valid on the closed full-dimensional cells containing that face.

\medskip\noindent
\textbf{Step 3: The upper bound \(\Lip_b(v)\le L_b^*\).}
Equations~\eqref{eq:cell-closed-partition} and
\eqref{eq:cell-segment-contained} give the exact increment identity
\begin{equation}\label{eq:cell-affine-increment}
 \begin{aligned}
 v(q(t_j))-v(q(t_{j-1}))
   &=A_{C_j}\{q(t_j)-q(t_{j-1})\}\\
   &=(t_j-t_{j-1})A_{C_j}J_bw.
 \end{aligned}
\end{equation}
Summing Equation~\eqref{eq:cell-affine-increment} telescopically,
then applying the triangle inequality and the definition of an operator
norm, yields
\begin{align}
 \|v(q+J_bw)-v(q)\|
 &=\left\|\sum_{j=1}^J(t_j-t_{j-1})A_{C_j}J_bw\right\|\nonumber\\
 &\le\sum_{j=1}^J(t_j-t_{j-1})\|A_{C_j}J_b\|_{\rm op}\|w\|
                                                        \nonumber\\
 &\le L_b^*\|w\|\sum_{j=1}^J(t_j-t_{j-1})
  =L_b^*\|w\|.                         \label{eq:cell-upper-bound}
\end{align}
Taking the supremum over \(q,w\) in
Equation~\eqref{eq:cell-block-definition} proves
\(\Lip_b(v)\le L_b^*\). In particular, this argument proves the
bound for every segment, not merely for almost every segment.

\medskip\noindent
\textbf{Step 4: The lower bound \(\Lip_b(v)\ge L_b^*\).}
Fix any full-dimensional cell \(C\in\mathscr C\). Choose
\(q_C\in\operatorname{int}C\) and \(r_C>0\) such that the
Euclidean ball of radius \(r_C\) centered at \(q_C\) is contained
in \(C\). For every unit vector \(w\in\R^{d_b}\) and every
\(0<t<r_C\), both \(q_C\) and \(q_C+tJ_bw\) lie in \(C\).
Therefore Equation~\eqref{eq:cell-closed-partition} and the definition
in Equation~\eqref{eq:cell-block-definition} give
\begin{equation}\label{eq:cell-directional-lower-bound}
 \Lip_b(v)
 \ge\frac{\|v(q_C+tJ_bw)-v(q_C)\|}{t}
 =\frac{\|tA_CJ_bw\|}{t}
 =\|A_CJ_bw\|.
\end{equation}
Taking the supremum over \(\|w\|=1\) proves
\(\Lip_b(v)\ge\|A_CJ_b\|_{\rm op}\). The cell \(C\) was
arbitrary, so Equation~\eqref{eq:cell-maximum-slope} implies
\(\Lip_b(v)\ge L_b^*\). Combining this lower bound with
Equation~\eqref{eq:cell-upper-bound} proves
Equation~\eqref{eq:cell-certificate}.

\medskip\noindent
\textbf{Step 5: The finite admissibility conditions for \(v\).}
Since \(A_{C,b}=A_CJ_b\), the equality just proved gives, for each
\(b\in\{x,u,\tau\}\),
\begin{equation}\label{eq:cell-finite-matrix-tests}
 \begin{aligned}
 \Lip_b(v)\le L_{b,j}
 &\quad\Longleftrightarrow\quad
   \|A_{C,b}\|_{\rm op}\le L_{b,j}
                   \quad\text{for every }C\in\mathscr C\\
 &\quad\Longleftrightarrow\quad
   A_{C,b}^{\mathsf T}A_{C,b}\preceq L_{b,j}^2I_{d_b}
                   \quad\text{for every }C\in\mathscr C.
 \end{aligned}
\end{equation}
The second equivalence follows from
\(\|A_{C,b}\|_{\rm op}^2
 =\sup_{\|w\|=1}w^{\mathsf T}A_{C,b}^{\mathsf T}A_{C,b}w\).
Thus Equation~\eqref{eq:lipschitz} is equivalent to finitely many
matrix inequalities in Equation~\eqref{eq:cell-finite-matrix-tests}.
All cells of the clipped network \(v\), including cells outside the
training rectangle, occur in this characterization. No additional
slope or jump condition on a lower-dimensional face is needed, because
continuity already ensures that the affine formulas agree there.
The proof establishes finiteness and exactness; it makes no claim that
constructing the cell collection \(\mathscr C\) is computationally
inexpensive.
\end{proof}

\subsection{Proper covers of constrained classes}
For a class \(\mathcal G\) of bounded functions, let
\(\mathcal N(\epsilon,\mathcal G,\|\cdot\|_\infty)\) be the smallest
size of an \(\epsilon\)-cover with centers in \(\mathcal G\), and write
\(H_{\mathcal G}(\epsilon)=\log\mathcal N(\epsilon,\mathcal G,
\|\cdot\|_\infty)\). These are proper covers: a center must obey all
constraints defining the class.

\begin{lemma}[Neural entropy and composition]\label{lem:entropy}
Fix a network architecture with \(P\) parameter positions, depth \(L\),
maximum width \(W\), at most \(S\le P\) nonzero parameters, parameter
magnitudes at most \(R\ge1\), bounded projected inputs, and fixed output
dimension. Every nonempty subclass obtained by additional functional
constraints has a proper cover satisfying
\begin{equation}\label{eq:network-entropy}
 H(\epsilon)\le
 S\log(eP/S)+S\log\{1+C R C_{\rm par}/\epsilon\},
 \qquad
 \log C_{\rm par}\le C(L+1)\log\{C(W+1)R\}.
\end{equation}
For the sieves in \cref{eq:architecture}, if \(N\) is bounded by a fixed
power of \(n\), then
\begin{equation}\label{eq:sieve-entropy}
 H_{\cV_{j,N}}(n^{-2})\le CN\log^3(en),\qquad
 H_{\cH_M}(n^{-2})\le CM\log^3(en),
\end{equation}
with the analogous polynomial-growth condition on the encoder
resolution \(M\). The supremum norm is global
because the inputs are projected onto fixed rectangles.
For \(\cF_j(N,M)=\{F_{j,v,h}:v\in\cV_{j,N},h\in\cH_M\}\),
\begin{equation}\label{eq:composition-entropy}
 H_{\cF_j(N,M)}(\epsilon)
 \le H_{\cV_{j,N}}(\epsilon/2)
      +H_{\cH_M}(\epsilon/(2L_{u,j})).
\end{equation}
For any fixed bounded representation \(h\), including a representation
fixed by independent source training,
\begin{equation}\label{eq:frozen-entropy}
 H_{\{F_{j,v,h}:v\in\cV_{j,N}\}}(\epsilon)
      \le H_{\cV_{j,N}}(\epsilon).
\end{equation}
\end{lemma}
\begin{proof}\proofanchor{lem:entropy}
Let \(\mathcal G\) be the nonempty function subclass in the first
assertion, and write
\(H_{\mathcal G}(\epsilon)=\log\mathcal N(\epsilon,\mathcal G,
\|\cdot\|_\infty)\), with the proper-cover convention stated immediately
before Lemma~\ref{lem:entropy}. Throughout the proof of
Equation~\eqref{eq:network-entropy}, a constant \(C\) may depend on
the fixed input dimension \(D\), output dimension \(m\), and input
rectangle \(K\), but not on \(\epsilon,P,S,L,W,R\) or on the
particular subclass \(\mathcal G\). When applying the bound to
Equation~\eqref{eq:architecture}, the constant \(C\) may additionally
depend on the fixed architecture constants in that equation.
If \(S=0\), all trainable parameters are zero, so the nonempty subclass
\(\mathcal G\) contains only one realized function and
\(H_{\mathcal G}(\epsilon)=0\). The first assertion then holds with
the convention \(0\log(eP/0)=0\). We henceforth suppose
\(1\le S\le P\).

\medskip\noindent
\textbf{Step 1: Network realizations \(f_\theta\) and the global norm.}
For a vector \(a\), put \(|a|_{\max}=\max_k|a_k|\); this
coordinate maximum norm is distinct from the Euclidean output norm
used in \(\|f\|_\infty\). Let \(\theta\in[-R,R]^P\) list the
trainable weights and biases in the fixed network architecture. Let
\(r\le L+1\) be the number of affine maps in its network core,
including the affine output map. Under the hidden-layer convention
in Section~\ref{sec:sieves}, \(r=L+1\). Write \(w_0=D\),
\(w_r=m\), and \(w_\ell\le W\) for the hidden-layer widths
\(1\le\ell<r\). For the weight matrix
\(A_\ell^\theta\in\R^{w_\ell\times w_{\ell-1}}\) and bias
\(b_\ell^\theta\in\R^{w_\ell}\), define
\begin{equation}\label{eq:entropy-layer-realization}
 \begin{aligned}
 y_0^\theta(q)&=q,\\
 y_\ell^\theta(q)
   &=\rho_\ell\bigl(A_\ell^\theta y_{\ell-1}^\theta(q)
                                      +b_\ell^\theta\bigr),
                      &&1\le\ell\le r,\\
 f_\theta(x)&=T_{\rm out}\bigl(y_r^\theta(\Pi_Kx)\bigr).
 \end{aligned}
\end{equation}
Here \(\rho_\ell\) is coordinatewise ReLU for \(\ell<r\),
\(\rho_r\) is the identity map, \(\Pi_K\) is the fixed input
projection, and \(T_{\rm out}\) is the fixed coordinatewise output
clip, or the identity map when no output clip is imposed. Each map
\(\rho_\ell\) is 1-Lipschitz in \(|\cdot|_{\max}\) and fixes zero;
the map \(T_{\rm out}\) is 1-Lipschitz in the Euclidean norm.
Absent connections in the architecture have coefficient zero.

Because \(\Pi_K(\R^D)=K\) and \(\Pi_Kq=q\) for \(q\in K\),
any two parameter vectors \(\theta,\theta'\) satisfy
\begin{equation}\label{eq:entropy-global-norm}
 \begin{aligned}
 \|f_\theta-f_{\theta'}\|_\infty
 &=\sup_{q\in K}
   \bigl\|T_{\rm out}(y_r^\theta(q))
                -T_{\rm out}(y_r^{\theta'}(q))\bigr\|\\
 &\le\sqrt m\sup_{q\in K}
                     |y_r^\theta(q)-y_r^{\theta'}(q)|_{\max}.
 \end{aligned}
\end{equation}
Thus a uniform bound on the fixed rectangle \(K\) in
Equation~\eqref{eq:entropy-global-norm} controls the global supremum
norm, not merely a norm on observed inputs.

\medskip\noindent
\textbf{Step 2: A parameter-to-function bound \(C_{\rm par}\).}
Define the fixed input bound \(B_0\), augmented width \(w_*\),
and layer factor \(Q_*\) by
\begin{equation}\label{eq:entropy-layer-constants}
 B_0=\max\{1,\sup_{q\in K}|q|_{\max}\},\qquad
 w_*=\max\{D,m,W,1\},\qquad Q_*=(w_*+1)R.
\end{equation}
In particular, \(Q_*\ge2\). Let
\(M_\ell=\sup_{\theta\in[-R,R]^P}\sup_{q\in K}
|y_\ell^\theta(q)|_{\max}\) be the uniform activation bound at
layer \(\ell\). The supremum defining \(M_\ell\) also includes
dense parameter vectors; a bound for this larger set applies to every
sparse parameter vector. Since every entry of
\(A_\ell^\theta,b_\ell^\theta\) has magnitude at most \(R\),
Equation~\eqref{eq:entropy-layer-realization} gives
\(M_0\le B_0\) and
\begin{equation}\label{eq:entropy-activation-bound}
 \begin{aligned}
 M_\ell
 &\le w_*R M_{\ell-1}+R\\
 &\le w_*R B_0Q_*^{\ell-1}+R
 \le (w_*+1)R B_0Q_*^{\ell-1}
   =B_0Q_*^\ell,\qquad 1\le\ell\le r.
 \end{aligned}
\end{equation}
The second line is an induction on \(\ell\); its last inequality
uses \(B_0Q_*^{\ell-1}\ge1\).

Now fix \(\theta,\theta'\in[-R,R]^P\), and define their parameter
distance and activation differences by
\begin{equation}\label{eq:entropy-parameter-differences}
 d_\theta=|\theta-\theta'|_{\max},\qquad
 E_\ell=\sup_{q\in K}|y_\ell^\theta(q)
                         -y_\ell^{\theta'}(q)|_{\max},\qquad E_0=0.
\end{equation}
For each input \(q\in K\), the difference of the two affine
preactivations in layer \(\ell\) is exactly
\begin{equation}\label{eq:entropy-preactivation-difference}
 \begin{aligned}
 &A_\ell^\theta y_{\ell-1}^\theta(q)+b_\ell^\theta
       -A_\ell^{\theta'}y_{\ell-1}^{\theta'}(q)-b_\ell^{\theta'}\\
 &\quad=A_\ell^\theta
              \{y_{\ell-1}^\theta(q)-y_{\ell-1}^{\theta'}(q)\}
       +(A_\ell^\theta-A_\ell^{\theta'})
                                  y_{\ell-1}^{\theta'}(q)
       +(b_\ell^\theta-b_\ell^{\theta'}).
 \end{aligned}
\end{equation}
Using the 1-Lipschitz property of \(\rho_\ell\), the row length
bound \(w_{\ell-1}\le w_*\), and the entrywise difference bound
\(d_\theta\) in Equation~\eqref{eq:entropy-preactivation-difference}
yields
\begin{equation}\label{eq:entropy-difference-recursion}
 \begin{aligned}
 E_\ell
 &\le w_*R E_{\ell-1}+w_*d_\theta M_{\ell-1}+d_\theta\\
 &\le Q_*E_{\ell-1}
           +(w_*+1)B_0Q_*^{\ell-1}d_\theta.
 \end{aligned}
\end{equation}
The second inequality follows from
Equations~\eqref{eq:entropy-layer-constants} and
\eqref{eq:entropy-activation-bound}. Iterating
Equation~\eqref{eq:entropy-difference-recursion} from \(E_0=0\)
gives the explicit finite sum
\begin{equation}\label{eq:entropy-difference-solution}
 \begin{aligned}
 E_\ell
 &\le (w_*+1)B_0d_\theta
             \sum_{k=1}^{\ell}Q_*^{\ell-k}Q_*^{k-1}\\
 &=\ell(w_*+1)B_0Q_*^{\ell-1}d_\theta
 \le\ell B_0Q_*^\ell d_\theta,
                   \qquad 1\le\ell\le r.
 \end{aligned}
\end{equation}
The last inequality uses \(w_*+1\le Q_*\), which follows from
\(R\ge1\). Combining
Equations~\eqref{eq:entropy-global-norm} and
\eqref{eq:entropy-difference-solution}, and using \(r\le L+1\),
proves
\begin{equation}\label{eq:entropy-parameter-lipschitz}
 \|f_\theta-f_{\theta'}\|_\infty
 \le C_{\rm par}|\theta-\theta'|_{\max},\qquad
 C_{\rm par}=\sqrt m\,B_0(L+1)Q_*^{L+1}.
\end{equation}
This comparison holds for all pairs \(\theta,\theta'\) in the
parameter box, hence in particular for pairs with a common support.
Since \(w_*+1\le C(W+1)\), with \(D,m\) fixed, the constant
\(C_{\rm par}\) in Equation~\eqref{eq:entropy-parameter-lipschitz}
satisfies
\begin{equation}\label{eq:entropy-log-parameter-lipschitz}
 \begin{aligned}
 \log C_{\rm par}
 &=\tfrac12\log m+\log B_0+\log(L+1)+(L+1)\log Q_*\\
 &\le C(L+1)\log\{C(W+1)R\}.
 \end{aligned}
\end{equation}
Here \(\log(L+1)\le L+1\), and the remaining fixed terms are
absorbed by increasing \(C\). This proves the second inequality
in Equation~\eqref{eq:network-entropy}.

\medskip\noindent
\textbf{Step 3: Supports \(J\), parameter cells \(B\), and proper centers.}
Define the sparse parameter set \(\Theta_S\), the feasible parameter
set \(\Theta_{\mathcal G}\), and the coordinate boxes \(\Theta_J\) by
\begin{equation}\label{eq:entropy-feasible-parameter-sets}
 \begin{aligned}
 \Theta_S&=\{\theta\in[-R,R]^P:|\supp(\theta)|\le S\},\\
 \Theta_{\mathcal G}&=\{\theta\in\Theta_S:f_\theta\in\mathcal G\},\\
 \mathcal J_S&=\{J\subseteq\{1,\ldots,P\}:|J|=S\},\\
 \Theta_J&=\{\theta\in[-R,R]^P:\theta_k=0\text{ for }k\notin J\},
                              \qquad J\in\mathcal J_S.
 \end{aligned}
\end{equation}
Every support \(\supp(\theta)\) with at most \(S\) elements is
contained in some set \(J\in\mathcal J_S\). Consequently,
\(\Theta_S=\bigcup_{J\in\mathcal J_S}\Theta_J\). Coordinates in
\(J\) are allowed to equal zero; thus this union includes all supports
of size less than \(S\) without separately counting their sizes.
Moreover,
\begin{equation}\label{eq:entropy-support-count}
 |\mathcal J_S|=\binom{P}{S}
 \le\frac{P^S}{S!}\le\left(\frac{eP}{S}\right)^S.
\end{equation}
For the last inequality, the exponential series gives
\(e^S\ge S^S/S!\), or equivalently \(S!\ge(S/e)^S\).

Fix \(\epsilon>0\) and put
\(\delta=\epsilon/(2C_{\rm par})\). Partition each active coordinate
interval \([-R,R]\) into
\(k_\delta=\max\{1,\lceil2R/\delta\rceil\}\) intervals of length
at most \(\delta\), assigning shared endpoints to one interval.
Then each box \(\Theta_J\) is partitioned into at most
\(k_\delta^S\) parameter cells \(B\), each of coordinate maximum
diameter at most \(\delta\). For every pair \((J,B)\) for which
\(B\cap\Theta_{\mathcal G}\ne\varnothing\), choose one parameter
vector
\(\theta^{J,B}\in B\cap\Theta_{\mathcal G}\). The finite set
\begin{equation}\label{eq:entropy-proper-center-set}
 \mathcal C_\epsilon
 =\{f_{\theta^{J,B}}:J\in\mathcal J_S,
                         \ B\cap\Theta_{\mathcal G}\ne\varnothing\}
 \subseteq\mathcal G
\end{equation}
is a proper cover. Indeed, for any \(f\in\mathcal G\), take a
feasible parameterization \(f=f_\theta\), select
\(J\in\mathcal J_S\) with \(\supp(\theta)\subseteq J\), and let
\(B\) be its parameter cell. Equation~\eqref{eq:entropy-parameter-lipschitz}
then gives
\begin{equation}\label{eq:entropy-cell-error}
 \|f-f_{\theta^{J,B}}\|_\infty
 \le C_{\rm par}|\theta-\theta^{J,B}|_{\max}
 \le C_{\rm par}\delta=\epsilon/2\le\epsilon.
\end{equation}
No closedness of \(\Theta_{\mathcal G}\) is required: only one
point from each of finitely many nonempty intersections is chosen.
In particular, the center \(f_{\theta^{J,B}}\) satisfies every
functional constraint defining \(\mathcal G\). The partition of the
parameter box is not a rounding rule for \(\theta\); a corner of its
cell need not satisfy those constraints.

The interval count satisfies
\(k_\delta\le1+2R/\delta=1+4RC_{\rm par}/\epsilon\).
Equations~\eqref{eq:entropy-support-count}--\eqref{eq:entropy-cell-error}
therefore imply
\begin{equation}\label{eq:entropy-proper-cover-count}
 \begin{aligned}
 \mathcal N(\epsilon,\mathcal G,\|\cdot\|_\infty)
 &\le |\mathcal C_\epsilon|
 \le \binom{P}{S}\,k_\delta^S\\
 &\le\left(\frac{eP}{S}\right)^S
       \left(1+\frac{4RC_{\rm par}}{\epsilon}\right)^S.
 \end{aligned}
\end{equation}
Taking logarithms in Equation~\eqref{eq:entropy-proper-cover-count}
proves the first inequality in Equation~\eqref{eq:network-entropy}.
Different parameter cells or support boxes may represent the same
function; such duplications only decrease the required cover size.

\medskip\noindent
\textbf{Step 4: The sieve bound \(CN\log^3(en)\).}
First consider a core sieve \(\cN_{D,m}(N)\) with fixed \(D,m\).
Let \(P_N\) be its number of trainable parameter positions, and let
\(\bar S_N=\min\{S_N,P_N\}\) be its effective sparsity budget.
The minimum defining \(\bar S_N\) accounts for the redundant-budget
convention in Section~\ref{sec:sieves}. Fixed input and output clips
are the maps \(\Pi_K,T_{\rm out}\) in
Equation~\eqref{eq:entropy-layer-realization}; their coefficients do
not need to be discretized. Counting the first hidden layer, the
remaining hidden layers, and the affine output layer separately gives
\begin{equation}\label{eq:entropy-sieve-position-count}
 \begin{aligned}
 P_N
 &=W_N(D+1)+(L_N-1)W_N(W_N+1)+m(W_N+1)\\
 &\le C(L_N+1)(W_N+1)^2
 \le CN^2\log(eN),\\
 \bar S_N&\le S_N\le CN\log(eN).
 \end{aligned}
\end{equation}
Here \(L_N\ge1\), and the last inequalities follow from
Equation~\eqref{eq:architecture}. A fixed sparse wiring pattern can
only reduce \(P_N\). The same architecture formula, together with
Equation~\eqref{eq:entropy-log-parameter-lipschitz}, gives
\begin{equation}\label{eq:entropy-sieve-log-bounds}
 \begin{aligned}
 \log(eP_N)&\le C\log(eN),&
 \log R_N&\le C\log(eN),\\
 \log C_{{\rm par},N}
 &\le C(L_N+1)\log\{C(W_N+1)R_N\}
 \le C\log^2(eN).
 \end{aligned}
\end{equation}
The constants in Equation~\eqref{eq:architecture} are chosen large
enough that \(R_N\ge1\); alternatively, replacing \(R_N\) by
\(\max\{1,R_N\}\) in this upper-bound calculation gives the same
orders.

Let \(\mathcal G_N\) be any nonempty constrained subclass of this
projected, possibly output-clipped sieve. Fix \(c_\epsilon>0\),
independent of \(n,N\), and take \(\epsilon=c_\epsilon n^{-2}\)
with \(n\ge2\). Since \(R_NC_{{\rm par},N}n^2\ge1\),
\begin{equation}\label{eq:entropy-grid-log-at-n}
 \begin{aligned}
 \log\left(1+\frac{CR_NC_{{\rm par},N}n^2}{c_\epsilon}\right)
 &\le\log(1+C/c_\epsilon)+\log R_N
                        +\log C_{{\rm par},N}+2\log n\\
 &\le C_{c_\epsilon}\{\log^2(eN)+\log(en)\}.
 \end{aligned}
\end{equation}
The first inequality uses \(1+ab\le(1+a)b\) for \(a\ge0,b\ge1\);
the second uses Equation~\eqref{eq:entropy-sieve-log-bounds}.
Substituting \(P=P_N\), \(S=\bar S_N\), and
\(\epsilon=c_\epsilon n^{-2}\) into
Equation~\eqref{eq:network-entropy} yields
\begin{equation}\label{eq:entropy-sieve-substitution}
 \begin{aligned}
 H_{\mathcal G_N}(c_\epsilon n^{-2})
 &\le \bar S_N\log(eP_N/\bar S_N)
   +\bar S_N\log\left(1+
               \frac{CR_NC_{{\rm par},N}n^2}{c_\epsilon}\right)\\
 &\le C_{c_\epsilon}N\log(eN)
                     \{\log^2(eN)+\log(en)\}.
 \end{aligned}
\end{equation}
For the support term, we used
\(\log(eP_N/\bar S_N)\le\log(eP_N)\) and
Equations~\eqref{eq:entropy-sieve-position-count} and
\eqref{eq:entropy-sieve-log-bounds}.
If \(N\le c_0n^\kappa\) for fixed \(c_0\ge1\), \(\kappa>0\),
then
\(\log(eN)\le1+\log c_0+\kappa\log n
\le C_{c_0,\kappa}\log(en)\). Hence
Equation~\eqref{eq:entropy-sieve-substitution} implies
\begin{equation}\label{eq:entropy-polynomial-resolution}
 H_{\mathcal G_N}(c_\epsilon n^{-2})
 \le C_{c_\epsilon,c_0,\kappa}N\log^3(en).
\end{equation}
The third logarithmic power in
Equation~\eqref{eq:entropy-polynomial-resolution} comes from
multiplying the sparsity order \(N\log(eN)\) by the parameter-cell
logarithm of order at most \(\log^2(en)\).

For the head sieve \(\cV_{j,N}\), set \(D=D_j\), \(m=d_j\),
\(K=\cK_j\), and \(T_{\rm out}=\clip_{b_j}\). The admissible
subclass is defined by Equation~\eqref{eq:lipschitz}; the feasible
center construction in Step~3 preserves precisely these constraints.
For the representation sieve \(\cH_M\), repeat the calculation with
resolution \(M\), input dimension \(p\), output dimension \(d_h\),
input rectangle \(\mathcal Z_0\), and output clip \(\clip_1\).
Under its analogous polynomial-growth condition on \(M\), this gives
\begin{equation}\label{eq:entropy-head-encoder-constant-radii}
 H_{\cV_{j,N}}(c_\epsilon n^{-2})\le CN\log^3(en),\qquad
 H_{\cH_M}(c_\epsilon n^{-2})\le CM\log^3(en).
\end{equation}
Taking \(c_\epsilon=1\) proves Equation~\eqref{eq:sieve-entropy}.
Equation~\eqref{eq:entropy-global-norm} establishes the claimed global
norm in both cases. Keeping arbitrary fixed \(c_\epsilon>0\) also
justifies the rescaled radii used in the next step.

\medskip\noindent
\textbf{Step 5: A proper product cover for \(\cF_j(N,M)\).}
Fix \(\epsilon>0\). The constant \(L_{u,j}\) is strictly positive
by the convention following Equation~\eqref{eq:lipschitz}. Choose
proper covers
\(\{v^{(a)}:1\le a\le K_V\}\subseteq\cV_{j,N}\) and
\(\{h^{(b)}:1\le b\le K_H\}\subseteq\cH_M\), where
\begin{equation}\label{eq:entropy-factor-cover-sizes}
 \begin{aligned}
 K_V&=\mathcal N(\epsilon/2,\cV_{j,N},\|\cdot\|_\infty),\\
 K_H&=\mathcal N(\epsilon/(2L_{u,j}),\cH_M,\|\cdot\|_\infty).
 \end{aligned}
\end{equation}
For any pair \((v,h)\in\cV_{j,N}\times\cH_M\), select indices
\(a,b\) with
\(\|v-v^{(a)}\|_\infty\le\epsilon/2\) and
\(\|h-h^{(b)}\|_\infty\le\epsilon/(2L_{u,j})\).
For each field input \((x,z,\tau)\), where \(z=(z_0,z_c)\),
Equation~\eqref{eq:field} gives the exact decomposition
\begin{equation}\label{eq:entropy-composition-decomposition}
 \begin{aligned}
 &F_{j,v,h}(x,z,\tau)-F_{j,v^{(a)},h^{(b)}}(x,z,\tau)\\
 &\quad=\{v(x,h(z_0),z_c,\tau)
                   -v^{(a)}(x,h(z_0),z_c,\tau)\}\\
 &\qquad\quad+\{v^{(a)}(x,h(z_0),z_c,\tau)
                   -v^{(a)}(x,h^{(b)}(z_0),z_c,\tau)\}.
 \end{aligned}
\end{equation}
The first brace in
Equation~\eqref{eq:entropy-composition-decomposition} has Euclidean
norm at most \(\|v-v^{(a)}\|_\infty\). The second brace has norm
at most \(L_{u,j}\|h(z_0)-h^{(b)}(z_0)\|\), because the proper
center \(v^{(a)}\in\cV_{j,N}\) satisfies the global blockwise bound
\(\Lip_u(v^{(a)})\le L_{u,j}\) in
Equation~\eqref{eq:lipschitz}. Taking the supremum over the field
inputs in Equation~\eqref{eq:entropy-composition-decomposition} gives
\begin{equation}\label{eq:entropy-composition-radius}
 \begin{aligned}
 \|F_{j,v,h}-F_{j,v^{(a)},h^{(b)}}\|_\infty
 &\le\|v-v^{(a)}\|_\infty
                    +L_{u,j}\|h-h^{(b)}\|_\infty\\
 &\le\epsilon/2+L_{u,j}\epsilon/(2L_{u,j})=\epsilon.
 \end{aligned}
\end{equation}
Every composed center \(F_{j,v^{(a)},h^{(b)}}\) belongs to
\(\cF_j(N,M)\). Thus these centers form a proper cover of size at
most \(K_VK_H\), and
\begin{equation}\label{eq:entropy-composition-count}
 \begin{aligned}
 H_{\cF_j(N,M)}(\epsilon)
 &\le\log(K_VK_H)=\log K_V+\log K_H\\
 &=H_{\cV_{j,N}}(\epsilon/2)
       +H_{\cH_M}(\epsilon/(2L_{u,j})).
 \end{aligned}
\end{equation}
This is Equation~\eqref{eq:composition-entropy}. The Lipschitz bound
used here is the representation-input bound of the head \(v^{(a)}\);
no Lipschitz bound on the fitted representation \(h\) is required.

\medskip\noindent
\textbf{Step 6: Frozen representations \(h\) and source conditioning.}
Fix a bounded representation \(h\), and put
\(\mathcal F_h=\{F_{j,v,h}:v\in\cV_{j,N}\}\).
For any two heads \(v,v'\in\cV_{j,N}\), their representation
argument is identical, so Equation~\eqref{eq:field} yields
\begin{equation}\label{eq:entropy-frozen-contraction}
 \begin{aligned}
 \|F_{j,v,h}-F_{j,v',h}\|_\infty
 &=\sup_{x,z,\tau}
     \|v(x,h(z_0),z_c,\tau)-v'(x,h(z_0),z_c,\tau)\|\\
 &\le\|v-v'\|_\infty.
 \end{aligned}
\end{equation}
Consequently, the images of any proper \(\epsilon\)-cover of
\(\cV_{j,N}\) under the map \(v\mapsto F_{j,v,h}\) form a proper
\(\epsilon\)-cover of \(\mathcal F_h\). In particular,
\begin{equation}\label{eq:entropy-frozen-cover-count}
 \mathcal N(\epsilon,\mathcal F_h,\|\cdot\|_\infty)
 \le\mathcal N(\epsilon,\cV_{j,N},\|\cdot\|_\infty).
\end{equation}
Taking logarithms in Equation~\eqref{eq:entropy-frozen-cover-count}
proves Equation~\eqref{eq:frozen-entropy}. The bound contains no
cover of \(h\), and hence no dependence on the number or values of
parameters used to realize \(h\).

For the source-trained case, take \(j=t\) and let \(\cS\) be the
source-and-anchor sigma-field defined in Section~\ref{sec:adaptation}.
The fitted representation \(\widehat h_s\) is \(\cS\)-measurable,
and \(\cS\) is independent of the target training sample. Choose
a deterministic proper head cover
\(\{v^{(a)}:1\le a\le K_\epsilon\}\), where
\(K_\epsilon=\mathcal N(\epsilon,\cV_{t,N},\|\cdot\|_\infty)\).
For each realization \(\omega\) of the source information, the
explicit center set
\begin{equation}\label{eq:entropy-source-conditional-centers}
 \mathcal C_\epsilon(\omega)
 =\{F_{t,v^{(a)},\widehat h_s(\omega)}:
                                   1\le a\le K_\epsilon\}
\end{equation}
is a proper \(\epsilon\)-cover of
\(\mathcal F_{\widehat h_s(\omega)}\), by
Equation~\eqref{eq:entropy-frozen-contraction}. These centers are
\(\cS\)-measurable at every field input because the head centers
\(v^{(a)}\) are deterministic continuous functions and
\(\widehat h_s\) has the measurable version stipulated in
Section~\ref{sec:model}. Therefore no additional measurable selection
of a source-dependent parameter cover is needed. The bound
\(H_{\mathcal F_{\widehat h_s(\omega)}}(\epsilon)
\le H_{\cV_{t,N}}(\epsilon)\) holds for every admissible realization
\(\omega\), with the same constant. Independence from target training
is not needed for the deterministic inequality
\eqref{eq:entropy-frozen-contraction}; it ensures that conditioning
on \(\cS\) fixes this cover without changing the target sampling
law when the cover is subsequently used in
Lemma~\ref{lem:relative}.
\end{proof}

\section{Regression bounds and main theorem proofs}
\label{app:mainproofs}
\subsection{A bounded relative-deviation inequality}
For an observation \(O=(W,U)\), let \(Pq=\E q(O)\) and
\(P_nq=n^{-1}\sum_{i=1}^nq(O_i)\). For a vector-valued predictor \(f\),
write \(\ell_f(O)=\|f(W)-U\|^2\).

\begin{lemma}[Relative deviations, least squares, and auxiliary clusters]
\label{lem:relative}
Let \((W_i,U_i)\), \(1\le i\le n\), be independent observations with
common law, \(\|U_i\|\le B_U\), and let
\(f_0(W)=\E(U\mid W)\). Let \(\mathcal F\) be a nonempty class of
predictors bounded by \(B_V\), fixed independently of these observations.
Set \(K=B_U+B_V\), \(K_0=1+B_U+B_V\),
\(s(f)=\|f-f_0\|_{L_2(P_W)}^2\), and
\(D_f=\ell_f-\ell_{f_0}\). For every \(x\ge0\), with probability at
least \(1-e^{-x}\), simultaneously for all \(f\in\mathcal F\),
\begin{equation}\label{eq:relative}
 |(P_n-P)D_f|\le\tfrac14s(f)+b_n(x),\qquad
 b_n(x)=C K_0^2\frac{H_{\mathcal F}(n^{-2})+x+1}{n}.
\end{equation}
Consequently any measurable \(\epsilon\)-minimizer of \(P_n\ell_f\)
over \(\mathcal F\) satisfies, on the same event,
\begin{equation}\label{eq:erm-oracle}
 s(\widehat f)\le\tfrac53\inf_{f\in\mathcal F}s(f)
                  +\tfrac83 b_n(x)+\tfrac43\epsilon.
\end{equation}
In expectation the same inequality holds with
\(b_n(x)\) replaced by
\(C K_0^2\{H_{\mathcal F}(n^{-2})+1\}/n\).
The results hold conditionally on independent information fixing the class.
They also hold when each original observation has a fixed number of
conditionally independent auxiliary draws and its loss is their average,
or when the loss is integrated over the auxiliary draws exactly. In these
cases \(n\) counts original observations.
\end{lemma}
\begin{proof}\proofanchor{lem:relative}
Throughout this proof, the population operator \(P\) refers to a fresh
observation \(O=(W,U)\) with the common law of the observations
\(O_i=(W_i,U_i)\). The predictor class \(\mathcal F\) is initially
regarded as deterministic. If
\(H_{\mathcal F}(n^{-2})=\infty\), the asserted upper bounds are
vacuous, so suppose that this covering entropy is finite.

\paragraph{Step 1: Mean, envelope, and second moment of the excess loss
\(D_f\).}
Conditional Jensen's inequality gives, for \(P_W\)-almost every \(w\),
\begin{equation}\label{eq:relative-proof-mean-envelope}
 \|f_0(w)\|=\|\E(U\mid W=w)\|
 \le\E(\|U\|\mid W=w)\le B_U.
\end{equation}
Fix \(f\in\mathcal F\), and write
\(a_f(W)=f(W)-f_0(W)\). Then
\(\|a_f(W)\|\le B_V+B_U=K\) almost surely. Expanding the two
squared losses in \(D_f=\ell_f-\ell_{f_0}\) gives
\begin{align}
 D_f(O)
 &=\|a_f(W)\|^2
       +2\ip{a_f(W)}{f_0(W)-U},
                  \label{eq:relative-proof-expansion}\\*
 PD_f
 &=\E\|a_f(W)\|^2
   +2\E\ip{a_f(W)}{\E(f_0(W)-U\mid W)}
 =s(f).           \label{eq:relative-proof-mean}
\end{align}
Because \(\|f_0(W)-U\|\le2B_U\),
\cref{eq:relative-proof-expansion} also implies
\begin{align}
 |D_f(O)|
 &\le\|a_f(W)\|\{\|a_f(W)\|+4B_U\}
 \le5K\|a_f(W)\|\le5K^2,
                     \label{eq:relative-proof-envelope}\\*
 PD_f^2&\le25K^2\E\|a_f(W)\|^2=25K^2s(f).
                     \label{eq:relative-proof-second-moment}
\end{align}
In particular, the centered variable
\(Y_{i,f}=D_f(O_i)-PD_f\) satisfies
\begin{equation}\label{eq:relative-proof-centered-bounds}
 \E Y_{i,f}=0,\qquad |Y_{i,f}|\le10K^2,\qquad
 \E Y_{i,f}^2=\Var\{D_f(O_i)\}\le25K^2s(f).
\end{equation}
When \(K=0\), both \(U\) and every predictor \(f\) vanish almost
surely, and all the assertions follow directly. Hence the subsequent
concentration calculation may assume \(K>0\).

\paragraph{Step 2: A scalar concentration calculation for a fixed
predictor \(f\).}
Let \(Y\) be any centered real random variable satisfying
\(|Y|\le b\) and \(\E Y^2\le\sigma^2\). For \(k\ge2\),
\(|\E Y^k|\le b^{k-2}\sigma^2\). The factorial inequality
\(k!\ge2\cdot3^{k-2}\) follows by induction from its equality at
\(k=2\). Consequently, for \(|t|b<3\), the absolutely convergent
exponential series yields
\begin{align}
 \E e^{tY}
 &\le1+\sum_{k=2}^{\infty}
                \frac{|t|^k b^{k-2}\sigma^2}{k!}
 \le1+\frac{t^2\sigma^2}{2}
                 \sum_{k=2}^{\infty}(|t|b/3)^{k-2}
 \nonumber\\*
 &\le\exp\left\{\frac{t^2\sigma^2}{2(1-|t|b/3)}\right\}.
                   \label{eq:relative-proof-mgf}
\end{align}
For independent centered variables \(Y_1,\ldots,Y_n\) with these
same bounds, independence and \cref{eq:relative-proof-mgf} imply
\[
 \E\exp\left(t\sum_{i=1}^nY_i\right)
 \le\exp\left\{\frac{n\sigma^2t^2}{2(1-bt/3)}\right\},
 \qquad 0<t<3/b.
\]
To make the exponential-tail calculation explicit, put
\(V=n\sigma^2\), \(c=b/3\), and, for \(V>0\) and \(u>0\),
\(\alpha=\sqrt{2u/V}\) and \(t=\alpha/(1+c\alpha)\).
Markov's inequality then gives
\begin{align*}
 \PP\left\{\sum_{i=1}^nY_i\ge\sqrt{2Vu}+cu\right\}
 &\le\exp\left\{-t(\sqrt{2Vu}+cu)
                   +\frac{Vt^2}{2(1-ct)}\right\}\\*
 &=\exp\left\{-\frac{2u+c\alpha u}{1+c\alpha}
                         +\frac{u}{1+c\alpha}\right\}=e^{-u}.
\end{align*}
The case \(V=0\) has \(Y_i=0\) almost surely. Applying the same
argument to \(-Y_i\), and enlarging the linear term harmlessly,
proves the two-sided inequality
\begin{equation}\label{eq:relative-proof-bernstein}
 \PP\left\{\left|\frac1n\sum_{i=1}^nY_i\right|
       >\sqrt{\frac{2\sigma^2u}{n}}+\frac{2bu}{3n}\right\}
 \le2e^{-u},\qquad u\ge0.
\end{equation}
Use \cref{eq:relative-proof-centered-bounds} in
\cref{eq:relative-proof-bernstein}, with \(b=10K^2\) and
\(\sigma^2=25K^2s(f)\). The elementary inequality
\(AB\le A^2/4+B^2\), applied with
\(A=\sqrt{s(f)}\) and \(B=\sqrt{50K^2u/n}\), gives
\[
 \sqrt{50K^2s(f)u/n}
 \le\tfrac14s(f)+50K^2u/n.
\]
It follows that, for a universal constant \(C\),
\begin{equation}\label{eq:relative-proof-fixed-tail}
 \PP\left\{|(P_n-P)D_f|>\tfrac14s(f)+CK^2u/n\right\}
 \le2e^{-u}.
\end{equation}

\paragraph{Step 3: A proper cover and extension to every predictor
\(f\in\mathcal F\).}
Set \(\delta_n=n^{-2}\), and choose a proper supremum-norm
\(\delta_n\)-cover \(\{g_1,\ldots,g_{M_n}\}\subseteq\mathcal F\),
where \(\log M_n=H_{\mathcal F}(\delta_n)\). Equivalently, one may
use a cover within a fixed factor of the minimum cardinality; its extra
logarithm is absorbed by the constant below. For
\(u=x+\log(2M_n)\), the union bound and
\cref{eq:relative-proof-fixed-tail} show that, with probability at
least \(1-e^{-x}\),
\begin{equation}\label{eq:relative-proof-net-event}
 |(P_n-P)D_{g_k}|
 \le\tfrac14s(g_k)+CK^2\{x+\log(2M_n)\}/n,
 \qquad 1\le k\le M_n.
\end{equation}
For an arbitrary \(f\in\mathcal F\), choose a center \(g_k\) such
that \(\|f-g_k\|_\infty\le\delta_n\). The identity
\(\|a\|^2-\|b\|^2=\ip{a-b}{a+b}\) implies
\begin{align}
 |\ell_f(O)-\ell_{g_k}(O)|
 &\le\|f(W)-g_k(W)\|
             \{\|f(W)-U\|+\|g_k(W)-U\|\}
 \le2K\delta_n,\nonumber\\*
 |s(f)-s(g_k)|
 &\le\E\big[\|f(W)-g_k(W)\|
       \{\|f(W)-f_0(W)\|+\|g_k(W)-f_0(W)\|\}\big]
 \le2K\delta_n.
                     \label{eq:relative-proof-net-continuity}
\end{align}
In particular,
\(|(P_n-P)(D_f-D_{g_k})|\le4K\delta_n\).
Combining this inequality with
\cref{eq:relative-proof-net-event,eq:relative-proof-net-continuity}
gives, on the same event and for every \(f\in\mathcal F\),
\begin{align*}
 |(P_n-P)D_f|
 &\le\tfrac14s(f)+CK^2\{x+\log(2M_n)\}/n
                         +\tfrac92K\delta_n\\*
 &\le\tfrac14s(f)
       +CK_0^2\{H_{\mathcal F}(n^{-2})+x+1\}/n.
\end{align*}
The last inequality uses \(K_0=1+K\), \(\delta_n=n^{-2}\le n^{-1}\),
and \(K\le K_0^2\). This is \cref{eq:relative}. The measurable,
separable versions stipulated after \cref{eq:empirical} ensure that
these simultaneous statements and the resulting fitted risks are
measurable.

\paragraph{Step 4: The empirical minimization inequality and its
expectation.}
Let \(\widehat f\) be the stated \(\epsilon\)-minimizer. Subtracting
\(P_n\ell_{f_0}\) from the empirical minimization inequality gives
\(P_nD_{\widehat f}\le P_nD_f+\epsilon\) for every
\(f\in\mathcal F\). On the event in \cref{eq:relative}, use
\(PD_f=s(f)\) from \cref{eq:relative-proof-mean} on both sides:
\begin{equation}\label{eq:relative-proof-basic-inequality}
 \tfrac34s(\widehat f)-b_n(x)
 \le P_nD_{\widehat f}
 \le P_nD_f+\epsilon
 \le\tfrac54s(f)+b_n(x)+\epsilon.
\end{equation}
Multiplication by \(4/3\), followed by the infimum over
\(f\in\mathcal F\), proves \cref{eq:erm-oracle}. No attainment of
the infimum \(s_*:=\inf_{f\in\mathcal F}s(f)\) is required, because
\cref{eq:relative-proof-basic-inequality} holds for every comparator
\(f\).

For the expectation assertion, write
\(H_n=H_{\mathcal F}(n^{-2})\), set
\(d_n=(8/3)CK_0^2/n\), and define the nonnegative variable
\[
 Z_n=\big\{s(\widehat f)-\tfrac53s_*
                                  -\tfrac43\epsilon\big\}_+,
 \qquad r_+=\max(r,0).
\]
The probability bound already proved states that
\(\PP\{Z_n>d_n(H_n+1+x)\}\le e^{-x}\) for all \(x\ge0\).
Therefore the tail-integral identity for a nonnegative variable gives
\begin{align}
 \E Z_n
 &=\int_0^\infty\PP(Z_n>u)\dd u\nonumber\\*
 &\le d_n(H_n+1)
       +d_n\int_0^\infty
           \PP\{Z_n>d_n(H_n+1+x)\}\dd x
 \le d_n(H_n+2).
                         \label{eq:relative-proof-tail-integral}
\end{align}
Since \(H_n+2\le2(H_n+1)\), increasing the universal constant in
\(b_n\) gives precisely the claimed expectation form of
\cref{eq:erm-oracle}.

\paragraph{Step 5: Conditioning on independent information and
averaging auxiliary draws.}
Let \(\mathcal B\) be a sigma-field independent of the observations
\(O_1,\ldots,O_n\), and suppose that \(\mathcal B\) fixes the
predictor class \(\mathcal F\). Conditional on almost every
realization of \(\mathcal B\), the observations retain their product
law, the mean \(f_0\) is unchanged, and the preceding proof applies
to the now fixed class \(\mathcal F\). In particular,
\cref{eq:relative-proof-tail-integral} can be integrated conditionally
on \(\mathcal B\). This establishes both conditional assertions.

For the cluster assertion, let \(V_i\) denote original observation
\(i\), and let \(O_{i,k}\), \(1\le k\le m\), be its regression
observations obtained from the \(m\) auxiliary draws. Each
\(O_{i,k}\) has the same marginal law as \(O\); the vectors
\((V_i,O_{i,1},\ldots,O_{i,m})\) are independent over \(i\).
Define the cluster excess loss
\(\overline D_{f,i}=m^{-1}\sum_{k=1}^mD_f(O_{i,k})\). Equations
\eqref{eq:relative-proof-mean}--\eqref{eq:relative-proof-second-moment}
and the inequality \((m^{-1}\sum_k d_k)^2\le m^{-1}\sum_kd_k^2\)
give
\begin{equation}\label{eq:relative-proof-cluster-moments}
 \E\overline D_{f,i}=s(f),\qquad
 |\overline D_{f,i}|\le5K^2,\qquad
 \E\overline D_{f,i}^{\,2}
 \le\frac1m\sum_{k=1}^m\E D_f(O_{i,k})^2
 \le25K^2s(f).
\end{equation}
The exact auxiliary-integrated excess loss is
\(D_{f,i}^{\rm int}=\E\{D_f(O_{i,1})\mid V_i\}\).
The tower property and conditional Jensen's inequality give the same
three bounds in \cref{eq:relative-proof-cluster-moments} for
\(D_{f,i}^{\rm int}\). The loss differences for either cluster
average also retain the bound \(2K\delta_n\) in
\cref{eq:relative-proof-net-continuity}, by averaging or conditional
expectation. Thus all concentration, covering, and minimization steps
apply to the \(n\) independent cluster losses. None of these steps
treats the \(nm\) auxiliary regression observations as independent;
the denominator in \cref{eq:relative} remains \(n\).
\end{proof}

\subsection{Joint neural learning and the frozen comparator}
\begin{lemma}[Joint neural regression and source deficiency]\label{lem:joint}
For task \(j\), suppose its bounded-regression part of \cref{ass:path}
and its part of \cref{ass:smooth} hold. Let
\((\widehat v_j,\widehat h_j)\) be an \(\epsilon_j\)-minimizer of the
empirical loss over \(\cV_{j,N}\times\cH_M\), with \(N,M\) at most
polynomial in \(n_j\) and above the fixed approximation threshold. Then
\begin{align}\label{eq:joint-risk}
 \E\cE_j(F_{j,\widehat v_j,\widehat h_j})
 \le C\left\{\zeta_j^2+N^{-2r_j/D_j}+M^{-2\beta_j/p}
          +\frac{(N+M)\log^3(en_j)}{n_j}+\epsilon_j\right\}.
\end{align}
Also \(\E\cD_j(\widehat h_j)\) is at most the left side of
\cref{eq:joint-risk}.
\end{lemma}
\begin{proof}\proofanchor{lem:joint}
Fix the task index \(j\in\{s,t\}\), and let
\(\cF_j(N,M)=\{F_{j,v,h}:v\in\cV_{j,N},\ h\in\cH_M\}\)
be the joint field class in \cref{lem:entropy}. Its population norm
is \(\|\cdot\|_j=\|\cdot\|_{L_2(\nu_j)}\), and its regression
mean is \(v_j^0\) from \cref{eq:full-mean}.

\paragraph{Step 1: Construct a jointly feasible comparator
\((v_{j,N},h_M)\).}
Apply \cref{lem:approximation} to the functions \(h_{j,0}\) and
\(g_{j,0}\) in \cref{ass:smooth}. There are comparators
\(h_M\in\cH_M\) and \(v_{j,N}\in\cV_{j,N}\) satisfying
\begin{equation}\label{eq:joint-proof-approximation}
 \|h_M-h_{j,0}\|_\infty\le C M^{-\beta_j/p},\qquad
 \|v_{j,N}-g_{j,0}\|_{\infty,\cK_j}\le C N^{-r_j/D_j}.
\end{equation}
Here feasibility of the head \(v_{j,N}\), not merely its value
approximation, follows from
\cref{eq:derivative-approximation,eq:smooth-margins}.
Indeed, with
\(\eta_{\min,j}=\min_{b\in\{x,u,\tau\}}\eta_{b,j}>0\),
the approximation threshold can be chosen so that
\(CN^{-(r_j-1)/D_j}\le\eta_{\min,j}\). The construction in
\cref{lem:approximation}, including the nonexpansive input and output
clips, then gives
\(\Lip_b(v_{j,N})\le\Lip_b(g_{j,0};\cK_j)
       +CN^{-(r_j-1)/D_j}\le L_{b,j}\)
for each \(b\in\{x,u,\tau\}\). For the encoder \(h_M\), the
value approximation in \cref{eq:joint-proof-approximation} suffices;
no encoder derivative bound is needed.

\paragraph{Step 2: Bound the comparator's population excess risk.}
For a raw input \(w=(x,z,\tau)\), where \(z=(z_0,z_c)\), define
\[
 q_M(w)=(x,h_M(z_0),z_c,\tau),\qquad
 q_0(w)=(x,h_{j,0}(z_0),z_c,\tau).
\]
The support condition in \cref{ass:path} and the bounded ranges of
\(h_M,h_{j,0}\) imply \(q_M(W_j),q_0(W_j)\in\cK_j\) almost
surely. Define the following three vector-valued functions on that
support:
\begin{align}
 e_1(w)&=v_{j,N}(q_M(w))-g_{j,0}(q_M(w)),\nonumber\\*
 e_2(w)&=g_{j,0}(q_M(w))-g_{j,0}(q_0(w)),\nonumber\\*
 e_3(w)&=g_{j,0}(q_0(w))-v_j^0(w).
                         \label{eq:joint-proof-error-terms}
\end{align}
By \cref{eq:field,eq:joint-proof-error-terms},
\(F_{j,v_{j,N},h_M}-v_j^0=e_1+e_2+e_3\). Equation
\eqref{eq:joint-proof-approximation}, the representation-block bound
\(\Lip_u(g_{j,0};\cK_j)\le L_{u,j}\) in
\cref{eq:smooth-margins}, and \cref{eq:structural-bias} yield
\begin{equation}\label{eq:joint-proof-error-norms}
 \|e_1\|_j^2\le C N^{-2r_j/D_j},\qquad
 \|e_2\|_j^2\le L_{u,j}^2\|h_M-h_{j,0}\|_\infty^2
                       \le C M^{-2\beta_j/p},\qquad
 \|e_3\|_j^2=\zeta_j^2.
\end{equation}
For any vectors \(u_1,u_2,u_3\), the Cauchy--Schwarz inequality
implies \(\|u_1+u_2+u_3\|^2\le3\sum_{k=1}^3\|u_k\|^2\).
Integrating this inequality and using
\cref{eq:joint-proof-error-norms} gives
\begin{align}
 \inf_{F\in\cF_j(N,M)}\cE_j(F)
 &\le\cE_j(F_{j,v_{j,N},h_M})
 \le3\sum_{k=1}^3\|e_k\|_j^2\nonumber\\*
 &\le C\{\zeta_j^2+N^{-2r_j/D_j}+M^{-2\beta_j/p}\}.
                         \label{eq:joint-proof-comparator-risk}
\end{align}

\paragraph{Step 3: Count both fitted networks and apply the regression
oracle.}
The proper composition cover in \cref{eq:composition-entropy} gives
\begin{align}
 H_{\cF_j(N,M)}(n_j^{-2})
 &\le H_{\cV_{j,N}}(n_j^{-2}/2)
       +H_{\cH_M}(n_j^{-2}/(2L_{u,j}))\nonumber\\*
 &\le C(N+M)\log^3(en_j).
                         \label{eq:joint-proof-entropy}
\end{align}
The last inequality uses \cref{eq:network-entropy,eq:sieve-entropy},
the fixed positive constant \(L_{u,j}\), and the assumed
polynomial-growth bounds on \(N,M\). Multiplying the covering radius
by a fixed positive constant changes only the constant in this bound.
Every \(F\in\cF_j(N,M)\) is bounded by \(B_{V,j}\), whereas the
regression label \(U_{j,T}\) is bounded by \(B_{U,j}\). The class
\(\cF_j(N,M)\) is fixed before the task-\(j\) observations are
used. Thus the expectation version of \cref{eq:erm-oracle}, with
\(f_0=v_j^0\), applies to the fitted field
\(\widehat F_j=F_{j,\widehat v_j,\widehat h_j}\):
\begin{align}
 \E\cE_j(\widehat F_j)
 &\le\tfrac53\inf_{F\in\cF_j(N,M)}\cE_j(F)
   +C(1+B_{U,j}+B_{V,j})^2
       \frac{H_{\cF_j(N,M)}(n_j^{-2})+1}{n_j}
   +\tfrac43\epsilon_j. \label{eq:joint-proof-regression}
\end{align}
Substitute \cref{eq:joint-proof-comparator-risk,eq:joint-proof-entropy}
into \cref{eq:joint-proof-regression}. Since \(N,M\ge2\) and
\(\log(en_j)\ge1\), the additional term \(n_j^{-1}\) is absorbed
by \((N+M)\log^3(en_j)/n_j\). This proves \cref{eq:joint-risk}.
The cluster version of \cref{lem:relative} proves the same conclusion
when the empirical loss averages or integrates auxiliary draws.

\paragraph{Step 4: Pass from prediction risk to representation deficiency.}
For every fixed feasible pair \((v,h)\), the projection identity
\cref{eq:decomposition} gives
\[
 \cE_j(F_{j,v,h})
 =\|F_{j,v,h}-m_{j,h}\|_j^2+\cD_j(h)
 \ge\cD_j(h).
\]
For each realized training sample, apply this deterministic identity to
\((v,h)=(\widehat v_j,\widehat h_j)\), with a fresh evaluation
observation defining the norm \(\|\cdot\|_j\). Taking expectations
then gives
\(\E\cD_j(\widehat h_j)\le\E\cE_j(\widehat F_j)\).
This substitution does not condition the training observations on the
fitted encoder \(\widehat h_j\); it uses the identity after training
has fixed the functions being evaluated.
\end{proof}

\begin{lemma}[Expected profile at the source representation]
\label{lem:profile-bound}
Under the target bounded-label condition and \cref{ass:transfer}, for
all sufficiently large \(N\),
\begin{equation}\label{eq:profile-source}
 \E Q_{t,N}(\widehat h_s)
 \le C N^{-2r_t/D_t}+\kappa\E\cD_s(\widehat h_s)
                      +\Delta^2+B_{U,t}^2\pi_s.
\end{equation}
No smoothness condition on a target-fitted representation is needed.
\end{lemma}
\begin{proof}\proofanchor{lem:profile-bound}
Let \(E_s=\{\widehat h_s\in\cH_{\rm good}\}\). By
\cref{ass:transfer}, the event \(E_s\) is measurable with respect to
source training and \(\PP(E_s^c)\le\pi_s\). We first prove a
pointwise profile bound for each deterministic representation
\(h\in\cH_{\rm good}\), and only then substitute \(\widehat h_s\).

\paragraph{Step 1: Approximate the reduced conditional mean at a good
representation \(h\).}
For \(h\in\cH_{\rm good}\), let \(g_{t,h}\) be the version
specified in \cref{ass:transfer}. For the reduced input
\(q_h(W_t)=(S_{t,T},h(Z_{t,0}),Z_{t,c},T)\), the version identity
in \cref{eq:reduced-mean} is
\(m_{t,h}(W_t)=g_{t,h}(q_h(W_t))\) almost surely. The smoothness
bound \(B_{g,t}\) and the three positive margins in
\cref{eq:reduced-regularity} do not depend on \(h\). Hence
\cref{lem:approximation} supplies, for every sufficiently large \(N\),
a feasible head \(v_{N,h}\in\cV_{t,N}\) such that
\begin{equation}\label{eq:profile-proof-uniform-approximation}
 \|v_{N,h}-g_{t,h}\|_{\infty,\cK_t}
 \le C N^{-r_t/D_t},
 \qquad h\in\cH_{\rm good}.
\end{equation}
Both the constant \(C\) and the lower threshold for \(N\) are
uniform in \(h\). In particular, the derivative approximation in
\cref{eq:derivative-approximation} fits inside each of the fixed
margins in \cref{eq:reduced-regularity}; the comparator head
\(v_{N,h}\) therefore satisfies the actual constraints in
\cref{eq:lipschitz}.

Using \cref{eq:profile,eq:profile-proof-uniform-approximation} and
the reduced-input version on \(\cK_t\), we obtain
\begin{align}
 \cA_{t,N}(h)
 &\le\|F_{t,v_{N,h},h}-m_{t,h}\|_t^2\nonumber\\*
 &=\E\|v_{N,h}(q_h(W_t))-g_{t,h}(q_h(W_t))\|^2
 \le C N^{-2r_t/D_t}.
                       \label{eq:profile-proof-head-error}
\end{align}
The profile identity \cref{eq:decomposition} and the calibration
inequality \cref{eq:calibration} now imply
\begin{equation}\label{eq:profile-proof-good-bound}
 Q_{t,N}(h)
 =\cD_t(h)+\cA_{t,N}(h)
 \le\kappa\cD_s(h)+\Delta^2+C N^{-2r_t/D_t},
 \qquad h\in\cH_{\rm good}.
\end{equation}
Only existence of each comparator head \(v_{N,h}\) is used in the
pointwise infimum defining \(Q_{t,N}(h)\); this step does not require
choosing a new target-trained representation or a measurable family of
minimizing heads.

\paragraph{Step 2: Control the profile outside the good set.}
The zero head \(v\equiv0\) belongs to \(\cV_{t,N}\) by the
sieve definition in \cref{sec:sieves}. Thus, for every feasible
representation \(h\), including those outside \(\cH_{\rm good}\),
\begin{align}
 0\le Q_{t,N}(h)
 &\le\cE_t(F_{t,0,h})
 =\|v_t^0\|_t^2\nonumber\\*
 &=\E\|\E(U_{t,T}\mid W_t)\|^2
 \le\E\|U_{t,T}\|^2\le B_{U,t}^2.
                       \label{eq:profile-proof-bad-bound}
\end{align}
The last two inequalities are conditional Jensen's inequality and the
bounded-label condition. No reduced-mean approximation is invoked on
\(E_s^c\).

\paragraph{Step 3: Average over the source-trained representation
\(\widehat h_s\).}
Substitute \(h=\widehat h_s\) in
\cref{eq:profile-proof-good-bound,eq:profile-proof-bad-bound}, multiply
by the respective event indicators, and take expectations:
\begin{align*}
 \E Q_{t,N}(\widehat h_s)
 &=\E\{\ind(E_s)Q_{t,N}(\widehat h_s)\}
   +\E\{\ind(E_s^c)Q_{t,N}(\widehat h_s)\}\\*
 &\le(CN^{-2r_t/D_t}+\Delta^2)\PP(E_s)
   +\kappa\E\{\ind(E_s)\cD_s(\widehat h_s)\}
   +B_{U,t}^2\PP(E_s^c)\\*
 &\le CN^{-2r_t/D_t}+\kappa\E\cD_s(\widehat h_s)
                   +\Delta^2+B_{U,t}^2\pi_s.
\end{align*}
The final inequality uses \(\kappa\ge0\),
\(\cD_s(\widehat h_s)\ge0\), and \(\PP(E_s^c)\le\pi_s\),
not independence between \(E_s\) and the source deficiency. This is
\cref{eq:profile-source}.
\end{proof}

\subsection{Proof of the frozen theorem}
\begin{proof}[Proof of \cref{thm:freeze}]\proofanchor{thm:freeze}
We first verify the deterministic rate calculation in \cref{eq:rho},
then analyze the source fit and the conditional target fit separately.

\paragraph{Step 1: Verify the resolution balance, including integer
rounding and the optimization tolerance.}
Fix \(a>0\) and an integer \(D\ge1\). Write
\(x_n=n/\ell_n^3\), \(\theta=D/(2a+D)\), and
\(\gamma=2a/(2a+D)=1-\theta\). Since \(x_n\to\infty\),
we have \(x_n^\theta\ge2\) for all sufficiently large \(n\).
For these \(n\), the definition of \(K_{a,D}(n)\) gives
\[
 x_n^\theta\le K_{a,D}(n)=\lceil x_n^\theta\rceil
          \le x_n^\theta+1\le2x_n^\theta.
\]
Consequently,
\begin{align*}
 K_{a,D}(n)^{-2a/D}&\le x_n^{-2a/(2a+D)}=\rho_{a,D}(n),\\*
 \frac{K_{a,D}(n)\ell_n^3}{n}
       &=\frac{K_{a,D}(n)}{x_n}
         \le2x_n^{\theta-1}=2\rho_{a,D}(n),\\*
 n^{-2}&\le n^{-\gamma}
          \le(\ell_n^3/n)^\gamma=\rho_{a,D}(n).
\end{align*}
The last line uses \(0<\gamma<1\) and \(\ell_n\ge1\). We have
therefore proved
\begin{equation}\label{eq:balance}
 K_{a,D}(n)^{-2a/D}
       +\frac{K_{a,D}(n)\ell_n^3}{n}+n^{-2}
       \le C_{a,D}\rho_{a,D}(n).
\end{equation}
Moreover, \(K_{a,D}(n)\le2n^\theta\) for sufficiently large \(n\),
so these resolutions satisfy the polynomial-growth requirements in
\cref{lem:entropy,lem:joint}. They also eventually exceed every fixed
approximation threshold required by \cref{lem:approximation}.

\paragraph{Step 2: Derive the source deficiency bound from joint source
regression.}
The source estimator \((\widehat v_s,\widehat h_s)\) in
\cref{eq:source-estimator} is an \(n_s^{-2}\)-minimizer over
\(\cV_{s,N_s}\times\cH_{M_s}\). The source parts of
\cref{ass:path,ass:smooth} therefore permit the application of
\cref{lem:joint}, which gives
\begin{align}
 \E\cD_s(\widehat h_s)
 &\le\E\cE_s(F_{s,\widehat v_s,\widehat h_s})\nonumber\\*
 &\le C\left\{\zeta_s^2+N_s^{-2r_s/D_s}+M_s^{-2\beta_s/p}
       +\frac{N_s\ell_{n_s}^3}{n_s}
       +\frac{M_s\ell_{n_s}^3}{n_s}+n_s^{-2}\right\}.
                         \label{eq:freeze-proof-source-unbalanced}
\end{align}
Substitute \(N_s=K_{r_s,D_s}(n_s)\) and
\(M_s=K_{\beta_s,p}(n_s)\) from \cref{eq:resolution-choice}.
Applying \cref{eq:balance} to \((a,D)=(r_s,D_s)\) and
\((a,D)=(\beta_s,p)\) yields
\[
 \E\cD_s(\widehat h_s)
 \le C\{\zeta_s^2+\rho_{r_s,D_s}(n_s)+\rho_{\beta_s,p}(n_s)\}
 =C R_s(n_s),
\]
where \(R_s\) is defined in \cref{eq:source-rate-symbol}.
This proves \cref{eq:source-final} without assuming a parametric
source-learning rate.

\paragraph{Step 3: Analyze target fitting conditional on the source
sample.}
Let \(\mathcal S_s=\sigma(\sD_s,\text{source auxiliary draws})\).
The source-fitted encoder \(\widehat h_s\) is
\(\mathcal S_s\)-measurable. Independence in \cref{eq:data} and the
auxiliary sampling construction preceding \cref{eq:empirical} imply
that the target regression observations are independent of
\(\mathcal S_s\). Conditional on \(\mathcal S_s\), define the
fixed class
\[
 \cF_F=\{F_{t,v,\widehat h_s}:v\in\cV_{t,N_t}\}.
\]
Its predictors are bounded by \(B_{V,t}\), its regression mean is
still \(v_t^0\), and its population comparator risk is
\(\inf_{F\in\cF_F}\cE_t(F)=Q_{t,N_t}(\widehat h_s)\) by
\cref{eq:profile}. The frozen covering bound in
\cref{eq:frozen-entropy} gives, for almost every source realization,
\begin{equation}\label{eq:freeze-proof-conditional-entropy}
 H_{\cF_F}(n_t^{-2})
 \le H_{\cV_{t,N_t}}(n_t^{-2})
 \le C N_t\ell_{n_t}^3.
\end{equation}
There is no covering term for \(\widehat h_s\) in this conditional
class.

The minimization property \cref{eq:freeze-estimator} and the
conditional expectation version of \cref{eq:erm-oracle} now imply
\begin{equation}\label{eq:freeze-proof-conditional-risk}
 \E\{\cE_t(\widehat F_F)\mid\mathcal S_s\}
 \le\tfrac53 Q_{t,N_t}(\widehat h_s)
       +C\frac{N_t\ell_{n_t}^3}{n_t}+\tfrac43n_t^{-2}.
\end{equation}
The extra \(n_t^{-1}\) in the entropy oracle is absorbed because
\(N_t\ge2\) and \(\ell_{n_t}\ge1\). This conditioning does not
fix the target-fitted head \(\widehat v_F\); the regression oracle
controls its entire feasible class simultaneously.

\paragraph{Step 4: Average the target profile and balance its resolution.}
Take the expectation of \cref{eq:freeze-proof-conditional-risk} over
\(\mathcal S_s\), and apply \cref{eq:profile-source}. This gives
\begin{align}
 \E\cE_t(\widehat F_F)
 \le C\bigg\{&N_t^{-2r_t/D_t}
       +\frac{N_t\ell_{n_t}^3}{n_t}+n_t^{-2}
       +\kappa\E\cD_s(\widehat h_s)
       +\Delta^2+B_{U,t}^2\pi_s\bigg\}.
                         \label{eq:freeze-proof-target-unbalanced}
\end{align}
Using \cref{eq:source-final} and \cref{eq:balance} with
\((a,D,n)=(r_t,D_t,n_t)\), the right side of
\cref{eq:freeze-proof-target-unbalanced} is at most
\[
 C\{\rho_{r_t,D_t}(n_t)+\kappa R_s(n_s)
                                      +\Delta^2+B_{U,t}^2\pi_s\}
 =C\{\rho_{r_t,D_t}(n_t)+T_s(n_s)\}.
\]
The equality uses \cref{eq:transfer-rate-symbol}; this is the
claimed flow-matching bound \cref{eq:freeze-risk}.

\paragraph{Step 5: Convert the velocity risk to exact and numerical
endpoint risks.}
For any realized feasible pair \((v,h)\), the encoder \(h\) depends
only on context. Hence the composition \(F_{t,v,h}\) inherits the
head's output, state, and time bounds:
\begin{align}
 \|F_{t,v,h}(x,z,\tau)\|&\le B_{V,t},\nonumber\\*
 \|F_{t,v,h}(x,z,\tau)-F_{t,v,h}(x',z,\tau)\|
       &\le L_{x,t}\|x-x'\|,\nonumber\\*
 \|F_{t,v,h}(x,z,\tau)-F_{t,v,h}(x,z,\tau')\|
       &\le L_{\tau,t}|\tau-\tau'|.
                         \label{eq:main-proof-inherited-budgets}
\end{align}
These inequalities follow directly from \cref{eq:field,eq:lipschitz};
no Lipschitz bound on \(h\) as a function of context is needed.
Apply \cref{lem:stability} to \(F=\widehat F_F\), conditional on
all training randomness, and then take expectations:
\begin{align}
 \E\cW_t^2(P_{\widehat F_F},P_t^0)
 &\le C_{L_{x,t}}\E\cE_t(\widehat F_F),\nonumber\\*
 \E\cW_t^2(P_{\widehat F_F}^{[J]},P_t^0)
 &\le2C_{L_{x,t}}\E\cE_t(\widehat F_F)
                              +2C_{\rm Euler}J^{-2}.
                         \label{eq:freeze-proof-endpoint-conversion}
\end{align}
The constants in \cref{eq:stability-constant,eq:euler-constant} depend
only on fixed head budgets. Substituting \cref{eq:freeze-risk} proves
\cref{eq:freeze-main} and its exact-flow version.

Finally, if \(\zeta_s=\Delta=\pi_s=0\), then
\(T_s(n_s)=\kappa\{\rho_{r_s,D_s}(n_s)
                              +\rho_{\beta_s,p}(n_s)\}\).
Inserting this identity into \cref{eq:freeze-main} and replacing each
\(\rho_{a,D}(n)\) by its formula in \cref{eq:rho} gives exactly
\cref{eq:freeze-clean}.
\end{proof}

\subsection{Neural residual approximation and conditional complexity}
\begin{lemma}[Residual comparators and conditional neural entropy]
\label{lem:residual-neural}
Condition on \(\cS\). For the class in \cref{eq:residual-encoder},
let
\(\cF_A(N,M)=\{F_{t,v,h_{\widehat h_s,a}}:
                   v\in\cV_{t,N},\ a\in\mathcal R_{A,M}\}\).
For \(N,M\) bounded by fixed powers of \(n\),
\begin{equation}\label{eq:residual-entropy}
 H_{\cF_A(N,M)}(n^{-2})\le C(N+M)\log^3(en).
\end{equation}
The constant is independent of the architecture and fitted weights of
\(\widehat h_s\). On \(E_{s,A}=\{\widehat h_s\in\cH_{\rm good}^A\}\),
for sufficiently large \(N,M\), there are feasible comparators
\(v_N\in\cV_{t,N}\), \(a_M\in\mathcal R_{A,M}\) satisfying
\begin{align}
 \cE_t(F_{t,v_N,h_{\widehat h_s,a_M}})
 &\le C\{\kappa_A\cD_s(\widehat h_s)+\zeta_A^2
                    +N^{-2r_t/D_t}+M^{-2\beta_A/q_A}\},
                      \label{eq:residual-comparator}\\
 A_s(h_{\widehat h_s,a_M})
 &\le 2\delta_A^2+CM^{-2\beta_A/q_A}.
                      \label{eq:residual-comparator-anchor}
\end{align}
There is also a feasible frozen comparator, using the same \(v_N\), with
\begin{equation}\label{eq:correctable-frozen-profile}
 Q_{t,N}(\widehat h_s)
 \le C\{\kappa_A\cD_s(\widehat h_s)+\zeta_A^2
                         +N^{-2r_t/D_t}+L_{u,t}^2\delta_A^2\}
 \quad\hbox{on }E_{s,A}.
\end{equation}
\end{lemma}
\begin{proof}\proofanchor{lem:residual-neural}
Fix a realization of the target-independent sigma-field \(\cS\).
In this conditional argument the fitted source representation
\(\widehat h_s\) is a fixed function taking values in
\([-1,1]^{d_h}\), and the residual class \(\mathcal R_{A,M}\) is
specified by \cref{eq:adapter-sieve}.

\paragraph{Step 1: Compare two residual encoders with the same source
branch \(\widehat h_s\).}
Coordinatewise clipping to a rectangle is Euclidean nonexpansive. Thus,
for any \(a,a'\in\mathcal R_{A,M}\) and \(z_0\in\mathcal Z_0\),
\begin{align*}
 &\|h_{\widehat h_s,a}(z_0)-h_{\widehat h_s,a'}(z_0)\|\\*
 &\quad=\big\|\clip_1\{\widehat h_s(z_0)+a(z_{0,I_A})\}
              -\clip_1\{\widehat h_s(z_0)+a'(z_{0,I_A})\}\big\|\\*
 &\quad\le\|a(z_{0,I_A})-a'(z_{0,I_A})\|.
\end{align*}
Taking the supremum over \(z_0\) proves
\begin{equation}\label{eq:residual-contraction}
 \|h_{\widehat h_s,a}-h_{\widehat h_s,a'}\|_\infty
          \le\|a-a'\|_\infty.
\end{equation}
The same inequality holds on the globally projected input domain. In
particular, its constant is exactly one and contains neither a parameter
norm nor a Lipschitz constant of \(\widehat h_s\).

For \(v,v'\in\cV_{t,N}\), add and subtract
\(F_{t,v',h_{\widehat h_s,a}}\). The representation-block constraint
\(\Lip_u(v')\le L_{u,t}\) from \cref{eq:lipschitz} then gives
\begin{align}
 &\|F_{t,v,h_{\widehat h_s,a}}
             -F_{t,v',h_{\widehat h_s,a'}}\|_\infty\nonumber\\*
 &\quad\le\|F_{t,v,h_{\widehat h_s,a}}
                       -F_{t,v',h_{\widehat h_s,a}}\|_\infty
       +\|F_{t,v',h_{\widehat h_s,a}}
                       -F_{t,v',h_{\widehat h_s,a'}}\|_\infty
       \nonumber\\*
 &\quad\le\|v-v'\|_\infty
       +L_{u,t}\|h_{\widehat h_s,a}-h_{\widehat h_s,a'}\|_\infty
 \le\|v-v'\|_\infty+L_{u,t}\|a-a'\|_\infty.
                         \label{eq:residual-proof-field-continuity}
\end{align}

\paragraph{Step 2: Build a proper conditional cover for the composed
field class \(\cF_A(N,M)\).}
Fix \(\varepsilon>0\). Let \(\{v_k\}_{k=1}^{m_v}\) be a proper
\(\varepsilon/2\)-cover of \(\cV_{t,N}\), and let
\(\{a_l\}_{l=1}^{m_a}\) be a proper
\(\varepsilon/(2L_{u,t})\)-cover of \(\mathcal R_{A,M}\).
Equation \eqref{eq:residual-proof-field-continuity} shows that the
\(m_vm_a\) feasible fields
\(F_{t,v_k,h_{\widehat h_s,a_l}}\) form a proper
\(\varepsilon\)-cover of \(\cF_A(N,M)\). Therefore
\begin{equation}\label{eq:residual-proof-product-cover}
 H_{\cF_A(N,M)}(\varepsilon)
 \le H_{\cV_{t,N}}(\varepsilon/2)
       +H_{\mathcal R_{A,M}}(\varepsilon/(2L_{u,t})).
\end{equation}
The residual sieve \(\mathcal R_{A,M}\) has the architecture in
\cref{eq:architecture}, with input dimension \(q_A\), output
dimension \(d_h\), and fixed clipping level \(B_R\). Applying
\cref{eq:network-entropy} to this sieve and to the head sieve, and
using the polynomial-growth assumptions on \(M,N\), gives
\begin{align*}
 H_{\cV_{t,N}}(n^{-2}/2)&\le CN\log^3(en),\\*
 H_{\mathcal R_{A,M}}(n^{-2}/(2L_{u,t}))&\le CM\log^3(en).
\end{align*}
Substituting these inequalities into
\cref{eq:residual-proof-product-cover} proves
\cref{eq:residual-entropy}. The source function \(\widehat h_s\)
is the identical fixed branch in every center and every covered field.
Thus the construction takes no cover of \(\widehat h_s\), and the
constant is independent of its architecture and fitted weights.

\paragraph{Step 3: Construct a feasible corrected comparator on the
good event \(E_{s,A}\).}
Now suppose that the fixed source realization lies in
\(E_{s,A}=\{\widehat h_s\in\cH_{\rm good}^A\}\). Choose the
functions promised by \cref{ass:correctable}, and abbreviate them by
\[
 a^*=a_{\widehat h_s},\qquad
 g^*=g_{\widehat h_s}^+,\qquad
 h^+=h_{\widehat h_s,a^*}.
\]
Equations \eqref{eq:correctable-functions} and
\eqref{eq:correctability} state that
\(\|a^*\|_\infty\le\delta_A\le B_R\) and
\begin{equation}\label{eq:residual-proof-ideal-risk}
 \|F_{t,g^*,h^+}-v_t^0\|_t^2
       \le\kappa_A\cD_s(\widehat h_s)+\zeta_A^2.
\end{equation}
Apply the value-approximation part of \cref{lem:approximation} to
\(a^*\) on \([-1,1]^{q_A}\), and the admissible head-approximation
part to \(g^*\) on \(\cK_t\). This gives
\(a_M\in\mathcal R_{A,M}\) and \(v_N\in\cV_{t,N}\) with
\begin{equation}\label{eq:residual-approximation}
 \|a_M-a^*\|_\infty\le CM^{-\beta_A/q_A},\qquad
 \|v_N-g^*\|_{\infty,\cK_t}\le CN^{-r_t/D_t}.
\end{equation}
For clarity, if \(\widetilde a_M\) is the pre-clipping residual
approximant, its clipped version obeys
\[
 \|\clip_{B_R}\widetilde a_M-a^*\|_\infty
 =\|\clip_{B_R}\widetilde a_M-\clip_{B_R}a^*\|_\infty
 \le\|\widetilde a_M-a^*\|_\infty,
\]
since every coordinate of \(a^*\) lies in \([-B_R,B_R]\).
No derivative approximation is needed for \(a_M\). For the head
\(v_N\), the derivative bound from \cref{eq:derivative-approximation}
and the fixed margins \(\eta_{b,A}\) in \cref{ass:correctable}
ensure \(\Lip_b(v_N)\le L_{b,t}\) for
\(b\in\{x,u,\tau\}\) once \(N\) exceeds a fixed threshold.
The uniform smoothness bounds make all these constants independent of
\(\widehat h_s\) on \(E_{s,A}\) and of \(\delta_A\).

\paragraph{Step 4: Bound the corrected comparator's risk and anchor.}
Write \(h_M^+=h_{\widehat h_s,a_M}\), and define
\begin{align}
 e_1&=F_{t,v_N,h_M^+}-F_{t,g^*,h_M^+},\nonumber\\*
 e_2&=F_{t,g^*,h_M^+}-F_{t,g^*,h^+},\nonumber\\*
 e_3&=F_{t,g^*,h^+}-v_t^0.
                         \label{eq:residual-proof-error-decomposition}
\end{align}
The data-support condition in \cref{ass:path} and the ranges of the
encoders \(h_M^+,h^+\) place both reduced inputs in \(\cK_t\).
Thus \cref{eq:residual-approximation,eq:residual-contraction} give
\begin{align}
 \|e_1\|_t^2&\le CN^{-2r_t/D_t},\nonumber\\*
 \|e_2\|_t^2
 &\le L_{u,t}^2\|h_M^+-h^+\|_\infty^2
 \le L_{u,t}^2\|a_M-a^*\|_\infty^2
 \le CM^{-2\beta_A/q_A},\nonumber\\*
 \|e_3\|_t^2&\le\kappa_A\cD_s(\widehat h_s)+\zeta_A^2.
                         \label{eq:residual-proof-error-norms}
\end{align}
The second line uses \(\Lip_u(g^*;\cK_t)\le L_{u,t}\) from
\cref{ass:correctable}; the third is
\cref{eq:residual-proof-ideal-risk}. Since
\(F_{t,v_N,h_M^+}-v_t^0=e_1+e_2+e_3\), integration of the
three-vector squared triangle inequality yields
\[
 \cE_t(F_{t,v_N,h_M^+})
 \le3\sum_{k=1}^3\|e_k\|_t^2
 \le C\{\kappa_A\cD_s(\widehat h_s)+\zeta_A^2
                 +N^{-2r_t/D_t}+M^{-2\beta_A/q_A}\}.
\]
This is \cref{eq:residual-comparator}.

For the same comparator encoder \(h_M^+\), the assumed upper
control \cref{eq:anchor-upper-control} gives
\begin{align*}
 A_s(h_M^+)
 &\le\|a_M\|_\infty^2
 \le\{\|a^*\|_\infty+\|a_M-a^*\|_\infty\}^2\\*
 &\le2\|a^*\|_\infty^2+2\|a_M-a^*\|_\infty^2
 \le2\delta_A^2+CM^{-2\beta_A/q_A}.
\end{align*}
This proves \cref{eq:residual-comparator-anchor}. The argument uses
only an upper bound on \(A_s(h_M^+)\) at the comparator; it does
not infer control of a population norm from the anchor design.

\paragraph{Step 5: Use the same head \(v_N\) with the frozen encoder.}
Because \(\widehat h_s=\clip_1\widehat h_s\), the contraction
calculation gives
\begin{equation}\label{eq:residual-proof-correction-size}
 \|h^+-\widehat h_s\|_\infty
 \le\|a^*\|_\infty\le\delta_A.
\end{equation}
Use the following decomposition, which does not involve the residual
approximant \(a_M\):
\begin{align*}
 F_{t,v_N,\widehat h_s}-v_t^0
 ={}&[F_{t,v_N,\widehat h_s}-F_{t,g^*,\widehat h_s}]\\*
    &+[F_{t,g^*,\widehat h_s}-F_{t,g^*,h^+}]
       +[F_{t,g^*,h^+}-v_t^0].
\end{align*}
The squared population norms of these three brackets are bounded,
respectively, by \(CN^{-2r_t/D_t}\),
\(L_{u,t}^2\delta_A^2\), and
\(\kappa_A\cD_s(\widehat h_s)+\zeta_A^2\), by
\cref{eq:residual-approximation,eq:residual-proof-correction-size,eq:residual-proof-ideal-risk}.
Therefore the head infimum in \cref{eq:profile} satisfies
\begin{align*}
 Q_{t,N}(\widehat h_s)
 &\le\cE_t(F_{t,v_N,\widehat h_s})\\*
 &\le C\{\kappa_A\cD_s(\widehat h_s)+\zeta_A^2
                     +N^{-2r_t/D_t}+L_{u,t}^2\delta_A^2\},
\end{align*}
which proves \cref{eq:correctable-frozen-profile}. The smooth function
used here is the corrected head \(g^*\), not an assumed smooth
version of the reduced conditional mean at \(\widehat h_s\).
\end{proof}

\subsection{A penalized oracle with joint-class entropy}
\begin{lemma}[Conditional penalized least-squares oracle]
\label{lem:penalized-joint}
Condition on target-independent information fixing a class of pairs
\(\mathcal P=\{(v,h)\}\), its composed field class \(\cF\), and a
nonnegative functional penalty \(A_s(h)\). Suppose the bounded-regression
conditions of \cref{lem:relative} hold. Write \(n=n_t\),
\(\epsilon\ge0\), and
\(b_{\cF}(x)=C(1+K_t)^2\{H_{\cF}(n^{-2})+x+1\}/n\).
On an event of conditional probability at least \(1-e^{-x}\), every
\(\epsilon\)-minimizer \((\widehat v_\lambda,\widehat h_\lambda)\)
of the empirical squared loss plus \(\lambda A_s(h)\), with
\(\lambda\ge0\), satisfies
\begin{align}
 &\cE_t(\widehat F_\lambda)
          +\tfrac{4\lambda}{3}A_s(\widehat h_\lambda)\nonumber\\
 &\quad\le
 \inf_{(v,h)\in\mathcal P}
       \left\{\tfrac53\cE_t(F_{t,v,h})
                    +\tfrac{4\lambda}{3}A_s(h)\right\}
       +\tfrac83 b_{\cF}(x)+\tfrac43\epsilon.
                         \label{eq:penalized-joint-oracle}
\end{align}
The event is simultaneous in \(\lambda\ge0\) for the fixed class.
At \(\lambda=0\), the penalty term is omitted throughout.
For a fixed \(\lambda\), the conditional expectation version replaces
\(b_{\cF}(x)\) by
\(C(1+K_t)^2\{H_{\cF}(n^{-2})+1\}/n\).
\end{lemma}
\begin{proof}\proofanchor{lem:penalized-joint}
Let \(\mathcal B\) be the target-independent sigma-field fixing the
pair class \(\mathcal P\), its composed field class \(\cF\), and
the functional penalty \(A_s\). We work conditional on a realization
of \(\mathcal B\). In particular, the population operator \(P\)
is still integration under the original target regression law, and
\(P_n\) is the target empirical operator with \(n=n_t\).
Put \(H_n=H_{\cF}(n^{-2})\); if \(H_n=\infty\), the asserted
bound is vacuous.

\paragraph{Step 1: Obtain a single event for every field \(F\in\cF\).}
For each \(F\in\cF\), define
\(D_F=\ell_F-\ell_{v_t^0}\), where
\(\ell_F(W_t,U_{t,T})=\|F(W_t)-U_{t,T}\|^2\).
By the regression identity in \cref{eq:excess},
\(PD_F=\cE_t(F)\). The conditional version of
\cref{eq:relative}, with \(K_0=1+K_t\), supplies an event
\(\Omega_x\) of conditional probability at least \(1-e^{-x}\)
on which
\begin{equation}\label{eq:penalized-proof-uniform-event}
 \tfrac34\cE_t(F)-b_{\cF}(x)
 \le P_nD_F
 \le\tfrac54\cE_t(F)+b_{\cF}(x),
 \qquad F\in\cF.
\end{equation}
The event \(\Omega_x\) depends on the field class \(\cF\) and
the observations, but its definition does not involve \(\lambda\)
or the penalty values \(A_s(h)\).

\paragraph{Step 2: Retain the penalty in the empirical comparison.}
Fix \(\lambda\ge0\), and write
\(\widehat F_\lambda=F_{t,\widehat v_\lambda,\widehat h_\lambda}\).
For every comparator pair \((v,h)\in\mathcal P\), the
\(\epsilon\)-minimization property gives
\[
 P_n\ell_{\widehat F_\lambda}
       +\lambda A_s(\widehat h_\lambda)
 \le P_n\ell_{F_{t,v,h}}+\lambda A_s(h)+\epsilon.
\]
Subtracting the common quantity \(P_n\ell_{v_t^0}\), and then
applying the lower and upper bounds in
\cref{eq:penalized-proof-uniform-event}, yields on \(\Omega_x\)
\begin{align}
 &\tfrac34\cE_t(\widehat F_\lambda)-b_{\cF}(x)
                  +\lambda A_s(\widehat h_\lambda)\nonumber\\*
 &\quad\le P_nD_{\widehat F_\lambda}
                  +\lambda A_s(\widehat h_\lambda)\nonumber\\*
 &\quad\le P_nD_{F_{t,v,h}}+\lambda A_s(h)+\epsilon\nonumber\\*
 &\quad\le\tfrac54\cE_t(F_{t,v,h})+b_{\cF}(x)
                  +\lambda A_s(h)+\epsilon.
                         \label{eq:penalized-proof-basic-inequality}
\end{align}
Move the two copies of \(b_{\cF}(x)\) to the right and multiply
by \(4/3\). Equation \eqref{eq:penalized-proof-basic-inequality}
then becomes
\begin{align*}
 \cE_t(\widehat F_\lambda)
           +\tfrac{4\lambda}{3}A_s(\widehat h_\lambda)
 &\le\tfrac53\cE_t(F_{t,v,h})+\tfrac{4\lambda}{3}A_s(h)
                   +\tfrac83b_{\cF}(x)+\tfrac43\epsilon.
\end{align*}
Taking the infimum over the pair \((v,h)\) proves
\cref{eq:penalized-joint-oracle}. This is a pair infimum: if two
pairs define the same field but have different penalties, the value
\(A_s(h)\) remains attached to its own pair. No identifiability of
\((v,h)\) from \(F_{t,v,h}\) is assumed.

Since \cref{eq:penalized-proof-uniform-event} holds for every
\(F\in\cF\) on the same event \(\Omega_x\), the preceding
algebra is valid there for every \(\lambda\ge0\) and every
corresponding \(\epsilon\)-minimizer. There is no union bound over
\(\lambda\), nor any division by \(\lambda\). In particular, at
\(\lambda=0\), \cref{eq:penalized-joint-oracle} reduces directly to
\[
 \cE_t(\widehat F_0)
 \le\tfrac53\inf_{(v,h)\in\mathcal P}\cE_t(F_{t,v,h})
       +\tfrac83b_{\cF}(x)+\tfrac43\epsilon.
\]
No limiting argument for \(\widehat F_\lambda\) as
\(\lambda\downarrow0\) is needed.

\paragraph{Step 3: Integrate the conditional tail for a fixed
\(\lambda\).}
Define the conditional comparator value
\[
 I_\lambda=\inf_{(v,h)\in\mathcal P}
       \{\tfrac53\cE_t(F_{t,v,h})
                              +\tfrac{4\lambda}{3}A_s(h)\}.
\]
If \(I_\lambda=\infty\), the expectation assertion is vacuous.
Otherwise put
\begin{align*}
 Z_\lambda
 &=\big\{\cE_t(\widehat F_\lambda)
       +\tfrac{4\lambda}{3}A_s(\widehat h_\lambda)
                     -I_\lambda-\tfrac43\epsilon\big\}_+,\\*
 d_n&=\frac{8C(1+K_t)^2}{3n}.
\end{align*}
All quantities in \(I_\lambda\) and \(H_n\) are fixed after
conditioning on \(\mathcal B\). The high-probability inequality
already proved gives
\[
 \PP\{Z_\lambda>d_n(H_n+1+x)\mid\mathcal B\}
       \le e^{-x},\qquad x\ge0.
\]
Consequently,
\begin{align*}
 \E(Z_\lambda\mid\mathcal B)
 &\le d_n(H_n+1)
   +d_n\int_0^\infty
       \PP\{Z_\lambda>d_n(H_n+1+x)\mid\mathcal B\}\dd x\\*
 &\le d_n(H_n+2)
 \le C(1+K_t)^2\{H_n+1\}/n.
\end{align*}
Add \(I_\lambda+(4/3)\epsilon\) to this inequality to obtain the
claimed conditional expectation version, with a possibly enlarged
universal constant. If the pair infimum \(I_\lambda\) is not
attained, its use is justified by taking pairs within an arbitrary
positive error of \(I_\lambda\) in
\cref{eq:penalized-proof-basic-inequality} and letting that error tend
to zero. The cluster extension follows from the conditional cluster
version of \cref{lem:relative} without changing this algebra.
\end{proof}

\subsection{Proof of the end-to-end adaptation theorem}
\begin{proof}[Proof of \cref{thm:anchor}]\proofanchor{thm:anchor}
The proof uses the source regression bound, a conditional penalized
comparison on \(E_{s,A}\), and a deterministic envelope bound on
\(E_{s,A}^c\). The frozen estimator is analyzed under the same
correctability assumption, not under \cref{ass:transfer}.

\paragraph{Step 1: Derive the source contribution.}
By \cref{eq:source-estimator}, the source pair
\((\widehat v_s,\widehat h_s)\) is an \(n_s^{-2}\)-minimizer over
\(\cV_{s,N_s}\times\cH_{M_s}\). Applying \cref{lem:joint} using
only the source parts of \cref{ass:path,ass:smooth}, and then using
\cref{eq:balance} at \((r_s,D_s)\) and \((\beta_s,p)\), gives
\begin{align}
 \E\cD_s(\widehat h_s)
 &\le\E\cE_s(F_{s,\widehat v_s,\widehat h_s})\nonumber\\*
 &\le C\left\{\zeta_s^2+N_s^{-2r_s/D_s}+M_s^{-2\beta_s/p}
                +\frac{(N_s+M_s)\ell_{n_s}^3}{n_s}+n_s^{-2}\right\}
 \nonumber\\*
 &\le C R_s(n_s).
                         \label{eq:adaptation-source-bound}
\end{align}
Thus the source bound used here has been proved independently of the
calibration and reduced-mean assumptions in \cref{ass:transfer}.

\paragraph{Step 2: Fix the target-independent information \(\cS\).}
Recall the sigma-field
\(\cS=\sigma(\sD_s,\text{source auxiliary draws},\sD_a)\)
from \cref{sec:adaptation}. Conditional on \(\cS\), the source
function \(\widehat h_s\), the anchor \(A_s\), and the pair class
\[
 \mathcal P_A=\cV_{t,N_t}\times\cH_A(M_A)
\]
are fixed. The target observations and their auxiliary draws retain
their original independent sampling law. Also
\(E_{s,A}=\{\widehat h_s\in\cH_{\rm good}^A\}\in\cS\).
The composed field class associated with \(\mathcal P_A\) is
\(\cF_A(N_t,M_A)\), and \cref{eq:residual-entropy} gives
\begin{equation}\label{eq:anchor-proof-conditional-entropy}
 H_{\cF_A(N_t,M_A)}(n_t^{-2})
       \le C(N_t+M_A)\ell_{n_t}^3.
\end{equation}
The approximate minimization property \cref{eq:anchor-estimator}
is precisely the hypothesis of \cref{lem:penalized-joint}, with
\(\epsilon=n_t^{-2}\). Its conditional expectation conclusion,
\cref{eq:anchor-proof-conditional-entropy}, and
\(\lambda A_s(\widehat h_\lambda)\ge0\) for
\(0 \leq \lambda\le1\) imply
\begin{align}
 \E\{\cE_t(\widehat F_\lambda)\mid\cS\}
 &\le\inf_{(v,h)\in\mathcal P_A}
      \{\tfrac53\cE_t(F_{t,v,h})+\tfrac{4\lambda}{3}A_s(h)\}
       \nonumber\\*
 &\quad+C\frac{(N_t+M_A)\ell_{n_t}^3}{n_t}
                                      +\tfrac43n_t^{-2}.
                         \label{eq:anchor-proof-conditional-oracle}
\end{align}
This inequality holds for almost every realization of \(\cS\),
before any restriction to \(E_{s,A}\).

\paragraph{Step 3: Insert the corrected comparator on \(E_{s,A}\).}
On \(E_{s,A}\), \cref{lem:residual-neural} supplies the pair
\((v_{N_t},h_{\widehat h_s,a_{M_A}})\in\mathcal P_A\). Its risk
and anchor satisfy
\cref{eq:residual-comparator,eq:residual-comparator-anchor}.
Substitute these two inequalities into the pair infimum in
\cref{eq:anchor-proof-conditional-oracle} to obtain
\begin{align}\label{eq:adaptive-conditional-rate}
 \E\{\cE_t(\widehat F_\lambda)\mid\cS\}
 \le C\bigg\{&\kappa_A\cD_s(\widehat h_s)+\zeta_A^2
          +N_t^{-2r_t/D_t}+M_A^{-2\beta_A/q_A}\nonumber\\*
 &+\lambda\delta_A^2+\lambda M_A^{-2\beta_A/q_A}
          +\frac{(N_t+M_A)\ell_{n_t}^3}{n_t}+n_t^{-2}\bigg\},
 \qquad\text{on }E_{s,A}.
\end{align}
In particular, the term \(\lambda\delta_A^2\) comes from the
comparator penalty \(\lambda A_s(h_{\widehat h_s,a_{M_A}})\),
not from an asserted lower norm comparison for the anchor. Since
\(0 \leq \lambda\le1\),
\[
 M_A^{-2\beta_A/q_A}+\lambda M_A^{-2\beta_A/q_A}
       \le2M_A^{-2\beta_A/q_A}.
\]
Thus the residual approximation penalty is absorbed without introducing
an inverse power of \(\lambda\).

\paragraph{Step 4: Keep the failure-event contribution and balance the
target resolutions.}
For every training realization, conditional Jensen's inequality gives
\(\|v_t^0(W_t)\|\le B_{U,t}\), while membership of the fitted
head \(\widehat v_\lambda\) in \(\cV_{t,N_t}\) gives
\(\|\widehat F_\lambda(W_t)\|\le B_{V,t}\). Therefore
\begin{equation}\label{eq:anchor-proof-uniform-risk-envelope}
 0\le\cE_t(\widehat F_\lambda)
 =\int\|\widehat F_\lambda(w)-v_t^0(w)\|^2\nu_t(\dd w)
 \le(B_{V,t}+B_{U,t})^2=K_t^2.
\end{equation}
Let \(e_\lambda(\cS)=\E\{\cE_t(\widehat F_\lambda)\mid\cS\}\).
Since \(E_{s,A}\in\cS\), the tower property gives the exact
split
\[
 \E\cE_t(\widehat F_\lambda)
 =\E\{\ind(E_{s,A})e_\lambda(\cS)\}
   +\E\{\ind(E_{s,A}^c)e_\lambda(\cS)\}.
\]
Use \cref{eq:adaptive-conditional-rate} on the first event and
\cref{eq:anchor-proof-uniform-risk-envelope} on the second. The
inequalities \(\PP(E_{s,A}^c)\le\pi_{s,A}\) and
\(\E\{\ind(E_{s,A})\cD_s(\widehat h_s)\}
     \le\E\cD_s(\widehat h_s)\) yield
\begin{align}
 \E\cE_t(\widehat F_\lambda)
 \le C\bigg\{&\kappa_A\E\cD_s(\widehat h_s)+\zeta_A^2
          +K_t^2\pi_{s,A}+\lambda\delta_A^2\nonumber\\*
 &+N_t^{-2r_t/D_t}+M_A^{-2\beta_A/q_A}
        +\frac{(N_t+M_A)\ell_{n_t}^3}{n_t}+n_t^{-2}\bigg\}.
                         \label{eq:anchor-proof-unconditional-unbalanced}
\end{align}
This calculation requires no independence between \(E_{s,A}\),
\(\widehat h_s\), and the anchor design \(\sD_a\).

Next substitute \cref{eq:adaptation-source-bound}. At the target
resolutions in \cref{eq:resolution-choice,eq:adapter-resolution},
\cref{eq:balance} gives
\begin{align*}
 N_t^{-2r_t/D_t}+\frac{N_t\ell_{n_t}^3}{n_t}+n_t^{-2}
       &\le C\rho_{r_t,D_t}(n_t),\\*
 M_A^{-2\beta_A/q_A}+\frac{M_A\ell_{n_t}^3}{n_t}+n_t^{-2}
       &\le C\rho_{\beta_A,q_A}(n_t).
\end{align*}
By \cref{eq:adapter-resolution,eq:adapter-source-term},
\cref{eq:anchor-proof-unconditional-unbalanced} consequently becomes
\[
 \E\cE_t(\widehat F_\lambda)
 \le C\{\mathfrak R_A(n_t)+T_A(n_s)+\lambda\delta_A^2\},
\]
which is \cref{eq:anchor-risk}.

\paragraph{Step 5: Prove the frozen comparison under the same
correctability assumption.}
Conditional on \(\cS\), the frozen estimator uses the class
\(\cF_F=\{F_{t,v,\widehat h_s}:v\in\cV_{t,N_t}\}\).
Equations \eqref{eq:frozen-entropy} and \eqref{eq:erm-oracle}, applied
to \cref{eq:freeze-estimator}, give
\[
 \E\{\cE_t(\widehat F_F)\mid\cS\}
 \le\tfrac53 Q_{t,N_t}(\widehat h_s)
          +C\frac{N_t\ell_{n_t}^3}{n_t}+\tfrac43n_t^{-2}.
\]
On \(E_{s,A}\), use the frozen comparator bound
\cref{eq:correctable-frozen-profile}, rather than the calibrated
profile bound \cref{eq:profile-source}. We obtain
\begin{align}
 \E\{\cE_t(\widehat F_F)\mid\cS\}
 \le C\bigg\{&\kappa_A\cD_s(\widehat h_s)+\zeta_A^2
          +N_t^{-2r_t/D_t}+L_{u,t}^2\delta_A^2\nonumber\\*
 &+\frac{N_t\ell_{n_t}^3}{n_t}+n_t^{-2}\bigg\},
 \qquad\text{on }E_{s,A}.
                         \label{eq:anchor-proof-frozen-conditional}
\end{align}
On \(E_{s,A}^c\), the proof of
\cref{eq:anchor-proof-uniform-risk-envelope} gives
\(\cE_t(\widehat F_F)\le K_t^2\) as well. Averaging
\cref{eq:anchor-proof-frozen-conditional}, retaining
\(K_t^2\pi_{s,A}\), and applying
\cref{eq:adaptation-source-bound,eq:balance} therefore yields
\begin{align*}
 \E\cE_t(\widehat F_F)
 &\le C\{\rho_{r_t,D_t}(n_t)+\kappa_A R_s(n_s)
             +\zeta_A^2+K_t^2\pi_{s,A}+L_{u,t}^2\delta_A^2\}\\*
 &=C\{\rho_{r_t,D_t}(n_t)+T_A(n_s)+L_{u,t}^2\delta_A^2\}.
\end{align*}
This proves \cref{eq:freeze-correctable-risk}. In this step the
comparator head is the smooth corrected head from
\cref{ass:correctable}, approximated by \(v_{N_t}\); no regularity
of \(m_{t,\widehat h_s}\) has been assumed.

\paragraph{Step 6: Convert both bounds to generation risks and choose
\(\lambda\).}
For \(\widehat F\in\{\widehat F_\lambda,\widehat F_F\}\), the
fitted head belongs to \(\cV_{t,N_t}\) and the fitted encoder is a
function only of context. Thus \(\widehat F\) satisfies all three
bounds in \cref{eq:main-proof-inherited-budgets}. The stability
inequalities \cref{eq:stability,eq:numerical-stability} apply for
each realized training sample and give, after averaging,
\begin{align*}
 \E\cW_t^2(P_{\widehat F},P_t^0)
       &\le C_{L_{x,t}}\E\cE_t(\widehat F),\\*
 \E\cW_t^2(P_{\widehat F}^{[J]},P_t^0)
       &\le2C_{L_{x,t}}\E\cE_t(\widehat F)
                             +2C_{\rm Euler}J^{-2}.
\end{align*}
Substitution of \cref{eq:anchor-risk,eq:freeze-correctable-risk}
proves \cref{eq:anchor-main,eq:freeze-correctable-main}, together
with their exact-integration versions.

Finally, let \(\lambda_{n_t}\) be any deterministic sequence satisfying
\cref{eq:lambda-choice}. By \cref{ass:correctable},
\(0\le\delta_A\le B_R\), so
\[
 0\le\lambda_{n_t}\delta_A^2
 \le B_R^2\lambda_{n_t}
 \le B_R^2\min\{1,\mathfrak R_A(n_t)\}
 \le B_R^2\mathfrak R_A(n_t).
\]
Insert this inequality into \cref{eq:anchor-main}, and expand
\(\mathfrak R_A\), \(T_A\), and \(R_s\) using
\cref{eq:adapter-resolution,eq:adapter-source-term,eq:source-rate-symbol}.
The result is exactly \cref{eq:anchor-clean}. Every constant used in
this proof depends only on the fixed envelopes, architecture constants,
smoothness bounds, dimensions, and Lipschitz margins, not on
\(n_s,n_t,\lambda,J,\delta_A\).
\end{proof}

\subsection{The information lower bound and strict separation}
\begin{lemma}[An endpoint lower bound for frozen context information]
\label{lem:frozen-information}
Fix a representation \(h\) and suppose \(\E\|X_t\|^2<\infty\).
Let \(Q(\cdot\mid z)\) be any conditional generator with finite
second moments whose dependence on \(z\) is only through
\(H_{t,h}(z)\). Then
\begin{equation}\label{eq:frozen-information-lower}
 \cW_t^2(Q,P_t^0)\ge\mathcal I_t(h).
\end{equation}
In particular, every exact or Euler-generated field
\(F_{t,v,h}\) has this lower bound. For a field with the fixed state
Lipschitz budget and the true-flow assumption,
\begin{equation}\label{eq:frozen-fm-lower}
 \cE_t(F_{t,v,h})\ge C_{L_{x,t}}^{-1}\mathcal I_t(h).
\end{equation}
Both statements hold conditionally on training and after averaging over
training when \(h\) is a fitted source representation.
\end{lemma}
\begin{proof}\proofanchor{lem:frozen-information}
Fix the representation \(h\), and let
\(H=H_{t,h}(Z_t)\). Define the target conditional mean
\(\mu_t(z)=\E(X_t\mid Z_t=z)\) as in
\cref{eq:frozen-mean-defect}. By conditional Jensen's inequality,
\begin{equation}\label{eq:information-proof-target-moment}
 \E\|\mu_t(Z_t)\|^2\le\E\|X_t\|^2<\infty.
\end{equation}

\paragraph{Step 1: Bound a transport cost by the discrepancy of its
conditional means.}
The assumed factorization of the generator means that there is a
probability kernel \(\widetilde Q\) with
\(Q(\cdot\mid z)=\widetilde Q(\cdot\mid H_{t,h}(z))\) for
almost every target context \(z\). Let
\(b(u)=\int y\,\widetilde Q(\dd y\mid u)\) denote its conditional
mean. For a fixed such \(z\) and any coupling
\(\pi_z\in\Pi\{Q(\cdot\mid z),P_t^0(\cdot\mid z)\}\),
Jensen's inequality under \(\pi_z\) gives
\begin{align*}
 \int\|y-x\|^2\pi_z(\dd y,\dd x)
 &\ge\left\|\int(y-x)\pi_z(\dd y,\dd x)\right\|^2\\*
 &=\|b(H_{t,h}(z))-\mu_t(z)\|^2.
\end{align*}
The final equality uses the two specified marginals of \(\pi_z\).
Taking the infimum over \(\pi_z\) in
\cref{eq:generation-loss} and then averaging over \(Z_t\) yields
\begin{equation}\label{eq:information-proof-mean-transport}
 \cW_t^2(Q,P_t^0)
 \ge\E\|b(H)-\mu_t(Z_t)\|^2.
\end{equation}
If the left side is infinite, the desired lower bound is immediate,
because \(\mathcal I_t(h)\le\E\|\mu_t(Z_t)\|^2<\infty\).
Otherwise \cref{eq:information-proof-mean-transport} and
\cref{eq:information-proof-target-moment} imply
\[
 \E\|b(H)\|^2
 \le2\E\|b(H)-\mu_t(Z_t)\|^2
                    +2\E\|\mu_t(Z_t)\|^2<\infty.
\]
Thus all conditional projection terms used next are square integrable.

\paragraph{Step 2: Project the target mean onto the frozen context
sigma-field \(\sigma(H)\).}
Write \(\overline\mu_h(H)=\E\{\mu_t(Z_t)\mid H\}\). The
vector \(b(H)-\overline\mu_h(H)\) is \(\sigma(H)\)-measurable,
and the defining conditional expectation identity gives
\begin{align*}
 &\E\ip{b(H)-\overline\mu_h(H)}
              {\overline\mu_h(H)-\mu_t(Z_t)}\\*
 &\quad=\E\ip{b(H)-\overline\mu_h(H)}
              {\E\{\overline\mu_h(H)-\mu_t(Z_t)\mid H\}}=0.
\end{align*}
Expanding
\(b(H)-\mu_t(Z_t)
 =[b(H)-\overline\mu_h(H)]+[\overline\mu_h(H)-\mu_t(Z_t)]\)
and using this zero cross term proves
\begin{align}
 \E\|b(H)-\mu_t(Z_t)\|^2
 &=\E\|b(H)-\overline\mu_h(H)\|^2
      +\E\|\mu_t(Z_t)-\overline\mu_h(H)\|^2\nonumber\\*
 &=\E\|b(H)-\overline\mu_h(H)\|^2+\mathcal I_t(h)
 \ge\mathcal I_t(h).
                         \label{eq:information-proof-projection}
\end{align}
Here the second equality is the definition
\cref{eq:frozen-mean-defect}. Combining
\cref{eq:information-proof-mean-transport,eq:information-proof-projection}
proves \cref{eq:frozen-information-lower}.

\paragraph{Step 3: Verify the required factorization for exact and
Euler-generated laws.}
Fix a feasible head \(v\) and a base initial value \(g\). For
\(F_{t,v,h}\), the exact equation \cref{eq:fitted-ode} reads
\[
 \dot X(\tau)=v(X(\tau),h(z_0),z_c,\tau),\qquad X(0)=g.
\]
If \(H_{t,h}(z)=H_{t,h}(z')\), then both the right-hand side and
the initial condition are identical for contexts \(z\) and \(z'\).
Uniqueness of the fitted ODE, supplied by its global state-Lipschitz
bound, implies identical exact endpoints. More explicitly, for each
reduced context \(u=(u_0,u_c)\), the solution of
\(\dot X=v(X,u_0,u_c,\tau)\) defines a measurable endpoint map
\(\varphi_v(g,u)\). The exact generator is the pushforward of
\(\gamma_t\) by \(g\mapsto\varphi_v(g,H_{t,h}(z))\), and
therefore depends on \(z\) only through \(H_{t,h}(z)\).

For Euler generation, define the iterates explicitly by
\begin{equation}\label{eq:information-proof-euler-recursion}
 x_0(g,z)=g,\qquad
 x_{k+1}(g,z)=x_k(g,z)
       +\frac1Jv(x_k(g,z),h(z_0),z_c,k/J),
 \quad 0\le k<J.
\end{equation}
An induction on \(k\) in \cref{eq:information-proof-euler-recursion}
shows that \(x_k(g,z)=x_k(g,z')\) whenever
\(H_{t,h}(z)=H_{t,h}(z')\). Hence the Euler endpoint law has the
same reduced-context factorization for every \(J\ge1\). The base
law \(\gamma_t\) is context independent by the model definition.
Both endpoint laws have finite second moments: boundedness of the head
\(v\) implies
\[
 \|X(1)\|\le\|G_t\|+B_{V,t},\qquad
 \|x_J(G_t,z)\|\le\|G_t\|+B_{V,t},
\]
so their second moments are at most
\(2\E\|G_t\|^2+2B_{V,t}^2\). They are therefore eligible for
\cref{eq:frozen-information-lower}.

\paragraph{Step 4: Derive the velocity lower bound using the exact
flow.}
Apply \cref{eq:frozen-information-lower} to the exact endpoint law
\(P_{F_{t,v,h}}\), and then use \cref{eq:stability} under the
stated true-flow assumption. The resulting inequalities are
\begin{equation}\label{eq:information-proof-stability-chain}
 \mathcal I_t(h)
 \le\cW_t^2(P_{F_{t,v,h}},P_t^0)
 \le C_{L_{x,t}}\cE_t(F_{t,v,h}).
\end{equation}
The constant \(C_{L_{x,t}}>0\) is given explicitly by
\cref{eq:stability-constant}. Division by this constant proves
\cref{eq:frozen-fm-lower}. The use of the exact-flow inequality is
important: no subtraction of an Euler discretization remainder is
needed to obtain the velocity lower bound.

\paragraph{Step 5: Condition on fitted functions and average over
training.}
Let \(\mathcal T\) contain all source and target fitting randomness,
and let the evaluation observation \((X_t,Z_t)\) and base draw
\(G_t\) be fresh and independent of \(\mathcal T\). Conditional
on \(\mathcal T\), the fitted functions are fixed and the evaluation
law is the original target law. Thus Steps 1--4 apply for almost every
training realization, even when the fitted head uses the target training
sample. For \(h=\widehat h_s\), integration over \(\mathcal T\)
gives a lower bound of \(\E\mathcal I_t(\widehat h_s)
=\mathcal I_F(n_s)\) for either expected endpoint risk, and
\(C_{L_{x,t}}^{-1}\mathcal I_F(n_s)\) for the expected velocity
risk. The equality with \(\mathcal I_F(n_s)\) uses
\cref{eq:expected-mean-defect} and the fact that \(\widehat h_s\)
is determined solely by source training.
\end{proof}

\begin{proof}[Proof of \cref{cor:strict-adaptation}]
\proofanchor{cor:strict-adaptation}
All limits below are along the sequence in
\cref{eq:strict-gap-conditions}, with \(\delta_A>0\). To distinguish
the deterministic expected risks from the fitted fields, write
\begin{align*}
 R_F^{[J]}&=\E\cW_t^2(P_{\widehat F_F}^{[J]},P_t^0),&
 R_\lambda^{[J]}&=\E\cW_t^2(P_{\widehat F_\lambda}^{[J]},P_t^0),\\*
 V_F&=\E\cE_t(\widehat F_F),&
 V_\lambda&=\E\cE_t(\widehat F_\lambda).
\end{align*}
Define the nonnegative deterministic remainder ratio
\begin{equation}\label{eq:separation-proof-remainder-ratio}
 r_{\rm gap}
   =\frac{\mathfrak R_A(n_t)+T_A(n_s)+J^{-2}}{\delta_A^2}.
\end{equation}
The second condition in \cref{eq:strict-gap-conditions} states exactly
that \(r_{\rm gap}\to0\).

\paragraph{Step 1: Match the frozen endpoint risk to \(\delta_A^2\).}
For each training realization, the Euler endpoint law of
\(\widehat F_F=F_{t,\widehat v_F,\widehat h_s}\) uses context only
through \(H_{t,\widehat h_s}\). The conditional assertion of
\cref{lem:frozen-information}, averaged over training, gives
\begin{equation}\label{eq:separation-proof-frozen-lower}
 R_F^{[J]}\ge\E\mathcal I_t(\widehat h_s)
       =\mathcal I_F(n_s)\ge c_0\delta_A^2.
\end{equation}
The last inequality is the first condition in
\cref{eq:strict-gap-conditions}, and it holds for every \(J\).
In particular, \(R_F^{[J]}>0\) along the sequence.

For the upper bound, \cref{eq:freeze-correctable-main} gives
\begin{align}
 R_F^{[J]}
 &\le C\{\rho_{r_t,D_t}(n_t)+T_A(n_s)+J^{-2}
                                      +L_{u,t}^2\delta_A^2\}
 \nonumber\\*
 &\le C\{r_{\rm gap}+L_{u,t}^2\}\delta_A^2.
                         \label{eq:separation-proof-frozen-upper}
\end{align}
Here \(\rho_{r_t,D_t}(n_t)\le\mathfrak R_A(n_t)\) by
\cref{eq:adapter-resolution}. For all sufficiently large sample sizes,
\(r_{\rm gap}\le1\), so
\cref{eq:separation-proof-frozen-lower,eq:separation-proof-frozen-upper}
imply
\[
 c_0\delta_A^2\le R_F^{[J]}
          \le C(1+L_{u,t}^2)\delta_A^2.
\]
Both constants are fixed along the sequence. This proves the two-sided
statement \cref{eq:freeze-matched-gap}.

\paragraph{Step 2: Compare the adaptive endpoint risk with the positive
frozen lower bound.}
Equation \eqref{eq:anchor-main} and the definition
\cref{eq:separation-proof-remainder-ratio} give
\begin{equation}\label{eq:separation-proof-adaptive-upper}
 0\le R_\lambda^{[J]}
 \le C\{\mathfrak R_A(n_t)+T_A(n_s)+J^{-2}
                                 +\lambda\delta_A^2\}
 =C(r_{\rm gap}+\lambda)\delta_A^2.
\end{equation}
The conditions \(r_{\rm gap}\to0\) and \(\lambda\to0\) imply
\(R_\lambda^{[J]}=o(\delta_A^2)\). More explicitly, division by
\cref{eq:separation-proof-frozen-lower} gives
\[
 0\le\frac{R_\lambda^{[J]}}{R_F^{[J]}}
       \le\frac{C}{c_0}(r_{\rm gap}+\lambda)\longrightarrow0.
\]
This proves \cref{eq:strict-generation-ratio} by a lower bound on
the denominator, rather than by comparing two upper bounds.

\paragraph{Step 3: Establish the exact-flow and velocity-risk assertions.}
For exact integration, define \(R_F,R_\lambda\) by replacing each
numerical law \(P_{\widehat F}^{[J]}\) with
\(P_{\widehat F}\), and replace \(r_{\rm gap}\) by
\[
 r_{\rm gap}^{\rm exact}
       =\{\mathfrak R_A(n_t)+T_A(n_s)\}/\delta_A^2.
\]
The exact-endpoint part of \cref{lem:frozen-information} supplies
\(R_F\ge c_0\delta_A^2\), and the exact-integration versions of
\cref{eq:anchor-main,eq:freeze-correctable-main} supply the same
upper bounds with \(r_{\rm gap}^{\rm exact}\) and no Euler term.
Thus the exact-flow conclusions follow under the stated corresponding
condition with \(J^{-2}\) omitted.

For velocity risk, averaging \cref{eq:frozen-fm-lower} gives
\[
 V_F\ge C_{L_{x,t}}^{-1}\mathcal I_F(n_s)
                 \ge(c_0/C_{L_{x,t}})\delta_A^2>0.
\]
On the other hand, \cref{eq:anchor-risk} gives
\(V_\lambda\le C\{\mathfrak R_A(n_t)+T_A(n_s)
                                      +\lambda\delta_A^2\}\).
Consequently,
\begin{align*}
 0\le\frac{V_\lambda}{V_F}
 &\le\frac{CC_{L_{x,t}}}{c_0}
       \left\{\frac{\mathfrak R_A(n_t)+T_A(n_s)}{\delta_A^2}
                    +\lambda\right\}\\*
 &\le\frac{CC_{L_{x,t}}}{c_0}(r_{\rm gap}+\lambda)
       \longrightarrow0.
\end{align*}
For the exact-integration formulation use
\(r_{\rm gap}^{\rm exact}\) in the last line. This proves the
claimed ratio of expected flow-matching excess risks.
\end{proof}

\subsection{Proof of the non-transfer theorem}
\begin{proof}[Proof of \cref{thm:nt}]\proofanchor{thm:nt}
The target-only estimator in \cref{eq:nt-estimator} fits both the
head \(\widehat v_{\rm NT}\) and the encoder
\(\widehat h_{\rm NT}\) on the same target observations. We therefore
apply the regression oracle to a joint class, not to a class obtained by
conditioning on \(\widehat h_{\rm NT}\).

\paragraph{Step 1: Identify the target-only class and its smooth
comparator.}
The deterministic class fixed before target training is
\[
 \cF_{\rm NT}
   =\{F_{t,v,h}:v\in\cV_{t,N_t},\ h\in\cH_{M_t}\},\qquad
 N_t=K_{r_t,D_t}(n_t),\quad M_t=K_{\beta_t,p}(n_t).
\]
The target part of \cref{ass:smooth} supplies the functions
\(h_{t,0}\) and \(g_{t,0}\), with structural error \(\zeta_t^2\)
as defined by \cref{eq:structural-bias}. By
\cref{lem:approximation}, choose the feasible approximants
\(h_{M_t}\in\cH_{M_t}\) and \(v_{t,N_t}\in\cV_{t,N_t}\)
so that
\[
 \|h_{M_t}-h_{t,0}\|_\infty\le CM_t^{-\beta_t/p},\qquad
 \|v_{t,N_t}-g_{t,0}\|_{\infty,\cK_t}
                                  \le CN_t^{-r_t/D_t}.
\]
For the comparator field, add and subtract the two smooth-head
compositions to obtain
\begin{align*}
 F_{t,v_{t,N_t},h_{M_t}}-v_t^0
 ={}&[F_{t,v_{t,N_t},h_{M_t}}-F_{t,g_{t,0},h_{M_t}}]\\*
 &+[F_{t,g_{t,0},h_{M_t}}-F_{t,g_{t,0},h_{t,0}}]
   +[F_{t,g_{t,0},h_{t,0}}-v_t^0].
\end{align*}
The target support condition in \cref{ass:path}, the representation
Lipschitz bound in \cref{eq:smooth-margins}, and the three-term
squared triangle inequality then give
\begin{align}
 \inf_{F\in\cF_{\rm NT}}\cE_t(F)
 &\le\cE_t(F_{t,v_{t,N_t},h_{M_t}})\nonumber\\*
 &\le3\|v_{t,N_t}-g_{t,0}\|_{\infty,\cK_t}^2
       +3L_{u,t}^2\|h_{M_t}-h_{t,0}\|_\infty^2+3\zeta_t^2
       \nonumber\\*
 &\le C\{\zeta_t^2+N_t^{-2r_t/D_t}+M_t^{-2\beta_t/p}\}.
                         \label{eq:nt-proof-comparator}
\end{align}
Only the target comparator \(h_{t,0}\) is used; it need not equal
any source comparator \(h_{s,0}\).

\paragraph{Step 2: Count the jointly fitted networks and obtain the
unbalanced risk bound.}
By \cref{eq:composition-entropy,eq:network-entropy}, the class
\(\cF_{\rm NT}\) satisfies
\begin{align}
 H_{\cF_{\rm NT}}(n_t^{-2})
 &\le H_{\cV_{t,N_t}}(n_t^{-2}/2)
       +H_{\cH_{M_t}}(n_t^{-2}/(2L_{u,t}))\nonumber\\*
 &\le C(N_t+M_t)\ell_{n_t}^3.
                         \label{eq:nt-proof-entropy}
\end{align}
Both resolutions are polynomial in \(n_t\), as verified in the
proof of \cref{eq:balance}. The fitted field
\(\widehat F_{\rm NT}\) is an \(n_t^{-2}\)-minimizer of empirical
squared loss over \(\cF_{\rm NT}\) by \cref{eq:nt-estimator}.
The bounded-regression assumptions of \cref{lem:relative} hold with
\(f_0=v_t^0\), label envelope \(B_{U,t}\), and field envelope
\(B_{V,t}\). Applying its expectation oracle to
\cref{eq:nt-proof-comparator,eq:nt-proof-entropy} gives
\begin{equation}\label{eq:nt-proof-unbalanced}
 \E\cE_t(\widehat F_{\rm NT})
 \le C\left\{\zeta_t^2+N_t^{-2r_t/D_t}+M_t^{-2\beta_t/p}
       +\frac{(N_t+M_t)\ell_{n_t}^3}{n_t}+n_t^{-2}\right\}.
\end{equation}
This is also \cref{eq:joint-risk} with \(j=t\). The explicit
calculation shows why the target entropy contains both \(N_t\) and
\(M_t\): neither fitted network is independent of target training.

\paragraph{Step 3: Balance head approximation, encoder approximation,
and estimation.}
Apply \cref{eq:balance} first with
\((a,D,n)=(r_t,D_t,n_t)\) and then with
\((a,D,n)=(\beta_t,p,n_t)\). Substitution into
\cref{eq:nt-proof-unbalanced} yields
\begin{align*}
 \E\cE_t(\widehat F_{\rm NT})
 &\le C\{\zeta_t^2+\rho_{r_t,D_t}(n_t)
                               +\rho_{\beta_t,p}(n_t)\}\\*
 &=C R_{\rm NT}(n_t),
\end{align*}
where the equality is \cref{eq:nt-rate-symbol}. This proves
\cref{eq:nt-risk}.

\paragraph{Step 4: Bound the endpoint error without source assumptions.}
For every training realization, \(\widehat v_{\rm NT}\in\cV_{t,N_t}\)
and the encoder \(\widehat h_{\rm NT}\) depends only on context.
Thus \(\widehat F_{\rm NT}\) satisfies the three inequalities in
\cref{eq:main-proof-inherited-budgets}. By the target true-flow
condition in \cref{ass:path}, \cref{lem:stability} applies and gives
\begin{align*}
 \E\cW_t^2(P_{\widehat F_{\rm NT}},P_t^0)
 &\le C_{L_{x,t}}\E\cE_t(\widehat F_{\rm NT})
 \le C R_{\rm NT}(n_t),\\*
 \E\cW_t^2(P_{\widehat F_{\rm NT}}^{[J]},P_t^0)
 &\le2C_{L_{x,t}}\E\cE_t(\widehat F_{\rm NT})
                             +2C_{\rm Euler}J^{-2}\\*
 &\le C\{R_{\rm NT}(n_t)+J^{-2}\}.
\end{align*}
These are the exact and numerical assertions in
\cref{eq:nt-main}. Every class, comparator, and observation used in
Steps 1--4 is target-only; neither source calibration nor source
correctability is invoked.
\end{proof}

\section{Verifying approximation and functional anchoring}
\label{app:verification}
\subsection{A verified  example with genuine adaptation gain}
\begin{proposition}[A conditional-flow separation with neural residuals]
\label{prop:adaptation-example}
Let \(q\ge1\), \(p=q+1\), \(d_s=d_t=d_h=1\), and
\(p_{s,c}=p_{t,c}=0\), so \(D_s=D_t=3\). In both tasks let
\(Z=(V,W)\), where \(V\) is uniform on \([-1,1]\), \(W\) is uniform
on \([-1,1]^q\), and \(V,W\) are independent. Let
\(f\in C^{\beta_A}([-1,1]^q;B_f)\), \(\beta_A>0\), satisfy
\(\|f\|_\infty\le1\), \(\E f(W)=0\), and
\(\sigma_f^2:=\E f(W)^2>0\). For \(0<\delta\le1/8\), set
\begin{equation}\label{eq:example-responses}
 X_s=V/4,\qquad X_t=V/4+\delta f(W),\qquad
 G_s=G_t=0,\qquad S_{j,\tau}=\tau X_j.
\end{equation}
Take \(K_{x,s}=K_{x,t}=1\), \(b_s=b_t=1\),
\(L_{x,s}=L_{x,t}=L_{\tau,s}=L_{\tau,t}=1\), and
\(L_{u,s}=L_{u,t}=2\). At all sufficiently large neural resolutions,
one valid measurable source minimizer is
\begin{equation}\label{eq:example-source-fit}
 \widehat c=\frac14+\frac{1}{4n_s}\sum_{i=1}^{n_s}V_{s,i}^2,
 \qquad \widehat h_s(z)=\widehat c V,\qquad
 \widehat v_s(x,u,\tau)=\clip_1\{u/(4\widehat c)\}.
\end{equation}
The source empirical loss is exactly zero, and
\(\cD_s(\widehat h_s)=0\).

Use residual coordinates \(I_A=\{2,\ldots,p\}\), \(q_A=q\),
\(B_R=1\), and either the empirical or supremum functional anchor.
Then \cref{ass:correctable} holds with
\(\delta_A=2\delta\), \(\kappa_A=\zeta_A=\pi_{s,A}=0\), any
fixed \(r_t>1\), and fixed smoothness bounds uniform over
\(0<\delta\le1/8\). Every frozen generator using
\(\widehat h_s\), including any fitted head in \cref{alg:freeze},
satisfies
\begin{equation}\label{eq:example-lower}
 \E\cW_t^2(P_{\widehat F_F}^{[J]},P_t^0)
       \ge\delta^2\sigma_f^2,
 \qquad
 \E\cW_t^2(P_{\widehat F_F},P_t^0)
       \ge\delta^2\sigma_f^2.
\end{equation}
For \cref{alg:anchor} with the residual sieve,
any deterministic \(0\le\lambda\le\min\{1,\mathfrak R_A(n_t)\}\),
and the source rule
\cref{eq:example-source-fit},
\begin{equation}\label{eq:example-upper}
 \E\cW_t^2(P_{\widehat F_\lambda}^{[J]},P_t^0)
       \le C\{\rho_{r_t,3}(n_t)+\rho_{\beta_A,q}(n_t)+J^{-2}\}.
\end{equation}
The exact-flow version omits \(J^{-2}\). In particular, for
\(q=2,\beta_A=2,r_t=4\), the right side is
\(O((\ell_{n_t}^3/n_t)^{2/3}+J^{-2})\). For these same indices, fix
\(0<\delta_0\le1/8\), take \(\delta=\delta_0n_t^{-1/4}\), and
choose \(J^{-2}\lesssim n_t^{-2/3}\ell_{n_t}^2\). Then
\begin{equation}\label{eq:example-shrinking}
 \E\cW_t^2(P_{\widehat F_F}^{[J]},P_t^0)\asymp n_t^{-1/2},
 \qquad
 \E\cW_t^2(P_{\widehat F_\lambda}^{[J]},P_t^0)
            =O(n_t^{-2/3}\ell_{n_t}^2).
\end{equation}
The ratio is \(O(n_t^{-1/6}\ell_{n_t}^2)\), which tends to zero.
Under the target-only smooth-class theorem with \(\beta_t=2,p=3\), its general upper envelope
instead contains \((\ell_{n_t}^3/n_t)^{4/7}\); this last statement is
only an upper-bound comparison with target-only estimation.
\end{proposition}
\begin{proof}\proofanchor{prop:adaptation-example}
The samples in the two tasks are independent copies of their specified
experiments. The base point mass at zero is independent of each sample,
and an independent uniform interpolation time is generated as in
\cref{eq:empirical}. The path derivative is
\(U_{j,\tau}=X_j\), and
\(|X_s|\le1/4\), \(|X_t|\le1/4+1/8=3/8\).
The state lies in \([-1,1]\), so the bounded-regression assumptions hold.
Since the responses are deterministic given the context, versions of the
full conditional velocities are
\[
 v_s^0(x,z,\tau)=V/4,\qquad
 v_t^0(x,z,\tau)=V/4+\delta f(W),
\]
on the entire state space. For fixed \(z\) these velocities are constant
in state and time. Their unique characteristics starting at zero are
\(\Phi_{j,\tau}(0,z)=\tau X_j(z)\), which have exactly the
interpolation marginals and the specified endpoint laws. Thus the
true-flow part of \cref{ass:path} also holds, including finite second
moments and joint measurability.

For the source smooth comparator, take
\(h_{s,0}(z)=V/4\) and \(g_{s,0}(x,u,\tau)=u\) on \(\cK_s\).
Both are smooth of any fixed finite order, \(\zeta_s=0\), and their
block-Lipschitz constants fit strictly inside the chosen budgets.
For every \(c\in[1/4,1/2]\), the representation \(h_c(z)=cV\)
and head \(v_c(x,u,\tau)=\clip_1\{u/(4c)\}\) are exact finite
ReLU implementations with a fixed number of nonzero coefficients.
Signed affine maps can be written as the difference of two ReLU units,
and the prescribed clips are also ReLU operations. Their coefficients
are bounded uniformly in \(c\); after padding, they belong to the
specified sieves for all sufficiently large resolutions. Their head
output bound is one, state/time Lipschitz constants are zero, and the
representation-block Lipschitz constant is \(1/(4c)\le1<2\).
At every augmented source observation,
\[
 v_c(S_{s,i,T_{s,i}},h_c(Z_{s,i}),T_{s,i})
       =\frac{cV_{s,i}}{4c}=X_{s,i}=U_{s,i}.
\]
Their empirical loss is zero, the global minimum of a sum of squares.
The source-data-dependent value \(\widehat c\) in
\cref{eq:example-source-fit} belongs to \([1/4,1/2]\), so that
formula is a measurable exact minimizer of the original source objective.
One can define the rule on other datasets by returning an arbitrary
measurable approximate minimizer whenever this proposed pair has nonzero
empirical loss. 
Since \(cV\) determines \(V\),
source sufficiency gives \(\cD_s(h_c)=0\).

Set \(\cH_{\rm good}^A=\{h_c:1/4\le c\le1/2\}\). For a
representation \(h_c\) in this set, choose
\[
 a_{h_c}(w)=4c\delta f(w),\qquad
 g_{h_c}^+(x,u,\tau)=u/(4c)\quad\hbox{on }\cK_t.
\]
The correction has supremum norm at most \(2\delta=\delta_A\)
and a fixed \(C^{\beta_A}\) bound, because \(4c\delta\le1/4\).
The head is affine, uniformly bounded by one on its rectangle, smooth
of every fixed finite order, and has representation-block Lipschitz
constant at most one. Fixed positive margins, for example
\(\eta_{x,A}=\eta_{u,A}=\eta_{\tau,A}=1/2\), are valid.
Moreover,
\[
 |cV+4c\delta f(W)|\le\tfrac12+\tfrac14=\tfrac34,
\]
so output clipping in the corrected encoder is inactive. Therefore
\[
 F_{t,g_{h_c}^+,h_{h_c,a_{h_c}}}(x,z,\tau)
 =\frac{cV+4c\delta f(W)}{4c}
 =V/4+\delta f(W)=v_t^0(x,z,\tau).
\]
This verifies \cref{eq:correctability} with zero right side, all required
uniform smoothness bounds, and \(\pi_{s,A}=0\). The anchor inequality
\cref{eq:anchor-upper-control} holds on every finite design. \Cref{thm:anchor}, with the tuning
range \cref{eq:lambda-choice}, now gives \cref{eq:example-upper},
since \(T_A(n_s)=0\).

Finally, conditional on any source realization, \(\widehat c V\)
generates the same sigma-field as \(V\). By independence and centering,
\[
 \E\{\mu_t(Z_t)\mid\widehat c V\}=V/4,
 \qquad
 \mathcal I_t(\widehat h_s)
      =\delta^2\E f(W)^2=\delta^2\sigma_f^2.
\]
\Cref{lem:frozen-information} proves both inequalities in
\cref{eq:example-lower}. For \(q=2,\beta_A=2,r_t=4\),
\(2r_t/(2r_t+3)=8/11>2/3=2\beta_A/(2\beta_A+q)\), so the residual
term dominates the stated adaptive upper bound. With \(\delta=\delta_0n_t^{-1/4}\), the lower bound
\cref{eq:example-lower} is a fixed positive multiple of \(n_t^{-1/2}\).
The frozen upper bound \cref{eq:freeze-correctable-main}, with
\(T_A=0\) and \(\delta_A=2\delta\), is of that same order, since
\(\rho_{4,3}(n_t)+J^{-2}=o(n_t^{-1/2})\). Equation
\eqref{eq:example-upper} gives the adaptive bound in
\cref{eq:example-shrinking}; division by the frozen lower bound proves
the stated ratio. All constants are uniform along this sequence.
A concrete smooth member is \(f(w_1,w_2)=w_1w_2\),
for which \(\sigma_f^2=1/9\).
For the target-only theorem, take
\(h_{t,0}(z)=V/4+\delta f(W)\), \(g_{t,0}(x,u,\tau)=u\), and
\(\zeta_t=0\). When \(f\) is in the specified \(C^2\) ball, this
representation has a fixed \(C^2\) bound in \(p=3\) coordinates.
Theorem~\ref{thm:nt} therefore contains the representation exponent
\(4/(4+3)=4/7\). No lower bound for that target-only estimator is
inferred from this comparison.

There is also no contradiction with \cref{thm:freeze}. In this example,
\(U_{t,T}=S_{t,T}/T\) for \(T>0\), so the unregularized conditional
projection has \(\cD_t(h_c)=0\). Nevertheless, its reduced conditional
mean must agree with \(x/\tau\) along the state support at each
positive time \(\tau\). Since \(\delta>0\) and
\(\Var\{f(W)\}>0\), changes in that state support require slope
\(1/\tau\); no fixed state-Lipschitz bound is possible uniformly as
\(\tau\) decreases to zero. Thus the reduced-mean regularity condition
of \cref{ass:transfer} is unavailable. The valid frozen comparison is
\cref{eq:freeze-correctable-main}, whose proof uses the smooth corrected
head as a comparator, not a purportedly admissible approximation of
\(x/\tau\). The endpoint mean defect \(\mathcal I_t(h_c)\) and the
velocity deficiency \(\cD_t(h_c)\) are genuinely different quantities.
\end{proof}

\subsection{Reduced-mean regularity and a plug-in alternative}
\begin{proposition}[A uniform reduced-density criterion]\label{prop:density}
For each \(h\) in a declared class, let
\(Q_h=(S_{t,T},h(Z_{t,0}),Z_{t,c},T)\), with law \(\mu_h\).
Suppose \(\mu_h\) has a density \(q_h\) on \(\cK_t\), and its vector
first-moment measure has a density \(b_h\), meaning
\begin{equation}\label{eq:moment-density}
 \E\{U_{t,T}\ind(Q_h\in A)\}=\int_A b_h(q)\dd q
              \quad\text{for Borel }A\subset\cK_t.
\end{equation}
If \(q_h,b_h\) belong uniformly to fixed \(C^{r_t}\) balls,
\(r_t>1\), and \(q_h\ge c_q>0\) on \(\cK_t\), then
\begin{equation}\label{eq:density-mean}
 g_{t,h}=b_h/q_h
\end{equation}
belongs uniformly to a fixed \(C^{r_t}\) ball. If also
\(\|U_{t,T}\|\le B_{U,t}\), these versions have norm at most
\(B_{U,t}\). Choosing output and block Lipschitz budgets strictly above
the resulting uniform bounds gives the regularity and margin conditions
in \cref{ass:transfer}. This criterion does not by itself imply calibration.
\end{proposition}
\begin{proof}\proofanchor{prop:density}
The defining integral identity for conditional expectation and
\cref{eq:moment-density} imply
\(g_{t,h}(q)q_h(q)=b_h(q)\) for almost every \(q\), giving
\cref{eq:density-mean}. Let \(k=\lceil r_t\rceil-1\) and
\(\alpha=r_t-k\). Since \(q_h\ge c_q\), its reciprocal is uniformly
bounded. Differentiating \(q_hq_h^{-1}=1\) gives, for a nonzero
multi-index \(\nu\) with \(|\nu|\le k\),
\[
 D^\nu(q_h^{-1})=-q_h^{-1}
       \sum_{0<\omega\le\nu}\binom{\nu}{\omega}
              D^\omega q_h\,D^{\nu-\omega}(q_h^{-1}).
\]
Induction on \(|\nu|\) bounds all these derivatives uniformly.
For bounded scalar functions, the H\"older seminorm obeys
\([ab]_\alpha\le\|a\|_\infty[b]_\alpha+
\|b\|_\infty[a]_\alpha\).
Also \(|q_h(x)^{-1}-q_h(y)^{-1}|
 \le c_q^{-2}|q_h(x)-q_h(y)|\).
Applying these inequalities to the differentiated identity bounds the
order-\(k\) H\"older seminorm of \(q_h^{-1}\).
The product rule for \(b_hq_h^{-1}\) now bounds its entire
\(C^{r_t}\) norm, uniformly in \(h\).

The original conditional expectation has norm at most \(B_{U,t}\)
almost everywhere. The continuous ratio version agrees with it almost
everywhere. Since \(q_h\ge c_q\), every relatively open subset of the
rectangle has positive \(\mu_h\)-measure. Continuity therefore extends
the norm bound to the whole rectangle, including its boundary.
The uniform first-derivative bounds bound each block Lipschitz constant;
strictly larger budgets supply fixed margins.
\end{proof}

\begin{proposition}[A single-head plug-in route]\label{prop:plugin}
Instead of reduced-mean regularity and deficiency calibration, suppose the
target comparator in \cref{ass:smooth} exists. On a source-measurable
event \(E_s\) with \(\PP(E_s^c)\le\pi_s\), assume the functional
representation recovery bound
\begin{equation}\label{eq:representation-recovery}
 \E_{Z_t}\|\widehat h_s(Z_{t,0})-h_{t,0}(Z_{t,0})\|^2
 \le C_{\mathrm{id}}\cD_s(\widehat h_s)+\delta_{\mathrm{id}}^2,
\end{equation}
where \(C_{\mathrm{id}},\delta_{\mathrm{id}}\) are deterministic and nonnegative.
Then, under the target bounded-label condition,
\begin{equation}\label{eq:plugin-profile}
 \E Q_{t,N}(\widehat h_s)
 \le C N^{-2r_t/D_t}+3\zeta_t^2
       +3L_{u,t}^2\{C_{\mathrm{id}}\E\cD_s(\widehat h_s)+\delta_{\mathrm{id}}^2\}
       +B_{U,t}^2\pi_s.
\end{equation}
Consequently \cref{thm:freeze} remains valid with its
transferred contribution \(T_s(n_s)\) replaced, up to the common
constant, by
\begin{equation}\label{eq:plugin-transfer-term}
 \zeta_t^2+L_{u,t}^2\{C_{\mathrm{id}}R_s(n_s)+\delta_{\mathrm{id}}^2\}
                       +B_{U,t}^2\pi_s.
\end{equation}
Only the single target head \(g_{t,0}\), not every reduced mean
\(g_{t,h}\), needs smoothness in this route.
\end{proposition}
\begin{proof}\proofanchor{prop:plugin}
Take the admissible neural approximation \(v_N\) of \(g_{t,0}\) in
\cref{lem:approximation}. For a fixed source realization on \(E_s\),
write its field error as the three summands
\begin{align*}
 F_{t,v_N,\widehat h_s}-v_t^0
 ={}&[v_N(x,\widehat h_s(z_0),z_c,\tau)
                 -g_{t,0}(x,\widehat h_s(z_0),z_c,\tau)]\\
 &+[g_{t,0}(x,\widehat h_s(z_0),z_c,\tau)
                 -g_{t,0}(x,h_{t,0}(z_0),z_c,\tau)]\\
 &+[g_{t,0}(x,h_{t,0}(z_0),z_c,\tau)-v_t^0(x,z,\tau)].
\end{align*}
The first has squared norm at most \(CN^{-2r_t/D_t}\), the second at
most \(L_{u,t}^2\E_{Z_t}\|\widehat h_s-h_{t,0}\|^2\), and the third
has squared norm \(\zeta_t^2\). The squared three-term triangle
inequality and \cref{eq:representation-recovery} give the desired bound
on \(Q_{t,N}(\widehat h_s)\) on \(E_s\).
On its complement the zero-head comparator bounds the profile by
\(B_{U,t}^2\). Averaging proves \cref{eq:plugin-profile}.
The frozen proof uses
\cref{ass:transfer} only through the
profile bound in \cref{lem:profile-bound}. Replacing that bound by
\cref{eq:plugin-profile}, and using the same source rate, proves
\cref{eq:plugin-transfer-term}. No deficiency term is added again:
\cref{eq:plugin-profile} already bounds the full profile risk.
\end{proof}

\section{Endpoint validation: detailed guarantees and proofs}
\label{app:endpoint}
This appendix proves the results used in
\cref{sec:endpoint-safeguards}. We retain the notation
\(n=n_{\rm val}\), \(d=d_t\), and
\(A_{\rm e}=D_{\rm e}^{\upsilon}\) from that section.

\subsection{Support, deployment, and a deterministic selection principle}
\begin{lemma}[Bounded endpoint support and solver changes]
\label{lem:endpoint-support}
Under the bounded-state part of \cref{ass:path} and the path condition
\cref{eq:path}, \(G_t,X_t\in[-K_{x,t},K_{x,t}]^d\) almost surely.
Every exact or Euler endpoint of a field bounded by \(B_{V,t}\) lies in
\([-K_{x,t}-B_{V,t},K_{x,t}+B_{V,t}]^d\) almost surely.
More generally, suppose two endpoint dictionaries \(Q_m,\widetilde Q_m\)
satisfy, on a common support of diameter \(D_{\rm e}\),
\(\sup_z W_2(Q_{m,z},\widetilde Q_{m,z})\le\epsilon_{\rm sol}\)
for every \(m\). Then
\begin{equation}\label{eq:solver-risk-change}
 \sup_m|\mathcal R_{\rm W}(Q_m)-\mathcal R_{\rm W}(\widetilde Q_m)|
                  \le2D_{\rm e}\epsilon_{\rm sol}.
\end{equation}
For exact versus Euler integration under \cref{lem:stability}, one may
take \(\epsilon_{\rm sol}=\sqrt{C_{\rm Euler}}/\min_m J_m\).
Thus validating and deploying the same numerical laws avoids a separate
solver-change remainder.
\end{lemma}
\begin{proof}\proofanchor{lem:endpoint-support}
\textbf{Step 1: recover the support of the interpolation endpoints
\(G_t\) and \(X_t\).}
Let \(C_t=[-K_{x,t},K_{x,t}]^d\), and let
\(\mathcal G_t=\sigma(G_t,X_t,Z_t)\). The interpolation path
\(\tau\mapsto S_{t,\tau}=\Psi_{t,\tau}(G_t,X_t)\) is determined by
\(\mathcal G_t\). The interpolation time
\(T\sim\operatorname{Unif}[0,1]\) is independent of
\(\mathcal G_t\), by the sampling construction preceding
\cref{eq:path}. The bounded-state assertion in \cref{ass:path} and
Tonelli's theorem therefore give
\begin{align}
 0
 &=\PP(S_{t,T}\notin C_t)\notag\\
 &=\E\bigl[\E\{\ind(S_{t,T}\notin C_t)\mid\mathcal G_t\}\bigr]
   =\E\int_0^1\ind\{S_{t,\tau}\notin C_t\}\dd\tau.
 \label{eq:endpoint-proof-path-integral}
\end{align}
Define the nonnegative random variable
\(I_t=\int_0^1\ind\{S_{t,\tau}\notin C_t\}\dd\tau\).
Equation~\eqref{eq:endpoint-proof-path-integral} implies
\(I_t=0\) almost surely. Intersect this probability-one event with
the event on which \(\tau\mapsto S_{t,\tau}\) is absolutely
continuous and has the endpoint values in \cref{eq:path}; denote the
intersection by \(\Omega_t^0\).

Fix a realization \(\omega_0\in\Omega_t^0\). Suppose, for a contradiction,
that \(S_{t,\tau_0}\notin C_t\) for some \(\tau_0\in[0,1]\).
Since \(C_t\) is closed,
\(a_0=\operatorname{dist}(S_{t,\tau_0},C_t)>0\), where
\(\operatorname{dist}(x,C_t)=\inf_{y\in C_t}\|x-y\|\).
Continuity of \(\tau\mapsto S_{t,\tau}\) yields a number
\(\varepsilon_0>0\) such that
\[
 \|S_{t,\tau}-S_{t,\tau_0}\|<a_0/2
 \quad\text{for all }\tau\in
 I_0=(\tau_0-\varepsilon_0,\tau_0+\varepsilon_0)\cap[0,1].
\]
For every \(\tau\in I_0\), the triangle inequality implies
\[
 \operatorname{dist}(S_{t,\tau},C_t)
 \ge a_0-\|S_{t,\tau}-S_{t,\tau_0}\|>a_0/2.
\]
The relative interval \(I_0\) has strictly positive Lebesgue length
\(|I_0|\), including when \(\tau_0=0\) or \(\tau_0=1\).
Consequently \(I_t\ge|I_0|>0\), contradicting the definition of
\(\Omega_t^0\). Hence \(S_{t,\tau}\in C_t\) for every
\(\tau\in[0,1]\) on \(\Omega_t^0\). Substituting \(\tau=0,1\)
in \cref{eq:path} proves
\begin{equation}
 \PP(G_t\in C_t,\ X_t\in C_t)=1.
 \label{eq:endpoint-proof-base-target-support}
\end{equation}

\textbf{Step 2: bound the exact and Euler displacements.}
Fix a field \(F\) with \(\sup_{x,z,\tau}\|F(x,z,\tau)\|\le B_{V,t}\),
a context \(z\), and an initial point \(g\in C_t\). Write
\(X_F(\tau;g,z)\) for the solution of \cref{eq:fitted-ode} starting
from \(g\). Its integral equation gives
\begin{equation}
 \|X_F(1;g,z)-g\|
 =\left\|\int_0^1F(X_F(u;g,z),z,u)\dd u\right\|
 \le\int_0^1 B_{V,t}\dd u=B_{V,t}.
 \label{eq:endpoint-proof-exact-displacement}
\end{equation}
For an integer \(J\ge1\), define the Euler iterates by
\(x^F_0=g\) and
\(x^F_{k+1}=x^F_k+J^{-1}F(x^F_k,z,k/J)\),
\(k=0,\ldots,J-1\). Their telescoping sum satisfies
\begin{equation}
 \|x^F_J-g\|
 \le\sum_{k=0}^{J-1}\|x^F_{k+1}-x^F_k\|
 \le\sum_{k=0}^{J-1}\frac{B_{V,t}}{J}=B_{V,t}.
 \label{eq:endpoint-proof-euler-displacement}
\end{equation}
For either endpoint \(y=X_F(1;g,z)\) or \(y=x^F_J\), and each
coordinate \(a\in\{1,\ldots,d\}\), these inequalities imply
\[
 |y_a|\le |g_a|+|y_a-g_a|
       \le K_{x,t}+\|y-g\|\le K_{x,t}+B_{V,t}.
\]
Together with \cref{eq:endpoint-proof-base-target-support}, this proves
the asserted support for every generated endpoint. In particular, the
compact rectangle
\(\mathcal K_{\rm e}=[-K_{x,t}-B_{V,t},K_{x,t}+B_{V,t}]^d\)
contains the target and generated supports; its Euclidean diameter is
\(2\sqrt d(K_{x,t}+B_{V,t})\).

\textbf{Step 3: compare the squared Wasserstein risks of
\(Q_m\) and \(\widetilde Q_m\).}
Write \(\pi_t=\mathcal L(Z_t)\) for the target context marginal.
For this comparison, the common support \(\mathcal K_{\rm e}\)
contains the supports of \(P_z\), \(Q_{m,z}\), and
\(\widetilde Q_{m,z}\), as in \cref{ass:endpoint-support}, and has
diameter at most \(D_{\rm e}\). Fix \(m\) and a context \(z\)
for which these support assertions hold, and put
\[
 a_m(z)=W_2(Q_{m,z},P_z),\qquad
 \widetilde a_m(z)=W_2(\widetilde Q_{m,z},P_z).
\]
The product couplings in the definition \cref{eq:generation-loss}
show that \(0\le a_m(z),\widetilde a_m(z)\le D_{\rm e}\).
The triangle inequality for \(W_2\), applied in both orders, gives
\[
 |a_m(z)-\widetilde a_m(z)|
 \le W_2(Q_{m,z},\widetilde Q_{m,z})\le\epsilon_{\rm sol}.
\]
Factoring the difference of squares therefore yields
\begin{equation}
 |a_m(z)^2-\widetilde a_m(z)^2|
 =(a_m(z)+\widetilde a_m(z))
       |a_m(z)-\widetilde a_m(z)|
 \le 2D_{\rm e}\epsilon_{\rm sol}.
 \label{eq:endpoint-proof-squared-distance-change}
\end{equation}
Using the definitions in \cref{eq:endpoint-risks,eq:generation-loss}
and then \cref{eq:endpoint-proof-squared-distance-change}, we obtain
\begin{align*}
 |\mathcal R_{\rm W}(Q_m)-\mathcal R_{\rm W}(\widetilde Q_m)|
 &\le\int |a_m(z)^2-\widetilde a_m(z)^2|\,\pi_t(\dd z)\\
 &\le 2D_{\rm e}\epsilon_{\rm sol}.
\end{align*}
Taking the supremum over \(m\) proves \cref{eq:solver-risk-change}.
Only target-almost-everywhere bounds in \(z\) are needed for this
integration; the stated uniform bound in \(z\) is sufficient.

\textbf{Step 4: verify the uniform exact-to-Euler comparison.}
For each candidate field \(F_m\), couple its exact and Euler
trajectories by the same initial point \(g\sim\gamma_t\). Put
\(h_m=J_m^{-1}\), \(t_k=kh_m\), and
\(e_{m,k}=\|x^{F_m}_k-X_{F_m}(t_k;g,z)\|\).
Let \(L=L_{x,t}\) and
\(C_0=(L_{\tau,t}+LB_{V,t})/2\). The common envelopes and Lipschitz
bounds in \cref{lem:stability} give
\begin{align*}
 &\|X_{F_m}(t_{k+1};g,z)-X_{F_m}(t_k;g,z)
          -h_mF_m(X_{F_m}(t_k;g,z),z,t_k)\|\\
 &\quad\le\int_{t_k}^{t_{k+1}}
       \{L\|X_{F_m}(u;g,z)-X_{F_m}(t_k;g,z)\|
                           +L_{\tau,t}(u-t_k)\}\dd u\\
 &\quad\le (LB_{V,t}+L_{\tau,t})
                 \int_{t_k}^{t_{k+1}}(u-t_k)\dd u=C_0h_m^2.
\end{align*}
Subtracting the Euler recursion and using the state-Lipschitz bound
with constant \(L\) yields
\(e_{m,k+1}\le(1+Lh_m)e_{m,k}+C_0h_m^2\), with \(e_{m,0}=0\).
Iteration gives, when \(L>0\),
\begin{align}
 e_{m,J_m}
 &\le C_0h_m^2\sum_{k=0}^{J_m-1}(1+Lh_m)^k\notag\\
 &=\frac{C_0}{J_mL}\{(1+L/J_m)^{J_m}-1\}
 \le\frac{C_0(e^L-1)}{J_mL}
 =\frac{\sqrt{C_{\rm Euler}}}{J_m},
 \label{eq:endpoint-proof-uniform-euler}
\end{align}
where the last equality is \cref{eq:euler-constant}. For \(L=0\),
the same recursion gives \(e_{m,J_m}\le C_0/J_m\), which is the
continuous extension of \cref{eq:endpoint-proof-uniform-euler}.
The bound is independent of both \(g\) and \(z\). The common-initial-point
coupling is admissible in \cref{eq:generation-loss}, so
\[
 W_2^2\{P_{F_m}(\cdot\mid z),P_{F_m}^{[J_m]}(\cdot\mid z)\}
 \le\int e_{m,J_m}^2\,\gamma_t(\dd g)
 \le C_{\rm Euler}J_m^{-2}.
\]
Thus \(\epsilon_{\rm sol}=\sqrt{C_{\rm Euler}}/\min_mJ_m\)
is valid in \cref{eq:solver-risk-change}. When the validated kernel
and deployed kernel are the same \(Q_m\), their risk difference is
zero, rather than an additional numerical comparison term.
\end{proof}

\begin{lemma}[Uniform contrasts, selection, and safe switching]
\label{lem:safe-selection}
Let \(R_1,\ldots,R_M\in[0,H]\) be fixed risks, let \(b\) be a
common offset, and suppose an event \(E\) satisfies
\begin{equation}\label{eq:generic-uniform-contrasts}
 \max_m|\widehat l_m-(R_m+b)|\le\epsilon.
\end{equation}
Let \(\widehat m\) minimize \(\widehat l_m\). For a fixed baseline
index \(m_0\), return \(\widehat m\) if
\(\widehat l_{m_0}-\widehat l_{\widehat m}>2\epsilon\), and
otherwise return \(m_0\); call the resulting index \(\widehat m_{\rm safe}\).
On \(E\),
\begin{equation}\label{eq:generic-safe-inequalities}
 R_{\widehat m}\le\min_m R_m+2\epsilon,\qquad
 R_{\widehat m_{\rm safe}}\le R_{m_0},\qquad
 R_{\widehat m_{\rm safe}}\le\min_m R_m+4\epsilon.
\end{equation}
If \(R_{m_0}-\min_mR_m>4\epsilon\), switching occurs on \(E\).
If \(\PP(E^c)\le\delta\), then
\(\E R_{\widehat m}\le\min_mR_m+2\epsilon+H\delta\) and
\(\E R_{\widehat m_{\rm safe}}\le R_{m_0}+H\delta\).
The same statements hold conditionally on information fixing the risks.
\end{lemma}
\begin{proof}\proofanchor{lem:safe-selection}
Let \(R_* =\min_{1\le m\le M}R_m\), and choose an index
\(m_*\in\argmin_{1\le m\le M}R_m\) by a fixed tie rule.
Define the switching event
\[
 A_{\rm sw}=
 \{\widehat l_{m_0}-\widehat l_{\widehat m}>2\epsilon\}.
\]
The index \(\widehat m_{\rm safe}\) equals \(\widehat m\) on
\(A_{\rm sw}\) and \(m_0\) on \(A_{\rm sw}^c\). All comparisons
in Steps~1--3 are made on the event \(E\) in
\cref{eq:generic-uniform-contrasts}.

\textbf{Step 1: the unguarded oracle inequality.}
For every index \(m\), Equation~\eqref{eq:generic-uniform-contrasts}
implies
\(\widehat l_m-b-\epsilon\le R_m\le\widehat l_m-b+\epsilon\).
The simultaneous nature of these inequalities permits their use at the
random index \(\widehat m\). Since
\(\widehat l_{\widehat m}\le\widehat l_{m_*}\),
\begin{equation}
 R_{\widehat m}
 \le\widehat l_{\widehat m}-b+\epsilon
 \le\widehat l_{m_*}-b+\epsilon
 \le R_*+2\epsilon.
 \label{eq:safe-proof-unguarded}
\end{equation}
This is the first inequality in \cref{eq:generic-safe-inequalities}.

\textbf{Step 2: the baseline and oracle inequalities for
\(\widehat m_{\rm safe}\).}
On \(E\cap A_{\rm sw}\), the definition of \(A_{\rm sw}\) and
\cref{eq:generic-uniform-contrasts} give
\begin{equation}
 R_{\widehat m_{\rm safe}}=R_{\widehat m}
 \le\widehat l_{\widehat m}-b+\epsilon
 <\widehat l_{m_0}-b-\epsilon
 \le R_{m_0}.
 \label{eq:safe-proof-switch-baseline}
\end{equation}
On \(E\cap A_{\rm sw}^c\), the returned index is \(m_0\), so
\(R_{\widehat m_{\rm safe}}=R_{m_0}\). These two cases prove the
second inequality in \cref{eq:generic-safe-inequalities}.

For the oracle comparison on \(E\cap A_{\rm sw}\),
Equation~\eqref{eq:safe-proof-unguarded} gives
\(R_{\widehat m_{\rm safe}}\le R_*+2\epsilon\le R_*+4\epsilon\).
For the remaining event \(E\cap A_{\rm sw}^c\), the no-switch
inequality is
\(\widehat l_{m_0}\le\widehat l_{\widehat m}+2\epsilon\).
Consequently
\begin{align}
 R_{\widehat m_{\rm safe}}=R_{m_0}
 &\le\widehat l_{m_0}-b+\epsilon\notag\\
 &\le\widehat l_{\widehat m}-b+3\epsilon
 \le\widehat l_{m_*}-b+3\epsilon
 \le R_*+4\epsilon.
 \label{eq:safe-proof-no-switch-oracle}
\end{align}
Combining the two cases proves the third inequality in
\cref{eq:generic-safe-inequalities}.

\textbf{Step 3: sufficient population improvement for switching.}
Suppose \(R_{m_0}-R_*>4\epsilon\). On \(E\),
\begin{align}
 \widehat l_{m_0}-\widehat l_{\widehat m}
 &\ge\widehat l_{m_0}-\widehat l_{m_*}\notag\\
 &\ge (R_{m_0}+b-\epsilon)-(R_*+b+\epsilon)
 =R_{m_0}-R_*-2\epsilon>2\epsilon.
 \label{eq:safe-proof-improvement}
\end{align}
Thus \(E\subseteq A_{\rm sw}\) under the stated improvement
condition. The strict inequalities in
\cref{eq:safe-proof-switch-baseline,eq:safe-proof-improvement}
match the strict switching threshold \(2\epsilon\); no assertion
of switching is needed at equality.

\textbf{Step 4: expectation inequalities and conditioning.}
The bounds \(0\le R_m\le H\) imply, on every realization,
\(R_{\widehat m}-R_*\le H\) and
\(R_{\widehat m_{\rm safe}}-R_{m_0}\le H\).
Split the first excess risk over \(E\) and \(E^c\), and use
\cref{eq:safe-proof-unguarded}:
\begin{align*}
 \E R_{\widehat m}-R_*
 &=\E\{(R_{\widehat m}-R_*)\ind(E)\}
    +\E\{(R_{\widehat m}-R_*)\ind(E^c)\}\\
 &\le 2\epsilon\PP(E)+H\PP(E^c)
 \le 2\epsilon+H\delta.
\end{align*}
The baseline inequality proved in Step~2 similarly gives
\begin{align*}
 \E R_{\widehat m_{\rm safe}}-R_{m_0}
 &=\E\{(R_{\widehat m_{\rm safe}}-R_{m_0})\ind(E)\}
   +\E\{(R_{\widehat m_{\rm safe}}-R_{m_0})\ind(E^c)\}\\
 &\le 0+H\PP(E^c)\le H\delta.
\end{align*}
These are the two claimed expectation bounds.

For the conditional statements, let \(\mathcal A\) be information
fixing \((R_1,\ldots,R_M)\), the offset \(b\), the baseline index
\(m_0\), and the chosen bounds \(H,\epsilon,\delta\). If
\(\PP(E^c\mid\mathcal A)\le\delta\) almost surely, the same
pathwise inequalities \cref{eq:safe-proof-unguarded,eq:safe-proof-switch-baseline,eq:safe-proof-no-switch-oracle,eq:safe-proof-improvement}
hold. Taking conditional expectations in the preceding decompositions
therefore yields
\[
 \E(R_{\widehat m}\mid\mathcal A)\le R_*+2\epsilon+H\delta,
 \qquad
 \E(R_{\widehat m_{\rm safe}}\mid\mathcal A)\le R_{m_0}+H\delta.
\]
No independence among the empirical contrasts
\(\widehat l_1,\ldots,\widehat l_M\) is used in this lemma.
\end{proof}

\subsection{The conditional energy identity and its Hilbert geometry}
\begin{lemma}[Energy scores as bounded squared Hilbert losses]
\label{lem:energy-hilbert}
For \(\upsilon\in(0,2)\), there is a real separable Hilbert space
\(\mathbb H_{\upsilon}\) and a continuous map
\(\phi:\mathcal K_{\rm e}\to\mathbb H_{\upsilon}\) such that
\begin{equation}\label{eq:energy-feature-distance}
 \|\phi(x)-\phi(y)\|_{\mathbb H_{\upsilon}}^2=\|x-y\|^{\upsilon}.
\end{equation}
For \(\mu_Q=\int\phi(y)Q(\dd y)\),
\begin{align}
 S_{\upsilon}(Q,x)&=\|\mu_Q-\phi(x)\|_{\mathbb H_{\upsilon}}^2,
                       \label{eq:energy-score-hilbert}\\
 \tfrac12\mathsf D_{\upsilon}(P,Q)
             &=\|\mu_P-\mu_Q\|_{\mathbb H_{\upsilon}}^2.
                       \label{eq:energy-distance-hilbert}
\end{align}
On the common support,
\(0\le S_{\upsilon}(Q,x)\le A_{\rm e}\) and
\(0\le\mathsf D_{\upsilon}(P,Q)/2\le A_{\rm e}\).
Moreover \(\mu_P=\mu_Q\) if and only if \(P=Q\).
The conditional identity \cref{eq:conditional-energy-identity} follows,
and \(\mathcal R_{{\rm E},\upsilon}\) is convex under mixtures of
probability kernels. For fixed \(x\), the function
\(\mu\mapsto\exp\{-\eta\|\mu-\phi(x)\|^2\}\) is concave on the
convex hull of the embeddings of supported probability laws whenever
\(0<\eta\le1/(2A_{\rm e})\).
\end{lemma}
\begin{proof}\proofanchor{lem:energy-hilbert}
Fix \(d=d_t\ge1\), \(\upsilon\in(0,2)\), and the compact endpoint
support \(\mathcal K_{\rm e}\) from \cref{ass:endpoint-support}.
Throughout this proof, all probability laws \(P,Q\) under discussion
are supported on \(\mathcal K_{\rm e}\).

\textbf{Step 1: construct the Hilbert space \(\mathbb H_{\upsilon}\)
and the feature map \(\phi\).}
For \(t=(t_1,\ldots,t_d)\in\R^d\), define
\begin{equation}
 I_{d,\upsilon}
 =\int_{\R^d}\frac{1-\cos(t_1)}{\|t\|^{d+\upsilon}}\dd t.
 \label{eq:energy-proof-fourier-constant}
\end{equation}
The value assigned to the integrand at \(t=0\) does not affect this
Lebesgue integral. Let \(\sigma_{d-1}\) be the surface measure of the
unit sphere in \(\R^d\). On \(\{\|t\|\le1\}\),
\(1-\cos(t_1)\le t_1^2/2\le\|t\|^2/2\); on
\(\{\|t\|>1\}\), \(1-\cos(t_1)\le2\). Polar integration gives
\begin{align*}
 I_{d,\upsilon}
 &\le\frac{\sigma_{d-1}}2\int_0^1r^{1-\upsilon}\dd r
       +2\sigma_{d-1}\int_1^\infty r^{-1-\upsilon}\dd r\\
 &=\frac{\sigma_{d-1}}{2(2-\upsilon)}
                       +\frac{2\sigma_{d-1}}{\upsilon}<\infty.
\end{align*}
The integrand in \cref{eq:energy-proof-fourier-constant} is strictly
positive on a neighborhood of \((\pi,0,\ldots,0)\), so
\(I_{d,\upsilon}>0\).

For \(a\in\R^d\setminus\{0\}\), choose an orthogonal matrix
\(O\) with \(Oe_1=a/\|a\|\), where \(e_1\) is the first coordinate
vector. Under the change of variables \(t=Ou/\|a\|\),
\(t^\top a=u_1\), \(\|t\|=\|u\|/\|a\|\), and
\(\dd t=\|a\|^{-d}\dd u\). Hence
\begin{equation}
 \int_{\R^d}\frac{1-\cos(t^\top a)}{\|t\|^{d+\upsilon}}\dd t
 =\|a\|^\upsilon
     \int_{\R^d}\frac{1-\cos(u_1)}{\|u\|^{d+\upsilon}}\dd u
 =I_{d,\upsilon}\|a\|^\upsilon.
 \label{eq:energy-proof-fourier-scaling}
\end{equation}
For \(a=0\), both sides of
\cref{eq:energy-proof-fourier-scaling} are zero.

Choose \(x_\circ\in\mathcal K_{\rm e}\), and define the measure
\(\omega\), the real Hilbert space \(\mathbb H\), and the map
\(\phi\) by
\begin{align}
 \omega(\dd t)&=\frac{\|t\|^{-d-\upsilon}}{2I_{d,\upsilon}}\dd t,
 \qquad
 \mathbb H=\mathbb H_\upsilon=L_2(\omega;\R^2),
 \label{eq:energy-proof-hilbert-space}\\
 \phi(x)(t)&=
 \bigl(\cos(t^\top x)-\cos(t^\top x_\circ),\,
       \sin(t^\top x)-\sin(t^\top x_\circ)\bigr).
 \label{eq:energy-proof-feature-map}
\end{align}
The measure \(\omega\) assigns zero mass to \(\{0\}\) and is finite
on every annulus \(\{k^{-1}\le\|t\|\le k\}\), \(k\ge2\).
Multiplication by
\((2I_{d,\upsilon})^{-1/2}\|t\|^{-(d+\upsilon)/2}\) is an isometry
of \(\mathbb H\) onto \(L_2(\R^d,\dd t;\R^2)\).
The latter space has a countable dense set of rational step functions,
so \(\mathbb H\) is separable.

For real numbers \(r,s\),
\((\cos r-\cos s)^2+(\sin r-\sin s)^2=2\{1-\cos(r-s)\}\).
Using this identity and \cref{eq:energy-proof-fourier-scaling},
\begin{align}
 \|\phi(x)-\phi(y)\|_{\mathbb H}^2
 &=\frac1{I_{d,\upsilon}}
   \int_{\R^d}\frac{1-\cos\{t^\top(x-y)\}}{\|t\|^{d+\upsilon}}\dd t
   =\|x-y\|^\upsilon.
 \label{eq:energy-proof-feature-isometry}
\end{align}
Since \(\phi(x_\circ)=0\), this calculation first verifies that
\(\phi(x)\in\mathbb H\), and then proves
\cref{eq:energy-feature-distance}. It also gives
\[
 \|\phi(x)\|_{\mathbb H}^2=\|x-x_\circ\|^\upsilon\le A_{\rm e},
 \qquad
 \|\phi(x)-\phi(y)\|_{\mathbb H}=\|x-y\|^{\upsilon/2}.
\]
Thus \(\phi\) is continuous and bounded. Separability of
\(\mathbb H\) implies strong measurability of \(\phi\), and
\(\int\|\phi(y)\|_{\mathbb H}Q(\dd y)\le\sqrt{A_{\rm e}}\).
The Bochner mean \(\mu_Q=\int\phi(y)Q(\dd y)\) is therefore
well defined for every supported law \(Q\).

\textbf{Step 2: derive the score and discrepancy identities.}
Let \(a_Q=\int\|\phi(y)\|_{\mathbb H}^2Q(\dd y)\), so
\(0\le a_Q\le A_{\rm e}\). For independent random variables
\(Y,Y'\sim Q\), boundedness permits Fubini's theorem, and hence
\[
 \E\langle\phi(Y),\phi(Y')\rangle_{\mathbb H}
                  =\langle\mu_Q,\mu_Q\rangle_{\mathbb H}.
\]
Expanding the squared norms in
\cref{eq:energy-proof-feature-isometry} gives, for each
\(x\in\mathcal K_{\rm e}\),
\begin{align}
 \E\|Y-x\|^\upsilon
 &=a_Q+\|\phi(x)\|_{\mathbb H}^2
               -2\langle\mu_Q,\phi(x)\rangle_{\mathbb H},\notag\\
 \frac12\E\|Y-Y'\|^\upsilon
 &=a_Q-\|\mu_Q\|_{\mathbb H}^2.
 \label{eq:energy-proof-score-expansion}
\end{align}
Subtracting the second line from the first, exactly as specified in
\cref{eq:energy-score}, proves
\(S_\upsilon(Q,x)=\|\mu_Q-\phi(x)\|_{\mathbb H}^2\), namely
\cref{eq:energy-score-hilbert}.

Define \(a_P=\int\|\phi(x)\|_{\mathbb H}^2P(\dd x)\).
For independent \(X\sim P\) and \(Y\sim Q\), the corresponding
cross-term expansion is
\[
 \E\|X-Y\|^\upsilon
 =a_P+a_Q-2\langle\mu_P,\mu_Q\rangle_{\mathbb H}.
\]
Substitution of this formula and both within-law versions of
\cref{eq:energy-proof-score-expansion} into
\cref{eq:energy-discrepancy} yields
\begin{align*}
 \frac12\mathsf D_\upsilon(P,Q)
 &=a_P+a_Q-2\langle\mu_P,\mu_Q\rangle_{\mathbb H}
       -(a_P-\|\mu_P\|_{\mathbb H}^2)
       -(a_Q-\|\mu_Q\|_{\mathbb H}^2)\\
 &=\|\mu_P-\mu_Q\|_{\mathbb H}^2.
\end{align*}
This proves \cref{eq:energy-distance-hilbert}.
The two squared norms are nonnegative. Jensen's inequality and the
diameter bound in \cref{ass:endpoint-support} further give
\begin{align}
 \|\mu_Q-\phi(x)\|_{\mathbb H}^2
 &\le\int\|\phi(y)-\phi(x)\|_{\mathbb H}^2Q(\dd y)
   =\int\|y-x\|^\upsilon Q(\dd y)\le A_{\rm e},\notag\\
 \|\mu_P-\mu_Q\|_{\mathbb H}^2
 &\le\iint\|\phi(x)-\phi(y)\|_{\mathbb H}^2P(\dd x)Q(\dd y)
   \le A_{\rm e}.
 \label{eq:energy-proof-uniform-bounds}
\end{align}
These are the asserted uniform score and risk bounds.

\textbf{Step 3: prove that the mean embedding \(\mu_Q\) identifies
 the law \(Q\).}
Define the characteristic function of a law \(Q\) by
\(\chi_Q(t)=\int e^{\mathrm i t^\top y}Q(\dd y)\), where
\(\mathrm i^2=-1\). On each finite-measure annulus
\(\{k^{-1}\le\|t\|\le k\}\), Fubini's theorem applies to the
bounded coordinates of \cref{eq:energy-proof-feature-map}. It identifies
a representative of the Bochner mean \(\mu_Q\) as
\begin{equation}
 \mu_Q(t)=
 \bigl(\operatorname{Re}\chi_Q(t)-\cos(t^\top x_\circ),\,
       \operatorname{Im}\chi_Q(t)-\sin(t^\top x_\circ)\bigr)
 \quad\text{for }\omega\text{-almost every }t.
 \label{eq:energy-proof-characteristic-embedding}
\end{equation}
The annuli cover \(\R^d\setminus\{0\}\), so the displayed
identity holds in \(\mathbb H\). If \(\mu_P=\mu_Q\), then
\cref{eq:energy-proof-characteristic-embedding} gives
\(\chi_P(t)=\chi_Q(t)\) for \(\omega\)-almost every \(t\).
The density of \(\omega\) is strictly positive and finite away
from zero, so this equality also holds Lebesgue-almost everywhere.
Both characteristic functions are continuous, by dominated convergence.
A nonzero continuous difference would be nonzero on an open set of
positive Lebesgue measure. Therefore
\(\chi_P(t)=\chi_Q(t)\) for every \(t\in\R^d\).

To verify directly that this equality identifies \(P\) and \(Q\),
let \(\sigma>0\) and let
\(g_\sigma(u)=(2\pi\sigma^2)^{-d/2}
             \exp\{-\|u\|^2/(2\sigma^2)\}\)
be the centered Gaussian density with covariance \(\sigma^2I_d\).
The product of the one-dimensional Gaussian integral identities gives
\[
 g_\sigma(u)=(2\pi)^{-d}\int_{\R^d}
       e^{-\mathrm i t^\top u}e^{-\sigma^2\|t\|^2/2}\dd t.
\]
For either law \(R=P\) or \(R=Q\), define its smoothed density
\(p_{R,\sigma}(u)=\int g_\sigma(u-y)R(\dd y)\).
Since \(\int e^{-\sigma^2\|t\|^2/2}\dd t<\infty\), Fubini's
theorem yields
\begin{equation}
 p_{R,\sigma}(u)
 =(2\pi)^{-d}\int_{\R^d}
        e^{-\mathrm i t^\top u}e^{-\sigma^2\|t\|^2/2}
                  \chi_R(t)\dd t.
 \label{eq:energy-proof-gaussian-identification}
\end{equation}
The equality \(\chi_P=\chi_Q\) therefore implies
\(p_{P,\sigma}=p_{Q,\sigma}\) for every \(\sigma>0\).
Let \(G\sim N(0,I_d)\) be independent of a random vector
\(Y_R\sim R\). For every bounded continuous function
\(f:\R^d\to\R\), dominated convergence gives
\[
 \int f(u)p_{R,\sigma}(u)\dd u
 =\E f(Y_R+\sigma G)\longrightarrow\E f(Y_R)
 =\int f(y)R(\dd y)
 \quad\text{as }\sigma\downarrow0.
\]
Thus \(\int f\dd P=\int f\dd Q\) for every bounded continuous
\(f\). Such functions determine Borel probability laws on
\(\R^d\), so \(P=Q\). Conversely, \(P=Q\) immediately implies
\(\mu_P=\mu_Q\) by the definition of the Bochner mean.

\textbf{Step 4: establish the conditional score identity and mixture
convexity.}
Let \(\pi_t=\mathcal L(Z_t)\). Measurability of the kernels
\(z\mapsto P_z,Q_z\), together with the bounded continuous map
\(\phi\), makes \(z\mapsto\mu_{P_z},\mu_{Q_z}\) measurable.
Fix a context \(z\) for which the support condition holds, and let
\(X\sim P_z\). The vector \(\mu_{P_z}-\phi(X)\) has mean zero.
Expansion of the square gives
\begin{align}
 &\int S_\upsilon(Q_z,x)P_z(\dd x)\notag\\
 &\quad=\|\mu_{Q_z}-\mu_{P_z}\|_{\mathbb H}^2
   +2\left\langle\mu_{Q_z}-\mu_{P_z},
        \int\{\mu_{P_z}-\phi(x)\}P_z(\dd x)
                          \right\rangle_{\mathbb H}\notag\\
 &\qquad\quad+\int\|\mu_{P_z}-\phi(x)\|_{\mathbb H}^2P_z(\dd x)
 \notag\\
 &\quad=\frac12\mathsf D_\upsilon(P_z,Q_z)
                         +\int S_\upsilon(P_z,x)P_z(\dd x).
 \label{eq:energy-proof-conditional-pythagoras}
\end{align}
Here \cref{eq:energy-score-hilbert,eq:energy-distance-hilbert} identify
the last two terms. All integrands are bounded by
\cref{eq:energy-proof-uniform-bounds}; hence integration of
\cref{eq:energy-proof-conditional-pythagoras} against \(\pi_t(\dd z)\)
proves \cref{eq:conditional-energy-identity}. In particular,
\begin{equation}
 \mathcal R_{{\rm E},\upsilon}(Q)
 =\int\|\mu_{P_z}-\mu_{Q_z}\|_{\mathbb H}^2\pi_t(\dd z)
 \in[0,A_{\rm e}].
 \label{eq:energy-proof-conditional-risk}
\end{equation}
The integral in \cref{eq:energy-proof-conditional-risk} is zero if
and only if \(\mu_{P_z}=\mu_{Q_z}\) for \(\pi_t\)-almost every
\(z\). Step~3 makes this equivalent to \(P_z=Q_z\) for
\(\pi_t\)-almost every \(z\), proving the equality assertion in
\cref{eq:conditional-energy-identity}.

For supported kernels \(Q_1,\ldots,Q_M\), choose mixture weights
\(\alpha=(\alpha_1,\ldots,\alpha_M)\) and define the mixture kernel
\(Q^\alpha\) by
\[
 \alpha_m\ge0,\qquad \sum_{m=1}^M\alpha_m=1,\qquad
 Q_z^\alpha=\sum_{m=1}^M\alpha_m Q_{m,z}.
\]
Linearity of the Bochner integral gives
\(\mu_{Q_z^\alpha}=\sum_m\alpha_m\mu_{Q_{m,z}}\).
Put \(a_m(z)=\mu_{Q_{m,z}}-\mu_{P_z}\). The Hilbert identity
\[
 \sum_m\alpha_m\|a_m(z)\|_{\mathbb H}^2
 -\left\|\sum_m\alpha_m a_m(z)\right\|_{\mathbb H}^2
 =\frac12\sum_{m,k}\alpha_m\alpha_k
               \|a_m(z)-a_k(z)\|_{\mathbb H}^2\ge0
\]
and \cref{eq:energy-proof-conditional-risk} imply
\[
 \mathcal R_{{\rm E},\upsilon}(Q^\alpha)
 \le\sum_{m=1}^M\alpha_m\mathcal R_{{\rm E},\upsilon}(Q_m).
\]
This proves convexity under mixtures of the probability kernels \(Q_m\).

\textbf{Step 5: prove exponential concavity on the mean-embedding set.}
Let
\(\mathcal M_{\rm e}=\{\mu_Q:Q(\mathcal K_{\rm e})=1\}\).
Linearity of the mean embedding shows that \(\mathcal M_{\rm e}\)
is convex, so its convex hull is \(\mathcal M_{\rm e}\) itself.
Fix \(x\in\mathcal K_{\rm e}\),
\(0<\eta\le(2A_{\rm e})^{-1}\), and two elements
\(\mu_0,\mu_1\in\mathcal M_{\rm e}\). Define
\(\xi=\mu_1-\mu_0\), \(\mu_s=\mu_0+s\xi\), and
\[
 h(s)=\exp\{-\eta\|\mu_s-\phi(x)\|_{\mathbb H}^2\},
 \qquad 0\le s\le1.
\]
Since \(\mu_s\in\mathcal M_{\rm e}\),
\cref{eq:energy-proof-uniform-bounds} gives
\(\|\mu_s-\phi(x)\|_{\mathbb H}^2\le A_{\rm e}\).
Differentiation and Cauchy--Schwarz yield
\begin{align}
 h''(s)
 &=h(s)\{4\eta^2\langle\mu_s-\phi(x),\xi\rangle_{\mathbb H}^2
                                      -2\eta\|\xi\|_{\mathbb H}^2\}
 \notag\\
 &\le h(s)\{4\eta^2\|\mu_s-\phi(x)\|_{\mathbb H}^2-2\eta\}
                                    \|\xi\|_{\mathbb H}^2\notag\\
 &\le 2\eta h(s)(2\eta A_{\rm e}-1)\|\xi\|_{\mathbb H}^2\le0.
 \label{eq:energy-proof-exponential-concavity}
\end{align}
Therefore \(h(s)\ge(1-s)h(0)+sh(1)\) for every \(s\in[0,1]\).
Since \(\mu_0,\mu_1\) were arbitrary, this proves the stated
concavity of \(\mu\mapsto\exp\{-\eta\|\mu-\phi(x)\|_{\mathbb H}^2\}\)
on the convex hull of the supported mean embeddings.
\end{proof}

\subsection{Simulation scores: unbiasedness and a conditional exponential bound}
\begin{lemma}[Unbiased energy simulation and its exponential error]
\label{lem:energy-mc}
Condition on a context, an observed endpoint \(x\), and a supported
law \(Q\). For \(B\ge2\) independent draws from \(Q\), the score
\(\widehat S\) in \cref{eq:energy-mc-score} satisfies
\begin{align}
 \E(\widehat S\mid Q,x)&=S_{\upsilon}(Q,x),\qquad
                 -A_{\rm e}/2\le\widehat S\le A_{\rm e},
                         \label{eq:energy-mc-unbiased}\\
 \E\left[\exp\{t(\widehat S-S_{\upsilon}(Q,x))\}\mid Q,x\right]
       &\le\exp\{t^2A_{\rm e}^2/(2B)\},\qquad t\in\R.
                         \label{eq:energy-mc-mgf}
\end{align}
Replacing \(B(B-1)\) in the pairwise term by \(B^2\) instead gives
conditional mean
\begin{equation}\label{eq:energy-vstat-bias}
 S_{\upsilon}(Q,x)+\frac1{2B}\E_{Y,Y'\sim Q}\|Y-Y'\|^{\upsilon}.
\end{equation}
This candidate-dependent term must not be discarded when comparing the
original generator laws.
\end{lemma}
\begin{proof}\proofanchor{lem:energy-mc}
Fix a context \(z\), the observed endpoint \(x\in\mathcal K_{\rm e}\),
and the supported probability law \(Q\). All expectations in this
proof are over \(Y_1,\ldots,Y_B\stackrel{\rm iid}{\sim}Q\), unless
another probability law is explicitly introduced. Define
\[
 \rho_\upsilon(y,y')=\|y-y'\|^\upsilon,\qquad
 a(Q,x)=\int\rho_\upsilon(y,x)Q(\dd y),\qquad
 p(Q)=\iint\rho_\upsilon(y,y')Q(\dd y)Q(\dd y').
\]
By \cref{ass:endpoint-support},
\(0\le\rho_\upsilon(y,y')\le A_{\rm e}\) on
\(\mathcal K_{\rm e}\times\mathcal K_{\rm e}\). In this notation,
\cref{eq:energy-score} reads
\(S_\upsilon(Q,x)=a(Q,x)-p(Q)/2\).

\textbf{Step 1: compute the mean and range of \(\widehat S\).}
For each \(b\),
\(\E\{\rho_\upsilon(Y_b,x)\mid Q,x\}=a(Q,x)\).
For each ordered pair \((b,b')\) with \(b\ne b'\), independence
of \(Y_b,Y_{b'}\) gives
\(\E\{\rho_\upsilon(Y_b,Y_{b'})\mid Q,x\}=p(Q)\).
There are exactly \(B(B-1)\) such ordered pairs. Therefore the score
in \cref{eq:energy-mc-score} satisfies
\begin{align}
 \E(\widehat S\mid Q,x)
 &=\frac{B}{B}a(Q,x)-\frac{B(B-1)}{2B(B-1)}p(Q)\notag\\
 &=a(Q,x)-\tfrac12p(Q)=S_\upsilon(Q,x).
 \label{eq:energy-proof-mc-mean}
\end{align}
This calculation uses linearity of expectation, not independence among
the overlapping pair terms \(\rho_\upsilon(Y_b,Y_{b'})\).
Moreover,
\[
 0\le\frac1B\sum_b\rho_\upsilon(Y_b,x)\le A_{\rm e},\qquad
 0\le\frac1{2B(B-1)}\sum_{b\ne b'}
             \rho_\upsilon(Y_b,Y_{b'})\le\frac{A_{\rm e}}2.
\]
Subtracting these two nonnegative terms gives
\(-A_{\rm e}/2\le\widehat S\le A_{\rm e}\), completing
\cref{eq:energy-mc-unbiased}.

\textbf{Step 2: bound the effect of one simulated endpoint.}
Let \(f_x(y_1,\ldots,y_B)\) be the expression in
\cref{eq:energy-mc-score}, with \(X_i=x\) and \(Y_{m,i,b}=y_b\).
Consider two input vectors \(y,\widetilde y\in\mathcal K_{\rm e}^B\)
that agree except at coordinate \(b\). In the first average, only
\(\rho_\upsilon(y_b,x)\) can change, and its change is at most
\(A_{\rm e}\). In the pairwise sum, only the \(2(B-1)\) ordered
pairs involving coordinate \(b\) can change; each change is at most
\(A_{\rm e}\). Thus
\begin{equation}
 |f_x(y)-f_x(\widetilde y)|
 \le\frac{A_{\rm e}}B+
       \frac{2(B-1)A_{\rm e}}{2B(B-1)}
 =\frac{2A_{\rm e}}B=:c_b.
 \label{eq:energy-proof-coordinate-sensitivity}
\end{equation}

\textbf{Step 3: prove the bounded-variable exponential inequality.}
Let \(V\) be a centered real random variable taking values in an
interval \([a,a+c]\) of length \(c\ge0\). Define
\(\psi(t)=\log\E e^{tV}\) for \(t\in\R\). Boundedness of
\(V\) justifies differentiation under the expectation. Under the
exponentially tilted probability measure
\(\dd\PP_t=e^{tV-\psi(t)}\dd\PP\), these derivatives are
\[
 \psi'(t)=\E_t V,\qquad
 \psi''(t)=\Var_t(V),
\]
where \(\E_t\) and \(\Var_t\) are expectation and variance under
\(\PP_t\). The tilted law still assigns probability one to
\([a,a+c]\). Since variance is the minimum squared distance to a
constant,
\[
 0\le\psi''(t)
 \le\E_t\{V-(a+c/2)\}^2\le c^2/4.
\]
The identities \(\psi(0)=0\) and \(\psi'(0)=\E V=0\), followed
by the integral form of Taylor's formula, give, for every real \(t\),
\[
 \psi(t)=t^2\int_0^1(1-s)\psi''(st)\dd s
        \le\frac{t^2c^2}{4}\int_0^1(1-s)\dd s
        =\frac{t^2c^2}{8}.
\]
Consequently
\begin{equation}
 \E e^{tV}\le\exp(t^2c^2/8),\qquad t\in\R.
 \label{eq:energy-proof-hoeffding-mgf}
\end{equation}
The integral formula also covers \(t<0\), so no one-sided restriction
on \(t\) is being imposed. Applying the same argument under a
conditional probability law proves the conditional version of
\cref{eq:energy-proof-hoeffding-mgf} whenever the conditional mean
is zero and the conditional range has length at most \(c\).

\textbf{Step 4: derive the bounded-differences exponential inequality.}
Let \(\Xi_1,\ldots,\Xi_B\) be independent random variables, and
let \(f=f(\Xi_1,\ldots,\Xi_B)\) be a bounded measurable function.
Suppose changing coordinate \(b\) changes \(f\) by at most a
deterministic number \(c_b\). Define
\[
 \mathcal F_b=\sigma(\Xi_1,\ldots,\Xi_b),\qquad
 M_b=\E(f\mid\mathcal F_b),\qquad D_b=M_b-M_{b-1},
 \quad b=1,\ldots,B,
\]
with \(\mathcal F_0\) trivial and \(M_0=\E f\).
Then \(\sum_{b=1}^BD_b=f-\E f\) and
\(\E(D_b\mid\mathcal F_{b-1})=0\).

To verify the conditional range of \(D_b\), fix the preceding
coordinates \((\Xi_1,\ldots,\Xi_{b-1})\), and average \(f\) over
the independent future coordinates \((\Xi_{b+1},\ldots,\Xi_B)\).
Denote the resulting function of coordinate \(u\) by
\[
 g_b(u)=\E_{\Xi_{b+1},\ldots,\Xi_B}
    f(\Xi_1,\ldots,\Xi_{b-1},u,\Xi_{b+1},\ldots,\Xi_B).
\]
The future distribution is the same for \(u\) and \(u'\), by
independence. Thus
\[
 |g_b(u)-g_b(u')|
 \le\E_{\Xi_{b+1},\ldots,\Xi_B}|f(\ldots,u,\ldots)
                                  -f(\ldots,u',\ldots)|\le c_b.
\]
Conditionally on \(\mathcal F_{b-1}\), the variable
\(D_b=g_b(\Xi_b)-\E\{g_b(\Xi_b)\mid\mathcal F_{b-1}\}\)
therefore has a range of length at most \(c_b\). The conditional
version of \cref{eq:energy-proof-hoeffding-mgf} implies
\[
 \E(e^{tD_b}\mid\mathcal F_{b-1})
                          \le\exp(t^2c_b^2/8).
\]
Iterating this inequality gives
\begin{align*}
 \E e^{t\sum_{b=1}^BD_b}
 &=\E\left[e^{t\sum_{b=1}^{B-1}D_b}
                     \E(e^{tD_B}\mid\mathcal F_{B-1})\right]\\
 &\le e^{t^2c_B^2/8}\E e^{t\sum_{b=1}^{B-1}D_b}
 \le\cdots\le\exp\left\{\frac{t^2}{8}\sum_{b=1}^Bc_b^2\right\}.
\end{align*}
Hence the bounded-differences inequality used below is
\begin{equation}\label{eq:bounded-differences-mgf}
 \E e^{t(f-\E f)}
 \le\exp\left\{\frac{t^2}{8}\sum_{b=1}^Bc_b^2\right\},
 \qquad t\in\R.
\end{equation}
The argument is valid under any conditional law making the coordinates
\(\Xi_1,\ldots,\Xi_B\) independent and satisfying the same
coordinate bounds \(c_b\).

Apply \cref{eq:bounded-differences-mgf} under the law conditional on
\((Q,x)\), with \(\Xi_b=Y_b\), \(f=f_x\), and
\(c_b=2A_{\rm e}/B\) from
\cref{eq:energy-proof-coordinate-sensitivity}. Since
\(\sum_{b=1}^Bc_b^2=4A_{\rm e}^2/B\) and
\(\E(f_x\mid Q,x)=S_\upsilon(Q,x)\) by
\cref{eq:energy-proof-mc-mean}, we obtain
\[
 \E\left[e^{t\{\widehat S-S_\upsilon(Q,x)\}}\mid Q,x\right]
 \le\exp\left\{\frac{t^2}{8}\frac{4A_{\rm e}^2}{B}\right\}
 =\exp\{t^2A_{\rm e}^2/(2B)\}.
\]
This proves \cref{eq:energy-mc-mgf} for every \(t\in\R\).

\textbf{Step 5: compute the bias from the denominator \(B^2\).}
Let \(\widetilde S\) be the score obtained by replacing the
pairwise denominator \(2B(B-1)\) with \(2B^2\). The diagonal
terms \(\rho_\upsilon(Y_b,Y_b)\) are zero, so including them in
that sum has no effect. Using the same pair count as in Step~1,
\begin{align*}
 \E(\widetilde S\mid Q,x)
 &=a(Q,x)-\frac{B(B-1)}{2B^2}p(Q)\\
 &=a(Q,x)-\frac12p(Q)+\frac1{2B}p(Q)
 =S_\upsilon(Q,x)+\frac1{2B}p(Q).
\end{align*}
The definition of \(p(Q)\) makes this exactly
\cref{eq:energy-vstat-bias}. To identify what \(\widetilde S\)
scores, let \(\widehat Q_B=B^{-1}\sum_{b=1}^B\delta_{Y_b}\),
where \(\delta_y\) is the point mass at \(y\). Direct substitution
in \cref{eq:energy-score} gives
\[
 S_\upsilon(\widehat Q_B,x)
 =\frac1B\sum_b\rho_\upsilon(Y_b,x)
  -\frac1{2B^2}\sum_{b,b'}\rho_\upsilon(Y_b,Y_{b'})
 =\widetilde S.
\]
Thus \(\widetilde S\) is the exact score of the realized empirical
law \(\widehat Q_B\), but its expectation as a score of the original
law \(Q\) contains the law-dependent term \(p(Q)/(2B)\).
For a dictionary of laws \(Q_m\), the corresponding terms
\(p(Q_m)/(2B)\) need not agree and cannot be treated as a common offset.
\end{proof}

\subsection{Proof of conditional energy-score selection}
\begin{proof}[Proof of \cref{thm:energy-selection}]
\proofanchor{thm:energy-selection}
Fix a confidence level \(\delta\in(0,1)\), and condition throughout
on the pre-validation information \(\cS_{\rm tr}\) specified in
\cref{sec:endpoint-safeguards}. Write
\[
 \PP^\circ(\cdot)=\PP(\cdot\mid\cS_{\rm tr}),\qquad
 \E^\circ[\cdot]=\E_{{\rm val},{\rm sim}}
                            [\cdot\mid\cS_{\rm tr}].
\]
Under this conditional law, the candidate kernels \(Q_1,\ldots,Q_M\)
and the target-only index \(m_0\) in \cref{eq:endpoint-dictionary}
are fixed. Set \(\pi_t=\mathcal L(Z_t)\) and define
\begin{align}
 R_m&=\mathcal R_{{\rm E},\upsilon}(Q_m),\qquad
 R_* =\min_{1\le m\le M}R_m,\qquad
 m_*\in\argmin_{1\le m\le M}R_m,\notag\\
 b_\star&=\int\int S_\upsilon(P_z,x)P_z(\dd x)\pi_t(\dd z),
 \qquad \Lambda_{\rm E}=c_{\rm E}A_{\rm e}.
 \label{eq:selection-proof-fixed-quantities}
\end{align}
Choose \(m_*\) using a fixed tie rule. By
\cref{eq:energy-proof-uniform-bounds,eq:energy-proof-conditional-risk},
\(0\le R_m\le A_{\rm e}\) and \(0\le b_\star\le A_{\rm e}\).
All quantities in \cref{eq:selection-proof-fixed-quantities} are
therefore finite and fixed under \(\PP^\circ\).

\textbf{Step 1: identify the mean and independent units of the
validation scores \(\widehat S_{m,i}\).}
For finite \(B\ge2\), the simulation construction following
\cref{eq:energy-mc-score} and \cref{eq:energy-mc-unbiased} imply
\[
 \E^\circ(\widehat S_{m,i}\mid X_i,Z_i)
                         =S_\upsilon(Q_{m,Z_i},X_i).
\]
For exact scores, the same identity holds because
\(\widehat S_{m,i}=S_\upsilon(Q_{m,Z_i},X_i)\).
The validation pair \((X_i,Z_i)\) is independent of
\(\cS_{\rm tr}\) and has law \(P_t^0\), by
\cref{ass:endpoint-support}. Taking its expectation and then using
\cref{eq:conditional-energy-identity} yields
\begin{align}
 \E^\circ\widehat S_{m,i}
 &=\int\int S_\upsilon(Q_{m,z},x)P_z(\dd x)\pi_t(\dd z)
   =R_m+b_\star.
 \label{eq:selection-proof-row-mean}
\end{align}

For finite \(B\), define the complete validation row
\(V_i=(X_i,Z_i,(Y_{m,i,b})_{1\le m\le M,\,1\le b\le B})\);
for exact scores, define \(V_i=(X_i,Z_i)\).
The rows \(V_1,\ldots,V_n\) are independent under \(\PP^\circ\),
by the stated validation and simulation experiment. For a fixed index
\(m\), the score \(\widehat S_{m,i}\) is a measurable function of
\(V_i\), so the scores \(\widehat S_{m,1},\ldots,\widehat S_{m,n}\)
are independent under \(\PP^\circ\). No independence across
indices \(m\) is required. In particular, common random numbers
within a row \(V_i\) do not affect this argument.

\textbf{Step 2: prove simultaneous concentration over the dictionary.}
For exact scores, \cref{eq:energy-proof-uniform-bounds} gives
\(\widehat S_{m,i}\in[0,A_{\rm e}]\), an interval of length
\(\Lambda_{\rm E}=A_{\rm e}\). For finite \(B\ge2\),
\cref{eq:energy-mc-unbiased} gives
\(\widehat S_{m,i}\in[-A_{\rm e}/2,A_{\rm e}]\), an interval
of length \(\Lambda_{\rm E}=3A_{\rm e}/2\). Thus both cases are
covered by the value of \(c_{\rm E}\) in
\cref{eq:energy-selection-rule}.

Define the centered empirical contrast
\(D_m=\widehat s_m-(R_m+b_\star)\). Equation~\eqref{eq:selection-proof-row-mean}
and the definition of \(\widehat s_m\) in
\cref{eq:energy-selection-rule} give
\[
 D_m=\frac1n\sum_{i=1}^n
               \{\widehat S_{m,i}-\E^\circ\widehat S_{m,i}\}.
\]
Apply the bounded-variable exponential inequality
\cref{eq:energy-proof-hoeffding-mgf} to each centered summand,
under \(\PP^\circ\). Conditional independence across \(i\)
then gives, for every \(u\in\R\),
\begin{align}
 \E^\circ e^{uD_m}
 &=\prod_{i=1}^n\E^\circ
       \exp\left[\frac un
               \{\widehat S_{m,i}-\E^\circ\widehat S_{m,i}\}\right]
 \notag\\
 &\le\prod_{i=1}^n\exp\{u^2\Lambda_{\rm E}^2/(8n^2)\}
 =\exp\{u^2\Lambda_{\rm E}^2/(8n)\}.
 \label{eq:selection-proof-contrast-mgf}
\end{align}
For \(t>0\) and \(u>0\), Markov's inequality applied to
\(e^{uD_m}\) gives
\[
 \PP^\circ(D_m>t)
 \le\exp\{-ut+u^2\Lambda_{\rm E}^2/(8n)\}.
\]
The right side is minimized at \(u=4nt/\Lambda_{\rm E}^2\).
Using \cref{eq:selection-proof-contrast-mgf} also for \(-u\)
therefore gives the two-sided bound
\begin{equation}
 \PP^\circ(|D_m|>t)
 \le2\exp\{-2nt^2/\Lambda_{\rm E}^2\}.
 \label{eq:selection-proof-contrast-tail}
\end{equation}
Use the confidence radius \(\varepsilon_{\rm E}(\delta)\) from
\cref{eq:energy-selection-rule}, and define
\[
 \varepsilon=\varepsilon_{\rm E}(\delta)
 =\Lambda_{\rm E}\sqrt{\frac{\log(2M/\delta)}{2n}},\qquad
 E_\delta=\left\{\max_{1\le m\le M}|D_m|\le\varepsilon\right\}.
\]
A union bound, which does not require independence across \(m\),
gives
\begin{align}
 \PP^\circ(E_\delta^c)
 &\le\sum_{m=1}^M\PP^\circ(|D_m|>\varepsilon)\notag\\
 &\le2M\exp\{-2n\varepsilon^2/\Lambda_{\rm E}^2\}
 =2M\exp\{-\log(2M/\delta)\}=\delta.
 \label{eq:selection-proof-confidence-event}
\end{align}

\textbf{Step 3: derive the three risk inequalities and the switching
condition on \(E_\delta\).}
On \(E_\delta\), the hypotheses of \cref{lem:safe-selection} hold
with the explicit substitutions
\[
 \widehat l_m=\widehat s_m,\quad b=b_\star,\quad H=A_{\rm e},
 \quad\epsilon=\varepsilon,\quad
 \widehat m=\widehat m_{\rm E},\quad
 \widehat m_{\rm safe}=\widehat m_{{\rm E},{\rm safe}}.
\]
The minimizing and switching definitions in that lemma agree exactly
with \cref{eq:energy-selection-rule,eq:energy-safe-rule}. Hence
\cref{eq:generic-safe-inequalities} gives, simultaneously,
\begin{equation}
 \begin{aligned}
 R_{\widehat m_{\rm E}}&\le R_*+2\varepsilon,\\
 R_{\widehat m_{{\rm E},{\rm safe}}}&\le R_{m_0},\\
 R_{\widehat m_{{\rm E},{\rm safe}}}&\le R_*+4\varepsilon.
 \end{aligned}
 \label{eq:selection-proof-eventwise-risks}
\end{equation}
Substitute the definition of \(R_m\) in
\cref{eq:selection-proof-fixed-quantities}, and use
\(Q_{m_0}=Q_{\rm NT}\). Equations~\eqref{eq:selection-proof-confidence-event}
and~\eqref{eq:selection-proof-eventwise-risks} then prove
\cref{eq:energy-selection-oracle,eq:energy-safe-baseline,eq:energy-safe-oracle}
with conditional probability at least \(1-\delta\).

For the claimed switching condition, if \(R_{m_0}-R_*>4\varepsilon\),
then on \(E_\delta\),
\begin{align*}
 \widehat s_{m_0}-\widehat s_{\widehat m_{\rm E}}
 &\ge\widehat s_{m_0}-\widehat s_{m_*}\\
 &\ge (R_{m_0}+b_\star-\varepsilon)
          -(R_*+b_\star+\varepsilon)
 >2\varepsilon.
\end{align*}
The first branch of \cref{eq:energy-safe-rule} is therefore taken.
This establishes the switching assertion on the same event
\(E_\delta\), without introducing another failure probability.

\textbf{Step 4: bound the conditional expectation of the safe risk.}
The second inequality of \cref{eq:selection-proof-eventwise-risks}
holds on \(E_\delta\). On \(E_\delta^c\),
\(R_{\widehat m_{{\rm E},{\rm safe}}}-R_{m_0}\le A_{\rm e}\)
since all candidate risks lie in \([0,A_{\rm e}]\). Thus
\begin{align*}
 &\E^\circ R_{\widehat m_{{\rm E},{\rm safe}}}-R_{m_0}\\
 &\quad=\E^\circ\bigl[
       (R_{\widehat m_{{\rm E},{\rm safe}}}-R_{m_0})\ind(E_\delta)
                         \bigr]\\
 &\qquad\quad+\E^\circ\bigl[
       (R_{\widehat m_{{\rm E},{\rm safe}}}-R_{m_0})\ind(E_\delta^c)
                         \bigr]\\
 &\quad\le0+A_{\rm e}\PP^\circ(E_\delta^c)\le A_{\rm e}\delta.
\end{align*}
Restoring \(R_{m_0}=\mathcal R_{{\rm E},\upsilon}(Q_{\rm NT})\)
proves \cref{eq:energy-safe-expectation}.

\textbf{Step 5: obtain the sharper expectation oracle for
\(\widehat m_{\rm E}\).}
Define \(D_\infty=\max_{1\le m\le M}|D_m|\).
For every \(u>0\), the deterministic inequality
\[
 e^{uD_\infty}\le\sum_{m=1}^M(e^{uD_m}+e^{-uD_m})
\]
and \cref{eq:selection-proof-contrast-mgf} imply
\(\E^\circ e^{uD_\infty}
       \le2M\exp\{u^2\Lambda_{\rm E}^2/(8n)\}\).
Jensen's inequality for the logarithm therefore gives
\begin{align}
 u\E^\circ D_\infty
 &=\E^\circ\log(e^{uD_\infty})
 \le\log\E^\circ e^{uD_\infty}
 \le\log(2M)+\frac{u^2\Lambda_{\rm E}^2}{8n}.
 \label{eq:selection-proof-maximum-mgf}
\end{align}
Dividing by \(u\) and choosing
\(u_*=\sqrt{8n\log(2M)}/\Lambda_{\rm E}\), the minimizer of
the resulting upper bound, yields
\begin{equation}
 \E^\circ D_\infty
 \le\frac{\log(2M)}{u_*}
                +\frac{u_*\Lambda_{\rm E}^2}{8n}
 =\Lambda_{\rm E}\sqrt{\frac{\log(2M)}{2n}}.
 \label{eq:selection-proof-expected-maximum}
\end{equation}
On every realization, without restriction to \(E_\delta\),
\begin{align*}
 R_{\widehat m_{\rm E}}
 &=\widehat s_{\widehat m_{\rm E}}-b_\star
                                      -D_{\widehat m_{\rm E}}\\
 &\le\widehat s_{m_*}-b_\star+D_\infty
 =R_*+D_{m_*}+D_\infty\le R_*+2D_\infty.
\end{align*}
Taking \(\E^\circ\), using \cref{eq:selection-proof-expected-maximum},
and substituting \(\Lambda_{\rm E}=c_{\rm E}A_{\rm e}\) gives
\[
 \E^\circ R_{\widehat m_{\rm E}}
 \le R_*+2c_{\rm E}A_{\rm e}\sqrt{\frac{\log(2M)}{2n}},
\]
which is \cref{eq:energy-selection-expectation}.

Every preceding conditional statement holds for almost every realization
of \(\cS_{\rm tr}\). The proof used \(n\) independent validation
rows \(V_i\), not \(nB\) independent target observations. For each
fixed integer \(B\ge2\), the mean identity
\cref{eq:selection-proof-row-mean} is exact and the same range bound
applies. Consequently, for fixed \(M\), the conditional expected
oracle excess risk \(\E^\circ(R_{\widehat m_{\rm E}}-R_*)\)
converges to zero as \(n\to\infty\), without requiring
\(B\to\infty\).
\end{proof}

\clearpage
\section{Details of the numerical studies}\label{app:numerical}
\subsection{Neural models, training, and validation}\label{app:numerical-training}
All procedures use a two-dimensional encoder with an affine skip and
two hidden ReLU layers of width 64. Encoder outputs are projected onto
the Euclidean ball of radius four. Velocity and score heads have two
hidden ReLU layers of width 64 and receive the response state, encoded
context, and time. Head outputs are clipped to $[-16,16]$ coordinatewise,
and state inputs to the neural residual are clipped to $[-10,10]$.
Layer weights and biases are projected to norm caps 12 and 5.
The diffusion score is the neural residual minus the current state.
Transfer procedures use a source-trained encoder and a fresh target head.
TFM-Freeze and Tian--Shen keep the encoder fixed; adaptation updates it
jointly with the target head. The Tian--Shen comparator implements the
frozen-embedding transfer mechanism of \citet{TianShen2026} with this
common architecture.

Positive-penalty adaptation uses an orthogonally aligned functional anchor,
\begin{equation}\label{eq:num-anchor}
 A_s^{\rm rot}(h)=\min_{Q\in O(2)}\frac1{256}
    \sum_{i=1}^{256}\|h(z_{a,i})-Q\widehat h_s(z_{a,i})\|^2,
\end{equation}
where the anchor contexts are drawn from the source sample. The
orthogonal Procrustes solution is updated during training. This specifies
the functional anchor used in the numerical implementation of
\cref{alg:anchor}.  Positive penalties
are selected from $\{0.01,0.1,1\}$; the separately tuned unpenalized
family fixes $\lambda=0$. The encoder/head learning-rate ratio is
$r=0.1$ for both adaptive families in all reported experiments.

FM uses $G\sim N_d(0,I_d)$ and $T\sim\operatorname{Unif}(0,1)$,
with interpolation $(1-T)G+TX$ and regression response $X-G$.
Diffusion uses the Ornstein--Uhlenbeck interpolation
$e^{-t}X+\sqrt{1-e^{-2t}}G$ on $[0.01,4]$ and uniform-time denoising
score matching, evaluated with log-uniform time sampling and the
corresponding importance weights. Adam uses minibatches of size 128,
head learning rates $\{10^{-3},3\times10^{-4}\}$, encoder weight
decay $10^{-4}$, and gradient-norm clipping at ten. Source encoder/head
learning-rate ratios are $0.3$ and $1$; source and target checkpoints
are $\{500,1500,3000,6000\}$ and $\{250,750,1500,3000\}$ updates.
Source encoders use regularized canonical correlation analysis (CCA)
initialization. Target-only fits compare CCA initialization with
encoder/head ratio $0.1$ and random initialization with ratio $1$.

Source observations are divided into fitting and tuning subsets in
proportions 90/10, with at most 256 held-out rows used for source tuning.
The target fitting, tuning, and validation counts are $(614,205,205)$
in the simulations, $(153,51,52)$ for Materials, and $(1228,410,410)$
for Diamonds. Hyperparameters and checkpoints minimize the ensemble
energy score on the tuning subset, using 48 generated responses and
128 field evaluations per response.

For each TFM procedure, the additional validation step compares its
 tuned candidate with tuned target-only FM. If $D_i$ is the
transfer score minus the target-only score on validation observation
$i$, transfer is retained when
\begin{equation}\label{eq:num-validation}
 \overline D+\Phi^{-1}(0.95)\frac{s_D}{\sqrt{n_g}}<0,
\end{equation}
where $s_D$ is the sample standard deviation and $n_g$ is the validation
sample size. Otherwise target-only FM is retained. Each comparison uses
96 generated responses and 256 field evaluations per response, with
paired random numbers. The three comparisons define separate reported
procedures. Once tuning and validation choices are fixed, the selected
specifications are fitted using all observations supplied to the
repetition, with update counts multiplied by the full-to-fitting
sample-size ratio and rounded upward. A validated procedure uses
exactly its selected transfer or target-only fit and generated samples.
Test observations are excluded from these fitting and selection steps.

\subsection{Construction of the real-data samples}\label{app:numerical-data}
For Materials, the original data contain 21,263 materials and 81
predictors. Element counts are normalized to composition proportions
and rounded to twelve decimal places. One recorded response per
composition is retained using response-independent random priorities.
The domain definitions in \cref{sec:numerical-real} leave 7155 source
and 1347 target compositions. The latter comprise 1011 compositions
eligible for training and the fixed test set of 336 compositions.
Each repetition samples 6000 source and 256 target training compositions.

For Diamonds, we remove twenty records with nonpositive physical
dimensions from the original 53,940 observations. Retaining one record
per exact non-cut predictor vector by response-independent random
priorities leaves 53,476 groups. Before applying the cut-grade domain
definitions, half the groups are randomly assigned to a source-only
cohort. Premium and Very Good cuts in that cohort form the source;
Ideal cuts in the remaining cohort form the target. This construction
yields 12,714 source and 10,585 target groups. The target groups are
divided into 9561 eligible for training and 1024 reserved for testing.
Each repetition samples 12000 source and 2048 target training groups.
The nineteen predictors contain six numerical variables and reference-coded
color and clarity indicators, with reference levels D and I1.
Reported scores are on the log-response scales defined in the main text.
The same 40 training seeds are used for both applications, and methods
share the samples and fixed test set within an application.

\subsection{Evaluation and uncertainty}\label{app:numerical-evaluation}
For $\rho(x,x')=\|x-x'\|/\sqrt d$ and an ensemble
$\widehat X_1,\ldots,\widehat X_M$ generated at a context $z$, the
unbiased energy-score estimate at an observed response $x$ is
\begin{equation}\label{eq:num-score}
 \widehat S(z,x)=\frac1M\sum_{a=1}^M\rho(\widehat X_a,x)
 -\frac1{M(M-1)}\sum_{a<b}\rho(\widehat X_a,\widehat X_b).
\end{equation}
For $d=1$, this is the ensemble CRPS estimator. In the simulations,
an independent ensemble $X_1,\ldots,X_M$ from the true conditional
distribution also permits estimation of the unrooted energy discrepancy,
\begin{equation}\label{eq:num-energy}
 \widehat{\mathcal E}(z)=
 \frac2{M^2}\sum_{a,b}\rho(\widehat X_a,X_b)
 -\frac2{M(M-1)}\sum_{a<b}
 \{\rho(\widehat X_a,\widehat X_b)+\rho(X_a,X_b)\}.
\end{equation}
This unbiased estimate can be slightly negative. For $q=1,2$, empirical
Wasserstein distance is
\begin{equation}\label{eq:num-wasserstein}
 \widehat W_q(z)=\left\{\min_{\pi\in\mathfrak S_M}
 \frac1M\sum_{a=1}^M\rho(\widehat X_a,X_{\pi(a)})^q\right\}^{1/q}.
\end{equation}
Scalar ensembles are sorted; multivariate ensembles use separate
equal-weight assignment problems for each cost. These are joint
empirical distances and retain finite-ensemble error. All criteria
are averaged over test contexts and then over training repetitions.

Final evaluation uses $M=256$ generated responses and 512 field
evaluations per response. FM uses 256 explicit-midpoint ODE steps;
diffusion uses 512 Euler--Maruyama steps on a geometric reverse-time
grid from four to 0.01, with Gaussian initialization and no terminal
denoising. Evaluation streams are shared across methods and independent
of training and selection streams. In the simulations, $B$ and $C$
are obtained once for each $p$ by seeded QR factorization. Source
samples are nested across source sizes within a repetition; target
observations and target-only fits are reused across those sizes.
The 100 training repetitions within each setting are the units for
Monte Carlo standard errors and Student-$t$ intervals.

For the real data, uncertainty is summarized by pointwise 95\% paired
Student-$t$ intervals across the 40 training repetitions. For each method
contrast, let $\Delta_b$ be the difference in mean test CRPS within
repetition $b$. The interval is
$\overline\Delta\pm t_{39,0.975}s_\Delta/\sqrt{40}$, where $s_\Delta$
is the sample standard deviation of the 40 paired differences and
$t_{39,0.975}$ is the 97.5th percentile of the Student-$t$ distribution
with 39 degrees of freedom. This  describes variation across training repetitions conditional on the
observed data pools and fixed target test set. We report Materials and
Diamonds at $r=0.1$.

A harmful repetition has a larger mean test criterion than its matched
target-only fit, with tolerance $10^{-12}$. The criterion is energy
discrepancy for simulations and CRPS for applications. The comparator
is target-only FM for all TFM procedures and target-only diffusion for
Tian--Shen. Exact fallback ties are excluded. Zero harmful repetitions
describes performance in the reported experiment.

\subsection{Supporting comparisons}\label{app:numerical-comparisons}
At $\theta=0.5$ and $n_s=8000$, the paired difference in mean energy
discrepancy between unpenalized adaptation and TFM-Freeze is
$-0.0833$ for $d=1$, with pointwise 95\% interval
$[-0.0892,-0.0774]$. For $d=5$, the
corresponding difference for TFM-Adapt is $-0.0470$, with
interval $[-0.0507,-0.0434]$.
Both adaptive procedures improve on each of the five nonadaptive
comparators in these dimensions after Holm adjustment over the 60
one-sided paired energy contrasts comparing four adaptive procedures
(positive or zero penalty, with or without validation) with five
nonadaptive controls in all three dimensions. Each adjusted $p$-value
for these $d=1,5$ comparisons is below $10^{-4}$.

\Cref{tab:num-validation} summarizes the effect of validation at
$n_s=8000$. At smaller source sizes in the unrelated regime, both
adaptive procedures with validation also have zero harmful repetitions,
except for one out of 100 when $(p,d,n_s)=(50,1,1000)$.
Validation can reject useful transfer: for related scalar responses
at $n_s=8000$, it retains the positive and zero-penalty fits in 49
and 51 repetitions, respectively. These decisions explain the
separation between the adaptive curves with and without validation
in the upper-left panels of the figures.

\begin{table}[htbp]\centering
\caption{Target validation at $n_s=8000$ in the simulations. Retained is the number of repetitions selecting transfer. Harm gives the count before / after validation in the unrelated regime, based on energy discrepancy. All counts are out of 100.}
\label{tab:num-validation}
\setlength{\tabcolsep}{5pt}
\begin{tabular}{rlrrc}
\toprule
 & & \multicolumn{2}{c}{Retained} & Unrelated \\
\cmidrule(lr){3-4}
$d$ & Procedure & Related & Unrelated & Harm \\
\midrule
1 & TFM-Adapt & 49 & 0 & 100 / 0 \\
1 & TFM-Adapt ($\lambda=0$) & 51 & 0 & 100 / 0 \\
\addlinespace
5 & TFM-Adapt & 100 & 0 & 100 / 0 \\
5 & TFM-Adapt ($\lambda=0$) & 100 & 0 & 100 / 0 \\
\addlinespace
10 & TFM-Adapt & 98 & 0 & 94 / 0 \\
10 & TFM-Adapt ($\lambda=0$) & 98 & 0 & 94 / 0 \\
\bottomrule
\end{tabular}
\end{table}

\Cref{tab:num-real-comparisons} gives the Diamonds comparisons that
support the accuracy gains in the main text. Both penalties yield
improvements over every nonadaptive comparator, whereas the difference
between the two adaptive fits is small: $-0.00007$,
with pointwise 95\% paired interval
$[-0.00024,0.00010]$.

\begin{table}[htbp]\centering
\caption{Paired CRPS comparisons on Diamonds. Differences are adaptive procedure minus comparator. Intervals are pointwise 95\% paired Student-$t$ intervals across 40 training repetitions.}
\label{tab:num-real-comparisons}
\setlength{\tabcolsep}{5pt}
\begin{tabular}{lrc}
\toprule
Comparator & Difference & Paired 95\% interval \\
\midrule
\multicolumn{3}{l}{TFM-Adapt} \\
Target-only diffusion & $-0.00706$ & $[-0.00785, -0.00628]$ \\
Target-only FM & $-0.00557$ & $[-0.00638, -0.00475]$ \\
Tian--Shen & $-0.00602$ & $[-0.00672, -0.00532]$ \\
TFM-Freeze & $-0.00241$ & $[-0.00299, -0.00183]$ \\
TFM-Freeze + TV & $-0.00248$ & $[-0.00308, -0.00188]$ \\
\addlinespace
\multicolumn{3}{l}{TFM-Adapt ($\lambda=0$)} \\
Target-only diffusion & $-0.00699$ & $[-0.00778, -0.00621]$ \\
Target-only FM & $-0.00549$ & $[-0.00633, -0.00465]$ \\
Tian--Shen & $-0.00595$ & $[-0.00664, -0.00526]$ \\
TFM-Freeze & $-0.00234$ & $[-0.00290, -0.00177]$ \\
TFM-Freeze + TV & $-0.00241$ & $[-0.00299, -0.00183]$ \\
\bottomrule
\end{tabular}
\end{table}

\clearpage
\begingroup
\setstretch{1.0}
\setlength{\bibsep}{2pt}

\endgroup

\end{document}